\documentclass[3p,authoryear,a4paper]{elsarticle}
\usepackage{graphicx}
\usepackage{slashbox}
\usepackage{color}
\usepackage{dsfont}
\usepackage{amsmath}
\usepackage{amssymb}
\usepackage{amsfonts}
\usepackage{amsbsy}
\usepackage{amsthm}
\usepackage{array}
\usepackage{booktabs} 
\usepackage{verbatim}
\usepackage[english]{babel}
\usepackage{natbib}
\usepackage{ulem}
\usepackage{physics}
\usepackage{booktabs}
\usepackage{enumitem}
\usepackage{mathtools}
\usepackage{url}
\usepackage{multirow}
\usepackage{float}
\usepackage{numprint}
\usepackage{enumitem}
\usepackage{centernot}
\usepackage{circuitikz}
\usepackage{algorithm}
\usepackage{algpseudocode}
\usepackage[title]{appendix}
\usepackage[font=normalsize, labelfont=bf, justification=centering]{caption}
\usepackage{tabularx}
\usepackage{actuarialangle}
\usepackage{ltablex}
\usepackage{makecell}
\usepackage[section]{placeins}
\usepackage[hidelinks]{hyperref}
\usepackage{bbm}
\usepackage[table,xcdraw]{xcolor}

\makeatletter
\def\ps@pprintTitle{
  \let\@oddhead\@empty
  \let\@evenhead\@empty
  \def\@oddfoot{\centerline{\thepage}}
  \def\@evenfoot{\thepage\hfill}}
\makeatother

\numberwithin{table}{section}
\numberwithin{figure}{section}

\newcolumntype{R}{>{\raggedleft\arraybackslash}X} 
\newcolumntype{P}[1]{>{\centering\arraybackslash}p{#1}} 

\renewcommand\appendix{\par
\setcounter{section}{0}
\setcounter{subsection}{0}
\setcounter{table}{0}
\setcounter{figure}{0}
\gdef\thetable{\Alph{table}}
\gdef\thefigure{\Alph{figure}}
\gdef\thesection{\Alph{section}}
\setcounter{section}{0}}

\newtheorem{theorem}{Theorem}[subsection]

\newtheorem{definition}[theorem]{Definition}

\newtheorem{remark}{Remark}[section]
\numberwithin{equation}{section}

\newcounter{arclist}

\newcounter{arcenum}

\graphicspath{{media/}}

\begin{document}

\normalem

\begin{frontmatter}

\title{Machine Learning with Multitype Protected
Attributes: \\ Intersectional Fairness through
Regularisation}

\author[a,b]{Ho Ming Lee}
\ead{homingl@student.unimelb.edu.au}
\author[b,c,d,e]{Katrien Antonio}
\ead{katrien.antonio@kuleuven.be}
\author[a]{Benjamin Avanzi}
\ead{b.avanzi@unimelb.edu.au}
\author[b,d]{Lorenzo Marchi}
\author[a]{Rui Zhou}
\ead{rui.zhou@unimelb.edu.au}

\address[a]{Centre for Actuarial Studies, Department of Economics, University of Melbourne, Australia}
\address[b]{Faculty of Economics and Business, KU Leuven, Belgium}
\address[c]{Faculty of Economics and Business, University of Amsterdam, The Netherlands}
\address[d]{LRisk, Leuven Research Center on Insurance and Financial Risk Analysis, KU Leuven, Belgium}
\address[e]{LStat, Leuven Statistics Research Centre, KU Leuven, Belgium}


\begin{abstract}
Ensuring equitable treatment (fairness) across protected attributes (such as gender or ethnicity) is a critical issue in machine learning. Most existing literature focuses on binary classification, but achieving fairness in regression tasks—such as insurance pricing or hiring score assessments—is equally important. Moreover, anti-discrimination laws also apply to continuous attributes, such as age, for which many existing methods are not applicable. In practice, multiple protected attributes can exist simultaneously; however, methods targeting fairness across several attributes often overlook so-called ``fairness gerrymandering", thereby ignoring disparities among intersectional subgroups (e.g., African-American women or Hispanic men). In this paper, we propose a distance covariance regularisation framework that mitigates the association between model predictions and protected attributes, in line with the fairness definition of demographic parity, and that captures both linear and nonlinear dependencies. To enhance applicability in the presence of multiple protected attributes, we extend our framework by incorporating two multivariate dependence measures based on distance covariance: the previously proposed joint distance covariance (JdCov) and our novel concatenated distance covariance (CCdCov), which effectively address fairness gerrymandering in both regression and classification tasks involving protected attributes of various types. We discuss and illustrate how to calibrate regularisation strength, including a method based on Jensen–Shannon divergence, which quantifies dissimilarities in prediction distributions across groups. We apply our framework to the \texttt{COMPAS} recidivism dataset and a large motor insurance claims dataset.
\end{abstract}

\end{frontmatter}

\section{Introduction}\label{sec:introduction}
\subsection{Background}
Machine learning (ML) has become an essential tool in management and financial decision-making, including customer relationship management, hiring, credit scoring, and insurance underwriting \citep{ngai2009application, gohar2023survey, pessach2023algorithmic, frees2023discriminating}. While ML enhances efficiency and predictive accuracy, it also risks exacerbating unwanted biases embedded in historical data or introducing new biases through model design \citep{mehrabi2021survey, jui2024fairness, xin2024antidiscrimination, charpentier2024insurance}. For instance, facial recognition systems misclassify darker-skinned women at higher rates \citep{buolamwini2018gender}. Such biases may not be obvious and may occur accidentally, which can lead to unfair, unethical, and sometimes even illegal outcomes.

In most of the existing literature, fairness interventions address individual protected attributes (e.g., gender or ethnicity) by typically assuming them as binary. However, real-world fairness challenges are more complex. They typically involve various attribute types, and unfairness can arise at the intersection of multiple attributes (e.g., being both African-American and female) \citep{liu2022practical}. Unfortunately, ensuring fairness across each protected attribute separately does not guarantee fairness for their intersectional subgroups. This issue, known as fairness gerrymandering \citep{kearns2018preventing}, can lead to hidden disparities that existing methods fail to mitigate.

To address these challenges, we propose a regularisation framework that promotes \textit{demographic parity}, thereby reducing disparities of the mean \textit{and} distribution of model predictions across multiple protected attributes. This fairness notion is effective for practical applications as it is easy to explain and interpret for stakeholders, and it is relatively simple to implement both mathematically and algorithmically \citep{barocas2023fairness, lindholm2024fair}. Specifically, demographic parity aims at matching prediction distributions and means across protected groups. 

Formally, our method optimises an objective that balances predictive accuracy and fairness:
\begin{equation}\label{eq:modelsetup1}
\min\{\underbrace{\mathcal{L}}_{\text{Accuracy loss}}+\lambda\cdot\underbrace{\psi}_{\text{Fairness loss}}\},
\end{equation}
The first term of our loss function in Equation~\eqref{eq:modelsetup1} minimises prediction error on the target variable, while the second penalises potential unfairness in the model predictions with respect to the protected attributes. The regularisation parameter $\lambda$ controls the trade-off between these two objectives. When $\lambda = 0$, the model is trained solely for accuracy; as $\lambda$ increases, fairness considerations become more prominent, with sufficiently large values enforcing prediction distributions and means that are nearly identical across all protected (sub)groups. This flexibility allows decision-makers to calibrate the model according to regulatory requirements or any other specific objectives. Furthermore, our proposed method does not require protected attributes to be an argument of the trained model, offering additional adaptability depending on the use case. For instance, constructing an insurance price with gender as an argument is illegal in the EU \citep{eu2004}, making such flexibility particularly valuable in regulated settings. It is important to note, however, that the protected attributes must be included in the training data to compute the regularisation term. 

\subsection{Motivating example}\label{sec:motive}
To motivate and illustrate our approach with a concrete example, we briefly discuss the \texttt{COMPAS} dataset, a benchmark dataset in algorithmic fairness research \citep{angwin2016machine, fabris2022algorithmic}. This dataset includes a non-binary protected attribute \textit{Ethnicity}, covering categories like Hispanic, Asian, and African-American. Some existing fairness interventions simplify ethnicity to a binary category (i.e., Caucasian vs. non-Caucasian) \citep{celis2020data, besse2022survey}, potentially overlooking disparities within non-Caucasian groups. Additionally, some fairness approaches focus on a single protected attribute, such as ethnicity \citep{wadsworth2018achieving, khademi2020algorithmic} or gender \citep{nagpal2024multi}, and thus fail to mitigate biases that arise across intersectional subgroups (e.g., African-American women vs. Caucasian men).

\begin{figure}[h]
    \centering
    \caption{Average predicted recidivism rates under different fairness interventions}
    \includegraphics[width=1\linewidth]{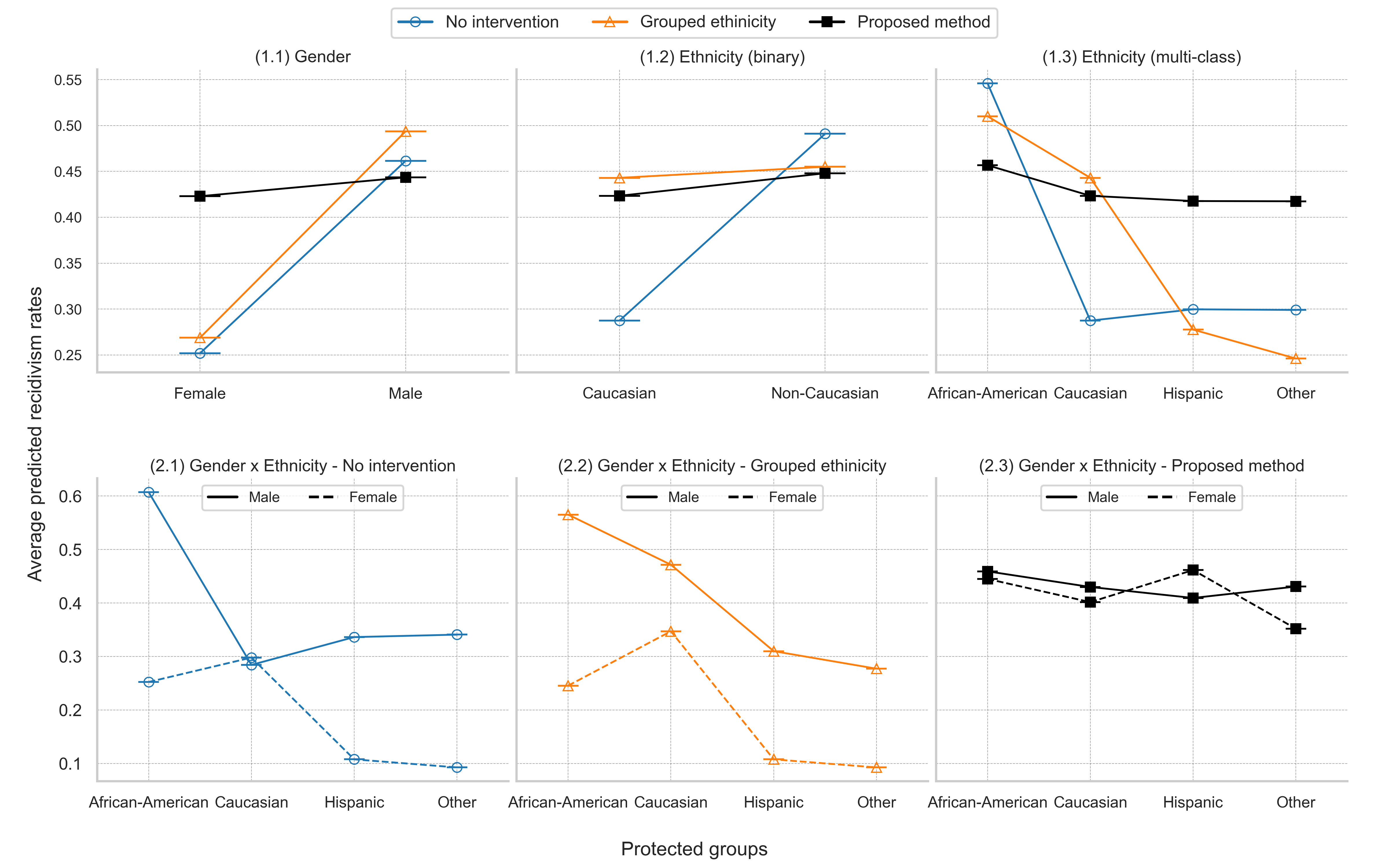}
    \caption*{     \begin{minipage}{0.95\textwidth}     \textit{Note}: This figure shows average predicted recidivism rates across protected groups under three fairness interventions: no regularisation, our framework mimicking existing methods by applying fairness to binary ethnicity only, and our proposed method addressing all protected attributes. Panels (1.1)–(1.3) show average rates by individual attributes, while Panels (2.1)–(2.3) show rates for intersectional subgroups under different fairness interventions. Our method achieves more uniform predicted rates across both groups and intersections.\end{minipage}} 
    \label{fig:compare_methods}
\end{figure}

Figure~\ref{fig:compare_methods} illustrates how different fairness interventions affect average predicted recidivism rates across protected groups. Panels (1.1) to (1.3) show average rates by gender and ethnicity. Without fairness intervention (blue), substantial disparities appear across gender, binary ethnicity, and multi-class ethnicity groups. The existing method (orange) applies our framework but imposes fairness only on binary ethnicity (Caucasian versus non-Caucasian), mimicking current approaches. This reduces disparities between the two ethnic groups (1.2) but fails to address disparities across gender (1.1) and within multi-class ethnicity (1.3). In comparison, our method (black) produces more aligned prediction rates across all individual groups, including multi-class ethnicity. 

Panels (2.1) to (2.3) show predicted rates for intersectional subgroups defined by gender and ethnicity. Only our method yields similar prediction rates across all subgroup combinations, demonstrating its capability to handle multiple categorical protected attributes and address fairness gerrymandering.

We acknowledge that introducing fairness constraints into the model results in a trade-off with predictive accuracy. However, this trade-off can be controlled by explicitly adjusting the regularisation strength $\lambda$, as defined in Equation~\eqref{eq:modelsetup1}.

\subsection{Contributions}
The \texttt{COMPAS} example illustrates the limitations of existing fairness interventions: they often treat protected attributes independently, assume binary protected attributes, and fail to address disparities at the intersectional level. Our proposed framework addresses these challenges; in particular:
\begin{enumerate}
\item We use distance covariance (dCov) \citep{szekely2007measuring} to measure fairness loss. dCov is capable of capturing both linear and nonlinear dependencies between model predictions and protected attributes. It supports protected attributes of any type, including binary, categorical, and continuous, and in any dimension. This addresses the limitations of prior methods that rely on linear dependence measures, such as Pearson’s correlation \citep{beutel2019putting, zhao2022towards}, or are constrained to specific attribute types and tasks \citep{kamishima2011fairness, goel2018non, berk2017convex, bechavod2017penalizing, steinberg2020fast}.

\item We extend fairness regularisation to multiple protected attributes by introducing two alternative regularisers. The first is joint distance covariance (JdCov) \citep{chakraborty2019distance}, which captures mutual dependence between the model output and the joint distribution of protected attributes of any type or dimension, ensuring fairness at both group and subgroup levels. To our knowledge, this is the first use of JdCov as a fairness regulariser. We also propose a novel alternative, which we named concatenated distance covariance (CCdCov), which applies standard distance covariance to a single concatenated vector of protected attributes. CCdCov achieves similar or sometimes better unfairness mitigation while relaxing the strong assumptions required by JdCov when attributes are dependent. Both regularisers are discussed and illustrated in detail in Section~\ref{sec:multifair}. Our framework improves upon prior work by supporting non-binary and continuous protected attributes \citep{kang2022infofair, huang2024bridging, lowy2021stochastic}, and is computationally more efficient than kernel-based fairness methods \citep{liu2022practical}.

\item We develop and discuss practical calibration criteria for choosing the strength of regularisation $\lambda$ in \eqref{eq:modelsetup1}. Such a choice can sometimes be challenging in regularisation tasks. Our recommended solution is based on the \textit{Jensen-Shannon (JS) divergence} \citep{lin1991divergence}, which quantifies divergence between prediction distributions across protected (sub)groups. This provides a visual and interpretable method for calibrating $\lambda$ according to users' needs. Our approach generalises across regression and classification tasks and is compatible with all regularisers introduced in this paper. Additionally, we outline complementary evaluation methods for assessing both fairness and predictive accuracy.

\end{enumerate}

To demonstrate the effectiveness of our methodology, we apply it to real-world datasets, including the \texttt{COMPAS} dataset for recidivism prediction and a motor insurance claims dataset.

\subsection{Paper structure}
Section \ref{sec2} provides more background on the current landscape and literature in fairness-aware ML. Section \ref{sec:regframework} introduces the model, as well as theoretical background and definitions of our proposed regularisers. Section \ref{sec:modelcal_met} discusses model calibration and model performance metrics. Section \ref{sec:appli} discusses our empirical results. Section \ref{sec:conclusion} concludes.

\section{Fairness in machine learning} \label{sec2}
This section introduces the main fairness notion used in this paper, highlights the challenge of fairness gerrymandering, and identifies research gaps in existing fairness regularisation methods.

\subsection{Fairness definitions}\label{sec:fairdef}
Fairness in machine learning is typically categorised into individual fairness and group fairness \citep{barocas2023fairness, caton2024fairness, wan2023processing}. Individual fairness requires that similar individuals receive similar predictions, irrespective of their protected attributes \citep{kusner2017counterfactual, petersen2021post}. Group fairness, in contrast, aims to equalise specific statistical quantities across demographic groups \citep{mehrabi2021survey, hardt2016equality}.

Given increasing regulatory awareness on group fairness in mitigating discrimination \citep{ahrc_ai_insurance_2022,algorithmic_accountability_act_2023, eu_ai_act_2024}, we adopt demographic parity, one of the most widely used fairness criteria \citep{barocas2023fairness, pessach2022review, charpentier2024quantifying}, as our fairness notion. Formally, given a protected attribute $\mathbf{S}\in\mathbb{R}^q$, demographic parity requires that the model prediction $\hat{\mathbf{Y}}\in\mathbb{R}^p$ is independent of $\mathbf{S}$. That is, for all Borel sets $B \subseteq \mathbb{R}^p$, we require:
\begin{equation} \label{dp:dist}
\mathbb{P}(\hat{\mathbf{Y}} \in B \mid \mathbf{S}) = \mathbb{P}(\hat{\mathbf{Y}} \in B) \, \iff \, \hat{\mathbf{Y}} \perp \mathbf{S}.
\end{equation}
This general formulation accommodates continuous, discrete, and multi-dimensional model predictions $\hat{\mathbf{Y}}$. It also allows protected attributes $\mathbf{S}$ to be represented as random vectors, such as one-hot encoded categorical variables, particularly when certain attributes are not well suited to specific encoding methods \citep{potdar2017comparative}. For instance, if $\mathbf{S}$ is a categorical protected attribute with $q$ classes, it can be one-hot encoded into a $q$-dimensional vector:
\begin{equation}
    \mathbf{S} = (S_{1},S_{2},\ldots,S_{q}),\;S_{k}\in\{0,1\},\;\sum_{k=1}^qS_k=1.
\end{equation}

For binary classification tasks, demographic parity is equivalent to requiring that the expected prediction does not vary across protected groups: \begin{equation} \mathbb{E}[\hat{\mathbf{Y}} | \mathbf{S}] = \mathbb{E}[\hat{\mathbf{Y}}]. \end{equation} However, this weaker form of demographic parity may fail in regression tasks where groups have identical means but distinct distributions \citep{charpentier2024quantifying, grari2022fair}. To ensure broader applicability, we adopt strong demographic parity as stated in Equation \eqref{dp:dist}, which enforces full distributional independence, aligning with the statistical properties of our proposed regulariser.

\subsection{Fairness gerrymandering}\label{sec:fairger}
In real-world applications, fairness concerns often extend beyond a single protected attribute to multiple attributes (e.g., gender, age, or ethnicity). However, enforcing fairness for each protected attribute separately does not necessarily prevent disparities at their intersections \citep{yang2020fairness, molina2022bounding, liu2022practical}. Suppose a model considers $d$ protected attributes $\mathbf{S}_1, \mathbf{S}_2, \dots, \mathbf{S}_d$ that can be either scalar-valued or vector-valued random variables. While demographic parity can be imposed on each attribute individually, this does not ensure fairness across their joint distribution, denoted by $(\mathbf{S}_1, \mathbf{S}_2,\cdots,\mathbf{S}_d)$, as:
\begin{equation} 
\begin{aligned}
    &\hat{\mathbf{Y}}\perp \mathbf{S}_1 \;\And\; \hat{\mathbf{Y}} \perp \mathbf{S}_2  \;\And\;  \cdots  \;\And\;  \hat{\mathbf{Y}} \perp \mathbf{S}_d \\
    &\qquad\centernot\implies \hat{\mathbf{Y}} \perp (\mathbf{S}_{i_1}, \mathbf{S}_{i_2},\cdots,\mathbf{S}_{i_k})
\end{aligned} 
\end{equation}
for any $\{i_1,\ldots,i_k\}\subseteq\{1,\ldots,d\}$ with $k\geq 2$. This discrepancy can give rise to fairness gerrymandering \citep{kearns2018preventing}, where a model appears fair across individual protected attributes but exhibits disparities among intersectional subgroups. To address this, recent work has emphasised subgroup fairness, which extends demographic parity to intersections of multiple protected attributes \citep{foulds2020intersectional, ghosh2021characterizing, gohar2023survey}. Formally, subgroup-level demographic parity requires that the model prediction $\hat{\mathbf{Y}}$ is independent of the joint distribution of protected attributes $(\mathbf{S}_1, \mathbf{S}_2, \dots, \mathbf{S}_d)$, such that for all Borel sets $B \subseteq \mathbb{R}^p$:
\begin{equation}\label{dp:dist2} 
\mathbb{P}(\hat{\mathbf{Y}} \in B |(\mathbf{S}_1, \mathbf{S}_2,\cdots,\mathbf{S}_d)) = \mathbb{P}(\hat{\mathbf{Y}} \in B). 
\end{equation}
Additionally, by the law of total probability, ensuring fairness across intersectional subgroups automatically guarantees fairness for each individual protected attribute, as well as for subgroups formed by any subset of protected attributes. Mathematically, this can be expressed as
\begin{equation}\label{jointindep}
\begin{aligned}
    & \hat{\mathbf{Y}} \perp (\mathbf{S}_1, \mathbf{S}_2,\cdots,\mathbf{S}_d)  \implies \hat{\mathbf{Y}} \perp (\mathbf{S}_{i_1}, \mathbf{S}_{i_2}, \ldots,\mathbf{S}_{i_k}) \\
    & \qquad\forall\; \{i_1, i_2, \ldots, i_k\} \subseteq \{1, 2, \ldots, d\}, 1\leq k\leq d.
\end{aligned}
\end{equation}

It is also important to note that fairness violations may arise even when protected attributes are excluded from the model inputs. This is due to \emph{proxy discrimination}, where other variables that are strongly associated with protected attributes encode sensitive information. A well-known example is redlining, where geographic location serves as a proxy for ethnicity, leading to discriminatory outcomes \citep{prince2019proxy, squires2003racial}. In such cases, excluding protected attributes (e.g., ethnicity) does not eliminate the risk of unfairness, as proxy variables (e.g., geographic location) can still introduce disparities across different ethnicity groups. The risk of proxy discrimination is exacerbated by the power of modern ML algorithms. Promoting subgroup fairness helps mitigate this issue by controlling for dependencies between model predictions and the joint distribution of protected attributes, including those indirectly encoded through proxies.

\subsection{Unfairness mitigation strategies} 
Approaches to mitigating unfairness in machine learning fall into three broad categories: pre-processing, in-processing, and post-processing methods \citep{mehrabi2021survey, caton2024fairness, pessach2022review, hort2024bias, rabonato2024systematic}. Pre-processing methods modify the input data to reduce bias before training, using techniques such as fair representation learning \citep{zemel2013learning, creager2019flexibly, guo2022learning} or re-sampling strategies \citep{calders2009building, chakraborty2020fairway}. Post-processing methods adjust model outputs after training, including decision threshold adjustments and distribution transformations \citep{kamiran2012decision, pope2011implementing, chzhen2019leveraging, lindholm2022discrimination, zeng2022bayes}.

In contrast, in-processing methods integrate fairness constraints into training, which allows for an adjustable trade-off between accuracy and fairness. These methods fall into two categories: implicit approaches, such as adversarial learning \citep{xu2021robust, beutel2017data, celis2019improved, grari2022fair}, which rely on black-box adversarial objectives to reduce dependence on protected attributes; and explicit approaches, which incorporate fairness constraints directly into the optimisation objective \citep{zafar2017fairness, donini2018empirical, zafar2019fairness}. 

Our approach is in-processing and explicit, as it uses regularisation to directly control fairness–accuracy trade-offs. We argue that this approach is the most suited for decision-making, as it offers transparency and enables control across the full continuum of trade-offs between accuracy and fairness.

\subsection{Current approaches and limitations}
Most fairness regularisation methods focus on binary classification and discrete protected attributes \citep{hort2024bias, mary2019fairness}. Early methods such as absolute value difference \citep{bechavod2017penalizing}, prejudice remover \citep{kamishima2011fairness} and the negative weighted sum of logs of the predicted probabilities \citep{goel2018non} are effective for classification but do not generalise to continuous attributes or regression tasks.

Another limitation is the inability to capture nonlinear dependencies. Approaches such as absolute correlation \citep{beutel2019putting}, and correlation between predictions and input attributes closely related to the protected attributes \citep{zhao2022towards}, rely on Pearson’s correlation or similar linear metrics. These methods do not ensure full independence between predictions and protected attributes. Moreover, correlation can only capture pairwise associations and cannot capture more complex dependencies or interactions \citep{johnson2002applied, makowski2020methods}, making it ineffective in real world settings where relationships are often nonlinear and multi-dimensional. It is also not suitable for scenarios involving multiple protected attributes, where fairness must be ensured across combinations of variables. 

More recent regularisation-based methods extend fairness constraints to regression and continuous attributes. For instance, \citet{steinberg2020fast} approximate mutual information to enforce fairness in regression tasks but their method remains limited to discrete protected attributes. Similarly, the Hirschfeld-Gebelein-Rényi maximum correlation coefficient used by \citet{mary2019fairness} offers greater flexibility by accommodating both continuous predictions and protected attributes, but remains restricted to a single protected attribute.

Ensuring fairness across multiple protected attributes remains an ongoing challenge, particularly in mitigating fairness gerrymandering \citep{jiang2022generalized, abraham2019fairness}. Some methods address intersectional unfairness by incorporating mutual information with concatenated protected attributes \citep{kang2022infofair} or by applying kernel density estimation-based fairness penalties as regularisers \citep{cho2020fair}. \citet{huang2024bridging} explored distance covariance as a fairness constraint, using the biased V-statistic estimator of distance covariance \citep{szekely2014partial}, and dynamically adjusted the regularisation strength during training to minimise dependence between model predictions and protected attributes. Their approach considers both individual protected attributes separately and their joint distribution, but is restricted to categorical attributes. However, none of these methods simultaneously support multiple protected attribute types, and applies across both regression and classification problems. 

A notable exception is the cross-covariance operator \citep{liu2022practical}, which supports multiple mixed-type protected attributes in both regression and classification tasks by characterising full statistical independence between model predictions and the joint distribution of protected attributes. However, its computational complexity of $O(n^3)$, where $n$ is the sample size, makes it computational inefficient for large-scale applications. In contrast, our proposed regularisers scale with $O(n^2)$ \citep{huo2016fast,chakraborty2019distance}, offering improved efficiency suitable for large-scale use. 

Our work addresses these computational and methodological gaps by introducing a fairness regularisation framework based on joint distance covariance \citep{chakraborty2019distance}. This regulariser efficiently captures nonlinear dependencies and extends fairness constraints to both regression and classification settings. However, joint distance covariance may suffer from numerical instability when protected attributes are dependent. To address this shortcoming, we further introduce \textit{CCdCov}, a regulariser designed to avoid this issue by capturing only the dependence between model predictions and the concatenated protected attributes.

\section{Fairness regulariser framework}\label{sec:regframework}
Building on the discussion of fairness definitions and the challenges associated with multiple protected attributes, we now introduce our regularisation framework. Let us now restate equation \eqref{eq:modelsetup1} using full mathematical notation:
\begin{equation}\label{eq:objective}
\mbox{$
\min\limits_{\Theta}\biggl\{\frac{1}{n}\sum^n_{i=1} 
\mathcal{L}(\hat{\mathbf{y}}_{\Theta i},\mathbf{y}_i) + 
\lambda \cdot \psi\left(\hat{\mathbf{y}}_{\Theta}, \mathbf{s}_1,\ldots\mathbf{s}_d\right) \biggr\}.
$}
\end{equation}
Here, $\mathcal{L}$ denotes the predictive loss function, which can correspond to any type of task (e.g., mean squared error or cross‐entropy); $\Theta$ represents the model parameters; $\hat{\mathbf{y}}_{\Theta i}$ is the $i$-th model prediction; $\mathbf{y}_i$ is the corresponding observed response; $\mathbf{s}_k= \{\mathbf{s}_{ki}\}_{i=1}^n$ for $k\in\{1,\ldots,d\}$ denotes the sample of the $k$-th protected attribute that can be either a scalar or a vector; and $n$ is the total number of samples. 

The function $\psi$ quantifies the dependence between model predictions and protected attributes, and $\lambda$ controls the regularisation strength. We quantify the dependence between model predictions $\hat{\mathbf{Y}}$ and protected attributes $\mathbf{S}$ using distance covariance, which not only captures both linear and nonlinear associations between random vectors of arbitrary dimensions, but is also uniquely zero \textit{if and only if} the variables are independent \citep{szekely2007measuring, huang2024bridging}. Additionally, it applies to a broad class of data types, including continuous, discrete, and categorical variables \citep{szekely2007measuring, shen2022chi}. We further detail the choice of $\psi$ in Sections~\ref{sec:single_reg} and~\ref{sec:multifair}. For completeness, Appendix \ref{apx:dcov} provides the formal definition and the unbiased estimator of distance covariance.

\subsection{Regularising a single protected attribute}\label{sec:single_reg}
We first consider the case where a single protected attribute $\mathbf{S}$ is present. Let $\hat{\mathbf{y}} = \{\hat{\mathbf{y}}_{\Theta i}\}_{i=1}^n$ denote the set of model predictions. We define the regularisation term as:
\begin{equation}
    \psi(\hat{\mathbf{y}}, \mathbf{s}) = \widetilde{dCov}^2\left(\hat{\mathbf{y}},\mathbf{s}\right).
\end{equation}
where $\widetilde{dCov}^2(\cdot,\cdot)$ is the unbiased estimator of distance covariance \citep{szekely2014partial}. Notably, it can be computed in $O(p_0n^2)$ time \citep{szekely2009brownian,huo2016fast}, where $p_0=p+q$ with $\hat{\mathbf{Y}}\in\mathbb{R}^p$ and ${\mathbf{S}}\in\mathbb{R}^q$. This makes our approach significantly more computationally efficient than the $O(n^3)$ complexity of the cross-covariance operator proposed by \citet{liu2022practical}.

\subsection{Extending to multiple protected attributes}\label{sec:multifair}
When multiple protected attributes are present, statistical associations can arise at both the pairwise and higher-order levels.

First, \textit{pairwise associations} arise between individual variables. These include dependencies between the model predictions $\hat{\mathbf{Y}}$ and each individual protected attribute, as well as between the protected attributes themselves. Figure~\ref{fig:asso1} (left panel) illustrates these associations in a simplified setting with two protected attributes: black arrows represent dependencies between $\hat{\mathbf{Y}}$ and each of $\mathbf{S}_1$ and $\mathbf{S}_2$, while the grey dashed arrow indicates the dependency between $\mathbf{S}_1$ and $\mathbf{S}_2$.

Second, \textit{higher-order associations} involve dependencies between joint distributions of variable subsets. Let $\mathbf{S}_1,\dots, \mathbf{S}_d$ denote $d \geq 2$ protected attributes, and let $\hat{\mathbf{Y}}$ denote the model prediction. These higher-order associations include not only dependencies between $\hat{\mathbf{Y}}$ and the joint distribution of any subset of protected attributes (e.g., $(\mathbf{S}_{i_1}, \ldots, \mathbf{S}_{i_k})$ for any index set $\{i_1,\dots,i_k\} \subseteq \{1,\dots,d\}$), but also between disjoint subsets of protected attributes themselves, or even between subsets involving $\hat{\mathbf{Y}}$ (e.g., between $(\hat{\mathbf{Y}}, \texttt{Gender})$ and $\texttt{Ethnicity}$ in the context of the protected attributes introduced in Section~\ref{sec:motive}).

In this paper, we focus specifically on dependencies between $\hat{\mathbf{Y}}$ and the joint distribution of any subset of protected attributes. Capturing these associations is essential for addressing fairness gerrymandering, as it ensures that disparities are not masked within intersectional subgroups. Figure~\ref{fig:asso1} (right panel) illustrates these higher-order associations in the same simplified setting: solid black arrows indicate the dependencies we aim to regularise, while grey dashed arrows represent additional higher-order associations that we acknowledge but do not intend to regularise.

\begin{figure}[!ht]
\centering
\caption{Pairwise and higher-order associations between model prediction and protected attributes \label{fig:asso1}}
{\resizebox{0.6\textwidth}{!}{%
\begin{circuitikz}
\tikzstyle{every node}=[font=\LARGE]
\node [font=\LARGE] at (10,-0.5) {$\hat{\mathbf{Y}}$};
\node [font=\LARGE] at (7.5,-4.5) {$\mathbf{S}_1$};
\node [font=\LARGE] at (12.5,-4.5) {$\mathbf{S}_2$};
\draw [ color={rgb,255:red,0; green,0; blue,0}, line width=2pt, <->, >=Stealth] (7.5,-4) -- (9.75,-1);
\draw [ color={rgb,255:red,0; green,0; blue,0}, line width=2pt, <->, >=Stealth] (10.25,-1) -- (12.5,-4);
\draw [ color={rgb,255:red,150; green,150; blue,150}, line width=2pt, <->, >=Stealth, dashed] (8,-4.5) -- (12,-4.5);
\node [font=\LARGE] at (18.75,-0.5) {$\hat{\mathbf{Y}}$};
\node [font=\LARGE] at (16.25,-4.5) {$\mathbf{S}_1$};
\node [font=\LARGE] at (21.25,-4.5) {$\mathbf{S}_2$};
\draw [ color={rgb,255:red,150; green,150; blue,150}, line width=2pt , rounded corners = 22.5, rotate around={55:(17.5, -2.5)}, dashed] (14.5,-1.75) rectangle  (20.5,-3.25);
\draw [ color={rgb,255:red,150; green,150; blue,150}, line width=2pt , rounded corners = 22.5, rotate around={-55:(20, -2.5)}, dashed] (17,-1.75) rectangle  (23,-3.25);
\draw [ line width=2pt , rounded corners = 18.5] (15.5,-3.75) rectangle  (22,-5.1);
\draw [line width=2pt, <->, >=Stealth] (18.75,-3.75) -- (18.75,-1);
\draw [color={rgb,255:red,150; green,150; blue,150}, line width=2pt, <->, >=Stealth, dashed] (20.75,-4.5) -- (18.1,-3);
\draw [color={rgb,255:red,150; green,150; blue,150}, line width=2pt, <->, >=Stealth, dashed] (16.75,-4.5) -- (19.4,-3);
\end{circuitikz}
}}
\vspace{0.2cm}     \caption*{     \begin{minipage}{0.95\textwidth}     \textit{Note}: This figure illustrates associations between the model prediction $\hat{\mathbf{Y}}$ and protected attributes $\mathbf{S}_1$ and $\mathbf{S}_2$. The left panel shows pairwise associations: between $\hat{\mathbf{Y}}$ and each protected attribute (black arrows), and between the protected attributes themselves (grey dashed arrow). The right panel shows higher-order associations, including the association between $\hat{\mathbf{Y}}$ and the joint distribution $(\mathbf{S}_1, \mathbf{S}_2)$ (black arrow), as well as other joint dependencies (grey dashed arrows) that may be present but should ideally not be in scope for regularisation. \end{minipage}}
\end{figure}

To implement these ideas, we next propose three regularisers to measure and mitigate unfairness in model predictions when multiple protected attributes are present. We formulate our proposed regularisers using $d \geq 2$ protected attributes, $\mathbf{S}_1, \ldots, \mathbf{S}_d$. As in the single-attribute case, these attributes may be discrete, continuous, or categorical, and may differ in type.


\subsubsection{Summing separate distance covariances}\label{reg:sep}
A naive approach is to sum the unbiased estimators of the pairwise distance covariances separately, that is, to define the regulariser as:
\begin{equation}\label{eq:sepreg}
    \begin{aligned}
    \psi(\hat{\mathbf{y}},\mathbf{s}_1,\ldots,\mathbf{s}_d) =&\sum_{k=1}^d 
    \lambda_k \widetilde{dCov}^2(\hat{\mathbf{y}},\textbf{s}_k).
    \end{aligned}
\end{equation}
We can set each $\lambda_k$ as a distinct tuning parameter or use a common value for all. This regulariser accounts only for the pairwise dependencies between $\hat{\mathbf{Y}}$ and each protected attribute separately, as represented by the solid black arrows in the left panel of Figure~\ref{fig:asso1}. However, it overlooks the dependence between $\hat{\mathbf{Y}}$ and the joint distribution of the protected attributes, which may lead to unaddressed disparities across intersectional subgroups and fail to mitigate fairness gerrymandering, as discussed in Section~\ref{sec:fairger}. Moreover, calibrating multiple $\lambda$ values is nontrivial. Hence, we do not include this regulariser in our applications, as our objective is to reduce disparities across joint subgroups rather than marginal attributes alone.

\subsubsection{Utilising the joint distance covariance}\label{sec:jdreg}
To address the disparities across intersectional subgroups, we consider the joint distance covariance introduced by \citet{chakraborty2019distance}, which captures the mutual association between random vectors. Unlike \citet{huang2024bridging}, which considers only multiple categorical protected attributes, this regulariser accommodates mixed-type protected attributes by accounting both types of associations discussed earlier in this section: the pairwise associations between each pair of random vectors and the higher-order associations among joint distributions of disjoint subsets. In the case with two protected attributes, this corresponds to all the black and grey associations illustrated in Figure~\ref{fig:asso1}.  Formal definitions of mutual association, joint distance covariance, and the bias-corrected estimator are provided in Appendix~\ref{apx:jdcov}. In addition, its estimator can be computed with time complexity $O(p_0n^2)$, where $p_0 = \text{dim}(\hat{\mathbf{Y}}) + \sum_{k=1}^d\text{dim}(\mathbf{S}_k)$.

To achieve this, we incorporate the bias-corrected estimator of joint distance covariance, $\widetilde{JdCov}^2(\cdots)$, as a fairness regulariser:
\begin{equation}
\psi(\hat{\mathbf{y}},\mathbf{s}_1,\ldots,\mathbf{s}_d) =  \widetilde{JdCov}^2(\hat{\mathbf{y}},\mathbf{s}_1,\ldots,\mathbf{s}_d).
\end{equation}
By definition, the bias-corrected estimator of joint distance covariance expands as follows:
\begin{equation}\label{jdcov_expand}
\mbox{$
\begin{aligned}
&\widetilde{JdCov}^2(\hat{\mathbf{y}},\mathbf{s}_1,\ldots,\mathbf{s}_d)=
\sum_{k=1}^d\widetilde{dCov}^2(\hat{\mathbf{y}},\mathbf{s}_k)\\
&\qquad+\sum_{1 \leq k < l \leq d}\widetilde{dCov}^2(\mathbf{s}_k,\mathbf{s}_l)+\zeta(\hat{\mathbf{y}}, \mathbf{s}_1,\ldots,\mathbf{s}_d), 
\end{aligned}$}
\end{equation}
where we provide the full decomposition and derivation in Appendix \ref{apx:jdcov}. In contrast with Equation \eqref{eq:sepreg}, which contains only the distance covariances between the predictions and each individual protected attribute, Equation \eqref{jdcov_expand} also adds (i) the pairwise covariances between the protected attributes themselves and (ii) the extra term $\zeta$ that captures joint dependencies involving three or more variables when $d\geq2$. As a result, this regulariser penalises both marginal associations between model predictions and individual protected attributes, as well as dependencies between model predictions and the joint distribution of the protected attributes, effectively mitigating fairness gerrymandering.

However, this regulariser captures more than just the associations we intend to penalise. As illustrated by the grey dashed arrows in Figure~\ref{fig:asso1}, it also accounts for associations between the protected attributes themselves, as well as dependencies between two joint distributions: one involving the model predictions and a subset of protected attributes, and the other involving the remaining protected attributes. When the protected attributes are not independent, these additional associations become significant. The associations among the protected attributes exclusively do not affect our optimisation process under gradient descent, since they are constant with respect to the model predictions, as shown in the decomposition in Equation~\eqref{jdcov_expand}. Consequently, higher-order associations (e.g. the grey associations on the right panel of Figure \ref{fig:asso1}) that involve the model output introduce dependencies that cannot be eliminated by adjusting $\hat{\mathbf{Y}}$ during training. Because the regulariser penalises these dependencies, and because such associations will not vanish regardless of changes to $\hat{\mathbf{Y}}$, the objective cannot theoretically reach zero. As $\lambda$ increases, the model continues to penalise these unavoidable terms, which will likely lead to numerical instability for large $\lambda$s. This issue becomes more prominent as the number of protected attributes increases, because the number of higher-order interactions and joint dependencies grows combinatorically. This increases the likelihood that the regulariser penalises associations that arise from the protected attributes themselves and that cannot be addressed through changes to the model predictions alone.

\subsubsection{Concatenating multiple protected attributes}\label{reg:cat}
To address the drawbacks and potential instability discussed immediately above, we introduce an alternative regulariser, \textit{CCdCov}, which concatenates the protected attributes into a single random vector and computes the distance covariance between this vector and the model predictions. Figure~\ref{fig:asso2} illustrates how the regulariser captures all relevant associations: the black arrows represent dependencies between the model prediction $\hat{\mathbf{Y}}$ and each individual protected attribute, while the grey dashed arrow captures the dependence with their joint distribution. The left panel shows the case with two protected attributes, and the right panel generalises this to any number of protected attributes. This construction allows the regulariser to penalise both marginal and joint associations with the protected attributes. Importantly, it avoids the additional dependencies considered by JdCov. As a result, the regulariser can in theory reach zero, making it less sensitive to numerical instability as the regularisation strength increases. It also focuses exclusively on the dependencies of interest.

Formally, we concatenate the protected attributes of any type into the random vector $(\mathbf{S}_1,\ldots, \mathbf{S}_d)$, and compute the distance covariance between this joint attribute and model predictions $\hat{\mathbf{Y}}$. The resulting regulariser is:
\begin{equation} 
\mbox{$\begin{aligned}
\psi(\hat{\mathbf{y}},\mathbf{s}_1,\ldots,\mathbf{s}_d) &= {CCdCov}(\hat{\mathbf{y}},(\mathbf{s}_1,\ldots,\mathbf{s}_d)).
\end{aligned}
$}
\end{equation}
To illustrate this construction, suppose $\mathbf{A}\in\mathbb{R}^{p_1}$ and $\mathbf{B}\in\mathbb{R}^{p_2}$ are two random vectors with samples $\mathbf{a}=\{(a_1,a_2,...,a_{p_1})_i\}_{i=1}^n$ and $\mathbf{b}=\{(b_1,b_2,...,b_{p_2})_i\}_{i=1}^n$, respectively. When we concatenate the samples $\mathbf{a}$ and $\mathbf{b}$, we construct a sample of size $n$ with random vectors in $\mathbb{R}^{p_1 + p_2}$ that represents realisations from the joint distribution $(\mathbf{A},\mathbf{B})$. Specifically, we have:
\begin{equation}
    (\mathbf{a},\mathbf{b}) = \{(a_1,a_2,...,a_{p_1},b_1,b_2,...,b_{p_2})_i\}_{i=1}^n.
\end{equation}
The following theorem demonstrates how the regulariser decomposes into marginal and joint terms:
\begin{theorem}\label{thm:1}
Let ${\mathbf{s}}=(\mathbf{s}_1,\ldots,\mathbf{s}_d)$ be the concatenated sample of the protected attributes and let ${\mathbf{s}}_{i,k}$ be the $i$-th sample of $\mathbf{S}_k$. We have the following decomposition:
\begin{equation}
    \begin{aligned}
        &CCdCov(\hat{\mathbf{y}},(\mathbf{s}_1,\ldots,\mathbf{s}_d))=\sum_{k=1}^d \widetilde{dCov}^2(\hat{\mathbf{y}},\mathbf{s}_k)+\eta(\hat{\mathbf{y}},{\mathbf{s}}),
    \end{aligned}
\end{equation}
where
\begin{equation}
\mbox{$\begin{aligned}
&\eta(\hat{\mathbf{y}},{\mathbf{s}})= -\frac{1}{n(n-3)}\Biggl(\sum^n_{i=1}\sum^n_{j=1}|\hat{\mathbf{y}}_i-\hat{\mathbf{y}}_j|\xi(i,j)\\
&\qquad+\frac{1}{(n-1)(n-2)}\biggl(\sum^n_{i=1}\sum^n_{j=1}|\hat{\mathbf{y}}_i-\hat{\mathbf{y}}_j|\biggr)\biggl(\sum^n_{i=1}\sum^n_{j=1}\xi(i,j)\biggr)\\
&\qquad -\frac{2}{n-2}\sum_{i=1}^n\biggl(\sum^n_{j=1}|\hat{\mathbf{y}}_i-\hat{\mathbf{y}}_j|\biggr)\biggl(\sum^n_{j=1}\xi(i,j)\biggr)\Biggr).
\end{aligned}$}
\end{equation}
with
\begin{equation}
\mbox{$
    \xi(i,j) = \begin{cases}
        - \frac{\sum_{1\leq k,l\leq d}2|{\mathbf{s}}_{i,k}-{\mathbf{s}}_{j,k}||{\mathbf{s}}_{i,l}-{\mathbf{s}}_{j,l}|}{|{\mathbf{s}}_{i}-{\mathbf{s}}_{j}|+\sum_{k=1}^d|{\mathbf{s}}_{i,k}-{\mathbf{s}}_{j,k}|}&,{\mathbf{s}}_i\neq{\mathbf{s}}_j\\
        0&, {\mathbf{s}}_i={\mathbf{s}}_j
    \end{cases},\;i,j\in\{1,\ldots n\}$.
}
\end{equation}
\end{theorem}
\begin{proof}
The full proof is provided in Appendix \ref{apx:thm1}. We also provide numerical examples of the decomposition in Online Appendix A.
\end{proof}

\begin{figure}[!ht]
\centering
\caption{Association captured by CCdCov \label{fig:asso2}}
{\resizebox{0.7\textwidth}{!}{%
\begin{circuitikz}
\tikzstyle{every node}=[font=\LARGE]
\node [font=\LARGE] at (10,-0.5) {$\hat{\mathbf{Y}}$};
\node [font=\LARGE] at (7.5,-4.5) {$\mathbf{S}_1$};
\node [font=\LARGE] at (12.5,-4.5) {$\mathbf{S}_2$};
\draw [line width=2pt, <->, >=Stealth] (7.5,-4) -- (9.75,-1);
\draw [line width=2pt, <->, >=Stealth] (10.25,-1) -- (12.5,-4);
\draw [color={rgb,255:red,150; green,150; blue,150},line width=2pt, <->, >=Stealth, dashed] (10,-4) -- (10,-1);
\draw [color={rgb,255:red,150; green,150; blue,150}, line width=2pt , rounded corners = 15.0, dashed] (6.75,-4) rectangle  (13.25,-5);
\node [font=\LARGE] at (21,-0.5) {$\hat{\mathbf{Y}}$};
\node [font=\LARGE] at (17,-4.5) {$\mathbf{S}_1$};
\node [font=\LARGE] at (25,-4.5) {$\mathbf{S}_d$};
\draw [line width=2pt, <->, >=Stealth] (17,-4) -- (20.5,-1);
\draw [line width=2pt, <->, >=Stealth] (21.5,-1) -- (25,-4);
\draw [color={rgb,255:red,150; green,150; blue,150}, line width=2pt, <->, >=Stealth, dashed] (21,-4) -- (21,-1);
\draw [color={rgb,255:red,150; green,150; blue,150}, line width=2pt , rounded corners = 15.0, dashed] (16.25,-4) rectangle  (25.75,-5);
\draw [line width=2pt, <->, >=Stealth] (19,-4) -- (20.75,-1);
\draw [line width=2pt, <->, >=Stealth] (21.25,-1) -- (23,-4);
\node [font=\LARGE] at (19,-4.5) {$\mathbf{S}_2$};
\node [font=\LARGE] at (23,-4.5) {$\mathbf{S}_{d-1}$};
\node [font=\LARGE] at (21,-4.6) {$\cdots$};
\end{circuitikz}
}} \vspace{0.2cm}   \caption*{     \begin{minipage}{0.95\textwidth}     \textit{Note}: This figure illustrates the associations captured by CCdCov. The left panel shows the case with two protected attributes, where the regulariser captures associations between $\hat{\mathbf{Y}}$ and $\mathbf{S}_1$ and between $\hat{\mathbf{Y}}$ and $\mathbf{S}_2$, as well as the joint distribution of $(\mathbf{S}_1, \mathbf{S}_2)$. The right panel generalises this to the case where $d > 2$ protected attributes are present. In this setting, the regulariser continues to capture associations between $\hat{\mathbf{Y}}$ and each protected attribute, along with the joint distribution of any subset of the protected attributes by virtue of Equation \eqref{jointindep}. \end{minipage}}
\end{figure}

The decomposition separates the regulariser into marginal association terms between the model predictions and each protected attribute, and a joint term $\eta$. The marginal terms penalise the dependence between the model predictions and each protected attribute individually, promoting fairness at the group level. The joint term $\eta$, on the other hand, penalises the dependence between the predictions and the joint distribution of the protected attributes, thereby addressing subgroup-level disparities.

\begin{remark}
    It is important to note that the regularisers can sometimes be negative by formulation \citep{chakraborty2019distance, szekely2023energy}. However, this negative value does not imply negative associations. In our experiments, we observed that with a sufficiently large sample size, any negative values were only marginally below zero. Importantly, including these slightly negative values in the objective function in Equation~\eqref{eq:objective} does not adversely affect the optimisation process, as they simply indicate that the association between the model predictions and the protected attributes is negligible, and therefore do not pose any issues during optimisation.
\end{remark} 

\section{Model performance metrics and calibration}\label{sec:modelcal_met}
With the objective function established, we now turn to the implementation and evaluation of our method. The proposed methodology applies to both regression and classification tasks, depending on the choice of loss function, and is compatible with any learning algorithm that optimises the specified objective function. In this paper, we implement it using a feedforward neural network due to its flexibility in accommodating custom loss components, in particular the inclusion of the regularisation term \citep{lecun2015deep, goodfellow2016deep, chollet2021deep}.

To optimise the objective function with a neural network, we apply gradient descent using the AdaHessian optimiser \citep{yao2021adahessian}. The justification for this choice is discussed in Appendix~\ref{apx:model_arch}, with full details of the network architecture and optimiser settings provided in Online Appendix B and C. In this section, we introduce the performance metrics used to evaluate model performance, beginning with accuracy metrics in Section~\ref{sec:acc_met}, followed by fairness metrics in Section~\ref{sec:fair_met}, and finally explaining how these metrics guide the calibration of regularisation strength in Section~\ref{sec:modelcal}.

\subsection{Accuracy metrics}\label{sec:acc_met}
We assess model accuracy using the Ranked Probability Score (RPS) \citep{murphy1970ranked}. Importantly, the choice of accuracy metric is flexible and should depend on the type of modelling task. We adopt RPS in this study because our applications involve binary classification (see Section~\ref{sec:COMPAS_app}) and Poisson regression (see Section~\ref{sec:pg15_app}), both of which return discrete probabilistic forecasts. For continuous probabilistic forecasts, the Continuous Ranked Probability Score (CRPS) is commonly used \citep{gneiting2007strictly}.

The RPS evaluates the accuracy of the predicted probability distribution by considering the entire cumulative distribution, rather than focusing solely on point estimates. A lower average RPS across predictions indicates better model performance as it reflects closer alignment between predicted and observed cumulative probabilities.

RPS is a strictly proper scoring rule, meaning that the expected score is maximised if and only if the predicted probability distribution matches the true underlying distribution \citep{gneiting2007strictly}. The definition of the RPS is provided in Appendix \ref{apx:performance_acc}.

\begin{remark}
    For a more comprehensive evaluation of model performance, we incorporate several additional metrics for accuracy in Section \ref{sec:appli}. We evaluate classification performance using accuracy (ACC) and measure predictive performance for the Poisson regression task using the Poisson deviance. Additionally, we conduct the Wilcoxon signed-rank test \citep{wilcoxon1992individual} to compare the accuracy between the regularised and unregularised models. Details of these complementary metrics are provided in Online Appendix D.
\end{remark}

\subsection{Fairness metrics}\label{sec:fair_met}
We introduce several fairness metrics used to evaluate our models. While all these metrics are useful for assessing the fairness of our model, not all are suitable for calibrating the regularisation strength. We discuss the strengths and limitations of each metric in this context here and justify our choice of calibration criterion in Section \ref{sec:modelcal}.

\subsubsection{Mean-based metrics}\label{sec:mean_metric}
To measure demographic disparities across subgroups, \citet{lindholm2024sensitivity} introduced the \textit{demographic unfairness metric} (UF), defined as:
\begin{equation} 0 \leq UF(\hat{\mathbf{Y}}) = \frac{\text{Var}(\mathbb{E}[\hat{\mathbf{Y}} \mid (\mathbf{S}_1, \ldots, \mathbf{S}_d)])}{\text{Var}(\hat{\mathbf{Y}})} \leq 1.
\end{equation}
Intuitively, $UF(\hat{\mathbf{Y}})$ quantifies the variation in prediction means across protected subgroups. A value of $UF(\hat{\mathbf{Y}}) = 0$ indicates identical subgroup means, while $UF(\hat{\mathbf{Y}}) = 1$ implies that predictions are constant within every protected subgroup and all variability is between subgroups. In particular, $UF(\hat{\mathbf{Y}}) = 0$ when demographic parity holds (as defined in Equation \eqref{dp:dist}), but the converse does not necessarily hold.

\subsubsection{Significance-based metrics}\label{sec:sig_metric}
Another approach to assessing fairness is through statistical hypothesis testing, which assesses whether model predictions are independent of the protected attributes. This aligns naturally with our framework, as dCov was originally developed as a test statistic for independence \citep{szekely2009brownian}. We consider two types of hypotheses.

First, when applying CCdCov, which penalises dependence between the model predictions and the joint distribution of the protected attributes, we can test for independence using the following hypothesis. Let $F$ denotes the CDF of the corresponding random vector:
\begin{equation}\label{eq:hyp_joint} 
\begin{aligned} 
&H_0:F_{\hat{\mathbf{Y}},(\mathbf{S}_1,\ldots,\mathbf{S}_d)} = F_{\hat{\mathbf{Y}}}F_{(\mathbf{S}_1,\ldots,\mathbf{S}_d)}\\ 
&H_A:F_{\hat{\mathbf{Y}},(\mathbf{S}_1,\ldots,\mathbf{S}_d)} \neq F_{\hat{\mathbf{Y}}}F_{(\mathbf{S}_1,\ldots,\mathbf{S}_d)},
\end{aligned} 
\end{equation}
where under the null hypothesis, the model predictions are independent of the joint distribution of the protected attributes.

Second, when applying JdCov, which penalises both pairwise and higher-order dependencies between the model predictions and all protected attributes, we can test for mutual independence:
\begin{equation}\label{eq:hyp_mutual} 
\begin{aligned} 
&H_0:F_{\hat{\mathbf{Y}},\mathbf{S}_1,\ldots,\mathbf{S}_d} = F_{\hat{\mathbf{Y}}}F_{\mathbf{S}_1}\cdots F_{\mathbf{S}_d}\\ 
&H_A:F_{\hat{\mathbf{Y}},\mathbf{S}_1,\ldots,\mathbf{S}_d} \neq F_{\hat{\mathbf{Y}}}F_{\mathbf{S}_1}\cdots F_{\mathbf{S}_d},
\end{aligned} 
\end{equation}
This is a stronger null hypothesis, as it requires mutual independence between the model predictions and all protected attributes, implying the absence of both pairwise and higher-order associations.

The hypothesis in \eqref{eq:hyp_joint} can be tested using the distance correlation $\chi^2$-test \citep{shen2022chi} or through a permutation test \citep{szekely2023energy}, while the hypothesis in \eqref{eq:hyp_mutual} can be evaluated using a JdCov-based permutation test proposed by \citet{chakraborty2019distance}. Details of both tests are provided in Appendix~\ref{apx:performance_fair}.

\subsubsection{Distribution-based metrics}\label{sec:JSD}
As discussed previously, our framework aims to promote statistical independence between the model prediction and the protected attribute subgroups. This objective is equivalent to seeking similarity in the prediction distributions across different subgroups. To quantify this, we use the Jensen–Shannon (JS) divergence, a symmetric and interpretable measure of distributional similarity introduced by \citet{lin1991divergence}.

The JS-divergence is computed as the weighted average of the Kullback–Leibler (KL) divergences ($D_{KL}$) between each subgroup’s conditional prediction distribution $\mathbb{P}(\hat{\mathbf{Y}} \mid (\mathbf{S}_1,\ldots,\mathbf{S}_d) = \mathbf{s})$ and the overall prediction distribution $\mathbb{P}(\hat{\mathbf{Y}})$: 
\begin{equation}
\mbox{\footnotesize
$D_{JS}=\displaystyle\sum_{\mathbf{s}} \pi_{\mathbf{s}} D_{KL}(\mathbb{P}(\hat{\mathbf{Y}}\mid (\mathbf{S}_1,\ldots,\mathbf{S}_d)=\mathbf{s})||\mathbb{P}(\hat{\mathbf{Y}}))$},
\end{equation}
where $\pi_{s}$ is the proportion of samples in each subgroup, and the KL-divergence between two probability distributions $P$ and $Q$ is defined as $D_{KL}(P||Q) = \sum\limits_{x} P(x)\log\frac{P(x)}{Q(x)}$, which quantifies how much information is lost when $Q$ is used to approximate $P$ \citep{kullback1951information}. For continuous protected attributes, the presence of infinitely many values leads to an uncountable number of conditional distributions, making direct computation of $D_{JS}$ challenging. As a practical remedy, we discretise the continuous attributes into intervals, enabling estimation of $D_{JS}$ over a finite set of subgroups. Importantly, since discretisation is a deterministic function of the original variables, independence between $\hat{\mathbf{Y}}$ and the original (continuous or joint) protected attributes implies independence from their discretised version. As a result, the JS-divergence computed from the discretised subgroups will also be zero in case of independence. A smaller $D_{JS}$ represents more aligned prediction distributions across subgroups, indicating a fairer model.

\subsection{Model calibration}\label{sec:modelcal}
Our model calibration process consists of two steps, followed by a final evaluation on a held-out test set. The full dataset is first split into a training set (80\%) and a test set (20\%). The test set is never used during model calibration and is kept aside for the final evaluation only.

First, we tune the hyperparameters of the neural network on the training set without regularisation ($\lambda=0$) to obtain a baseline model. Here, hyperparameters refer to model-architecture and training settings (e.g., number of layers/neurons per layer, dropout, learning rate, batch size). This is done using 5-fold cross-validation with Gaussian-process Bayesian optimisation and early stopping within each fold. The result is a single set of baseline hyperparameters, which are then fixed and used in all subsequent experiments. Further details are provided in Appendix~\ref{apx:model_arch}.

Next, we calibrate the regularisation strength $\lambda$ based on application-specific trade-offs between fairness and accuracy. For this step, the training set is further divided into a subtraining and a validation set. Models are trained on the subtraining set, with early stopping, and evaluated on the validation set for a range of $\lambda$ values. The choice of $\lambda$ can then be made according to fairness priorities and application-specific objectives, while using the same hyperparameters obtained from the first step.

Finally, after the hyperparameters and $\lambda$ have been chosen, the model is retrained on the full 80\% training set and evaluated once on the 20\% held-out test set.

In this section, we focus on how to calibrate the regularisation strength $\lambda$ in the objective function \eqref{eq:objective}, as choosing $\lambda$ carefully is crucial: a small $\lambda$ results in minimal fairness regularisation, while a large $\lambda$ may overemphasise the fairness penalty and lead to unnecessary sacrifices in predictive accuracy.

Although we use fairness metrics to guide the selection of $\lambda$, we concurrently track the accuracy metrics of the models. This ensures that the chosen value of $\lambda$ reflects a practical trade-off between fairness and accuracy. Selecting $\lambda$ based solely on fairness could lead to trivial models (e.g., intercept-only predictions), which may satisfy fairness constraints but offer no utility. Our approach aims to identify a model that improves fairness while retaining acceptable accuracy for the application at hand.

\subsubsection{Selecting fairness metrics for $\lambda$ calibration}
An important part of this process is selecting an appropriate fairness metric. We revisit the fairness metrics used to measure the fairness of model predictions introduced in Section~\ref{sec:fair_met} and assess their suitability for guiding the choice of regularisation strength.

\paragraph{\textbf{Mean-based metrics}} Although a lower $UF(\hat{\mathbf{Y}})$ indicates more similar subgroup means, it does not reflect differences in the full distribution of predictions. It is possible to observe $UF(\hat{\mathbf{Y}}) = 0$ even when subgroup distributions differ substantially. Since our framework promotes independence between predictions and protected attributes, we require a measure that captures differences in distributions rather than just means. We therefore do not use $UF(\hat{\mathbf{Y}})$ for calibration, although we report it in Section~\ref{sec:appli} for interpretability.

\paragraph{\textbf{Significance-based metrics}} While hypothesis tests can indicate whether predictions and protected attributes are statistically independent, they are not well suited for selecting $\lambda$. Associated $p$-values are highly sensitive to sample size: as sample size increases, even minor dependencies can become statistically significant \citep{sullivan2012using, lakens2022sample}, which may lead to excessive regularisation and reduced model accuracy.

In addition, the mutual-independence test in \eqref{eq:hyp_mutual}, which relies on the JdCov between the predictions and the protected attributes, becomes uninformative when the protected attributes themselves are dependent. In that situation mutual independence is violated by definition, so a significant $p$-value could reflect either this inherent dependence among the protected attributes or unfairness introduced by the model (see Section \ref{sec:jdreg}); at present we have no reliable way to disentangle these two sources.

\paragraph{\textbf{Distribution-based metrics}} By directly quantifying distributional similarity, JS-divergence (see Section~\ref{sec:JSD}) offers an intuitive and interpretable measure for assessing how closely subgroup predictions align with the overall distribution. In addition, excessively large values of $\lambda$ may distort the output distributions across protected subgroups when the fairness term dominates optimisation. Importantly, such distortions are not always evident from the size of the regulariser term itself: even when the regulariser appears small after training (e.g., after early stopping), the resulting subgroup distributions may still become dispersed. JS-divergence thus also serves as a valuable tool for detecting discrepancies caused by over-regularisation. Accordingly, we advocate its use as the primary criterion for calibrating regularisation strength.

\subsubsection{Procedure for tuning regularisation strength}
We use a holdout validation approach for its computational efficiency \citep{raschka2018model}. The training data $\mathcal{D}_{\text{train}}$ is split into a subtraining set $\mathcal{D}_{\text{subtrain}}$ (70\%) and a validation set $\mathcal{D}_{\text{valid}}$ (30\%). We train the model using various predetermined values of $\lambda$, including $\lambda = 0$ as a baseline, using $\mathcal{D}_{\text{subtrain}}$, and evaluate the resulting models on $\mathcal{D}_{\text{valid}}$ using the fairness and accuracy metrics discussed above. Full pseudocode is provided in Appendix~\ref{apx:lambdacal}.

The choice of $\lambda$ is fully exogenous and has to be made by the modeller. Here we explain and illustrate how to make such a choice, but the appropriate level of regularisation ultimately depends on the broader fairness objectives of the application, such as ethical standards or regulatory requirements. The calibration range for $\lambda$ should reflect the importance of fairness in the specific context. Moreover, because (joint) distance covariance is not scale-invariant under different datasets even when input features are standardised, the range of candidate $\lambda$ values will need to be selected so as to ensure that the regularisation term meaningfully contributes to the loss function. One possible starting point is to select $\lambda$ values with magnitudes comparable to the ratio of the baseline ($\lambda=0$) model’s validation loss to the regulariser (JdCov or CCdCov) computed using the predictions of the baseline model. Specifically, this ratio is given by 
\begin{equation}
    \left. {\frac{1}{n_\text{valid}}\sum_{i=1}^{n_\text{valid}}\mathcal{L}(\hat{\mathbf{y}}_{\Theta i},\mathbf{y}_i)} \right/ {\psi(\hat{\mathbf{y}},\mathbf{s}_1,\ldots\mathbf{s}_d)}
\end{equation} 
where $\hat{\mathbf{y}}$ denotes the model predictions on the validation set obtained using the model trained without regularisation.

\begin{remark}
    We acknowledge that tuning the neural network hyperparameters without regularisation ($\lambda=0$) may not yield optimal accuracy once regularisation is introduced ($\lambda>0$). Of course, one could adjust the hyperparameters for each value of $\lambda$. However, in this paper, we choose to fix the hyperparameters tuned under $\lambda = 0$ and apply the same configuration across all levels of regularisation when selecting a final $\lambda$.This setup allows us to isolate and clearly assess the effect of regularisation strength $\lambda$ on fairness outcomes under a consistent hyperparameter configuration. Moreover, re-tuning hyperparameters for each $\lambda$ would be computationally expensive. After selecting the preferred $\lambda$, we re-tune the hyperparameters with regularisation applied to the objective function for final model deployment; in our experiments, this re-tuning occurs immediately before evaluating the final model on the test set.
\end{remark}

\section{Applications}\label{sec:appli}
We apply our framework to two datasets that reflect different fairness challenges. The \texttt{COMPAS} dataset involves a \textit{classification} task with mixed type protected attributes (binary, categorical, continuous). These attributes are included in the predictive model input and are empirically associated based on the Kruskal-Wallis H test. Such dependence can lead to numerical instability during training when using the JdCov-regularised model, as discussed in Section~\ref{sec:jdreg}. In contrast, the motor insurance claims dataset involves a Poisson \textit{regression} task with empirically independent categorical protected attributes, which are excluded from the model input. Both applications demonstrate the flexibility of our method across tasks, dependency structures, and modelling contexts.

\subsection{Recidivism prediction with \texttt{COMPAS}}\label{sec:COMPAS_app}
\subsubsection{Background}
As introduced earlier, the Correctional Offender Management Profiling for Alternative Sanctions (\texttt{COMPAS}) tool is an assessment instrument developed by Northpointe to estimate a defendant’s likelihood of recidivism based on responses to a survey \citep{corbett_davies_algorithm_2016}. Although Northpointe asserts that \texttt{COMPAS} is race-neutral \citep{bornstein2021algorithms}, ProPublica’s analysis revealed that African-American defendants were significantly more likely to be classified as having a high likelihood of recidivism than Caucasian defendants. In particular, African-American males had the highest likelihood of being incorrectly classified as high risk, raising serious concerns about potential racial bias in the \texttt{COMPAS} risk assessment \citep{angwin2016machine}. 

Additionally, recent fairness concerns in the legal system have arisen over the use of algorithmic recidivism assessment tools in guiding judgements. Written warnings now explicitly acknowledge that \texttt{COMPAS} may ``disproportionately classify minority offenders as having a higher risk of recidivism" \citep{harvard2017loomis}. These concerns motivate us to mitigate unfairness in the predicted recidivism rates, while still maintaining adequate accuracy to ensure the predictions remain informative for guiding legal judgements.

\subsubsection{Data description}
We follow the same pre-processing steps as in the analysis by \citet{angwin2016machine}. Therein, \texttt{Ethnicity} was classified into 6 categories (African-American, Caucasian, Hispanic, Asian, Native American, Other). We merged the Asian and Native American groups into the ``Other" category, as these groups account for less than 1\% of the whole sample. Appendix \ref{apx:compas_prep} lists out the variables used and the pre-processing done in the analysis. We consider the following protected attributes: \texttt{Female} (binary), \texttt{Ethnicity} (categorical with four classes), and \texttt{Age} (continuous). Compared to the motivating example in Section~\ref{sec:motive}, only \texttt{Age} is newly included. Even though one may not want to consider age as a protected feature in this particular context, we do so as to illustrate how a continuous attribute can be included in our regulariser, and to demonstrate the ability of our proposed regulariser to handle disparities across mixed protected attribute types. In addition, we apply one-hot encoding to \texttt{Ethnicity} to show that our regulariser can operate on multidimensional random vectors.

\begin{figure}[h!]
    \centering
    \caption{Empirical average recidivism rate by protected attributes \label{fig:datadesc}}
    \includegraphics[width=1\linewidth]{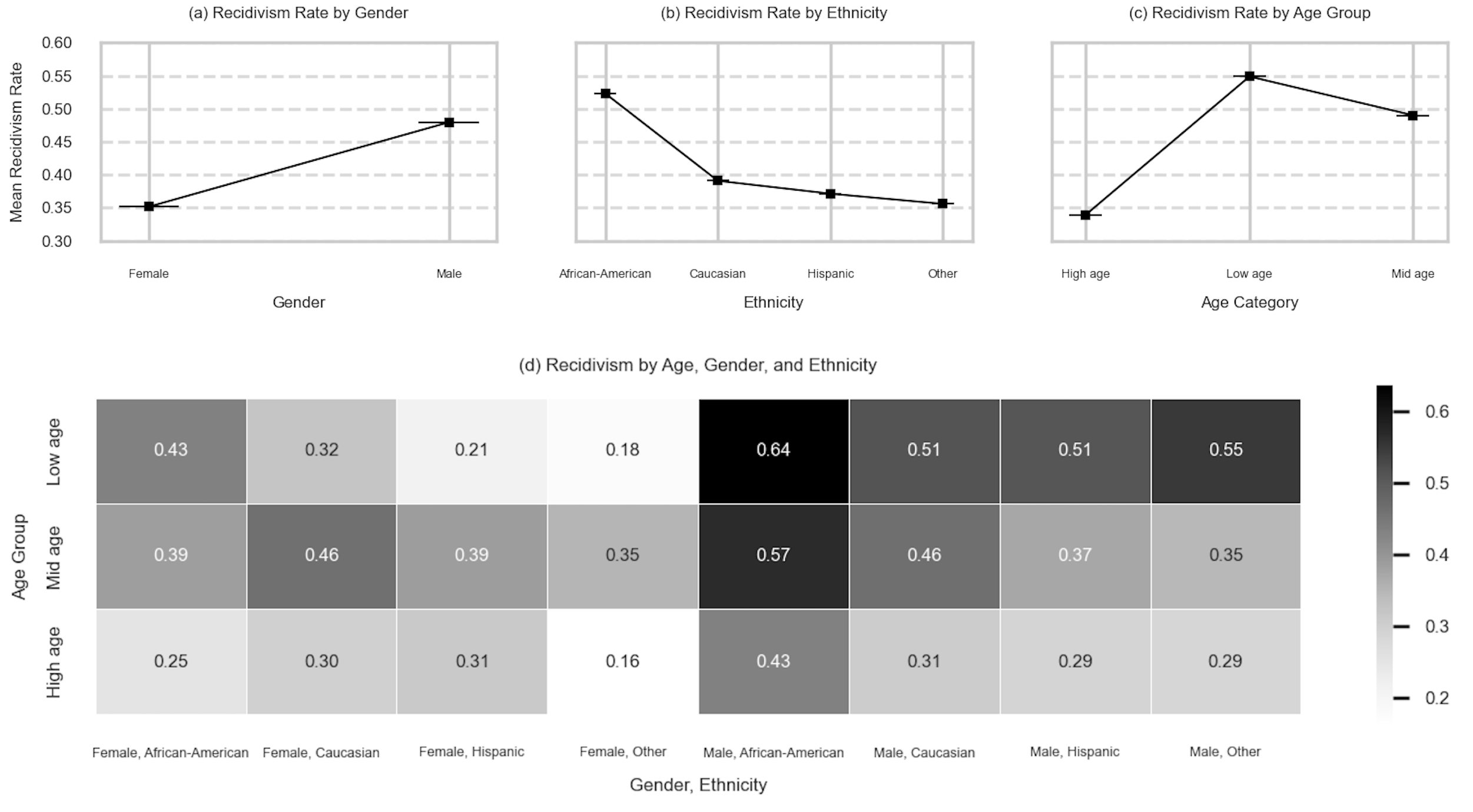}
    \caption*{\begin{minipage}{0.95\textwidth}     \textit{Note}: This figure reports exploratory data analysis of average recidivism rates across protected attributes in the \texttt{COMPAS} dataset. Panels (a)–(c) show average recidivism by gender, race, and age group, respectively. Panel (d) presents subgroup averages defined by the interaction of all three protected attributes. The results highlight disparities across individual attributes as well as their intersections, with notably higher rates for younger African-American males.\end{minipage}}
\end{figure}

Figure~\ref{fig:datadesc} displays the average recidivism rates across the protected attributes. On average, males and African-Americans exhibit the highest recidivism rates. This aligns with the unregularised model predictions in Figure~\hyperref[fig:compare_methods]{\ref*{fig:compare_methods}.1} and \hyperref[fig:compare_methods]{\ref*{fig:compare_methods}.3}.

We next consider interactions between protected attributes, as they may reveal hidden sources of bias not apparent when attributes are examined individually. Figure~\hyperref[fig:datadesc]{\ref*{fig:datadesc}(d)} shows the average recidivism rates across subgroups defined by the combination of gender, race, and age. For the purpose of this exploratory analysis, age is categorised into three groups based on the 33rd and 67th percentiles of the overall age distribution: 0–27, 28–37, and 38 or above. Notably, African-American males under 27 years of age exhibit the highest average recidivism rates. Without fairness constraints, the model may learn and reproduce this data pattern, leading to subgroup disparities in predictions; that is, higher predicted recidivism rates for some groups.
\clearpage

These observed disparities highlight the potential for bias in machine learning models trained on \texttt{COMPAS} data. In addition, we applied the Kruskal-Wallis H test by \citet{kruskal1952use} to compare median age across ethnicities, and obtained a $p$-value of $6.3435 \times 10^{-46}$,  indicating a statistically significant difference in median age across ethnicities. This finding suggests a strong association between age and ethnicity. Given this association, we may encounter numerical instability during model training with the JdCov regulariser, as discussed in Section~\ref{sec:jdreg}.

\subsubsection{Model specification and regularisation calibration}
We fit a binary classification task using the objective function:
\begin{equation}\label{eq:COMPAS_obj}
\mbox{\small$
    \begin{aligned}
        \min_\Theta\Biggl\{&-\frac{1}{n}\sum^n_{i=1}[\mathbf{y}_i\log(\hat{\mathbf{y}}_{\Theta i}) + (1-\mathbf{y}_i)\log(1-\hat{\mathbf{y}}_{\Theta i})]+\lambda\cdot\psi\left(\hat{\mathbf{y}},\texttt{\textbf{Female}},\texttt{\textbf{Ethnicity}}, \texttt{\textbf{Age}}\right)\Biggr\},
    \end{aligned}$
}
\end{equation}
where $\mathbf{y}_i\in\{0,1\}$ represents whether an individual re-offended ($\mathbf{y}_i = 1$) within two years after conviction, and where $\hat{\mathbf{y}}_{\Theta i}$ denotes the predicted recidivism probability, with protected attributes included among the covariates used as model input (see Table \ref{tab:details_COMPAS} in Appendix~\ref{apx:compas} for the full list of covariates).

\begin{figure}[ht]
    \centering
    \caption{JS-divergence vs regularisation strength and accuracy on validation set (\texttt{COMPAS}) \label{fig:calibration_diagnostics}}
    \includegraphics[width=1
    \linewidth]{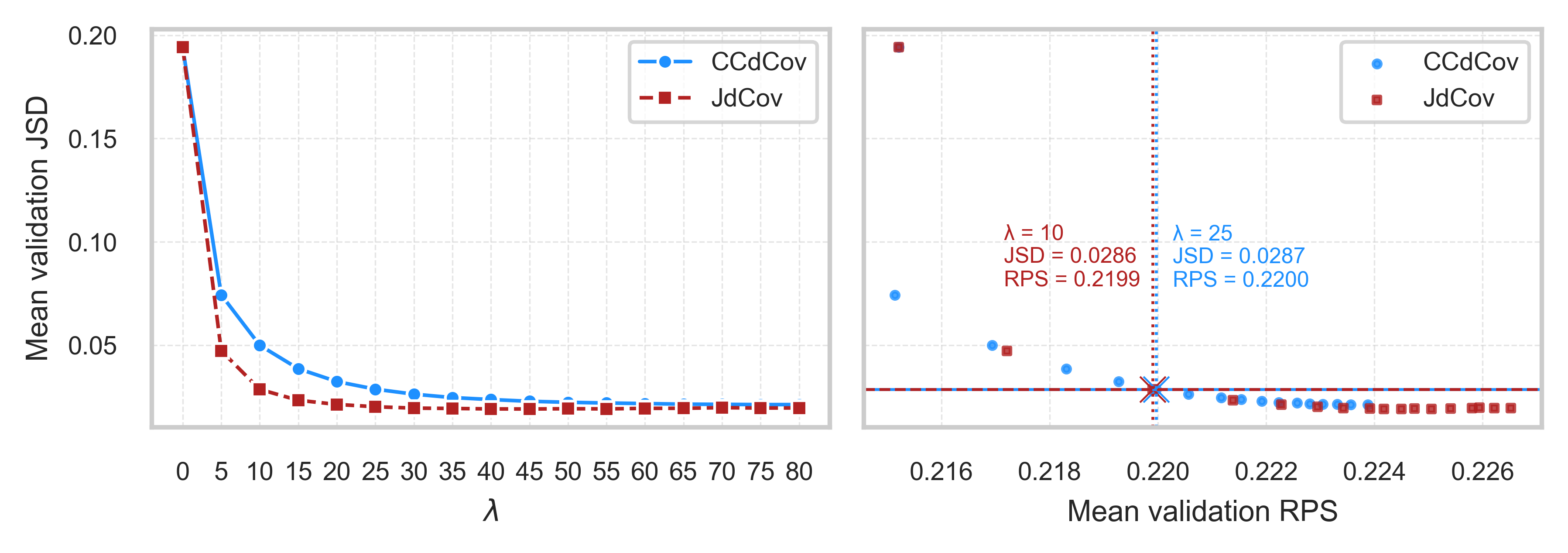}
    \caption*{     \begin{minipage}{0.95\textwidth}     \textit{Note}: This figure presents the mean validation JS-divergence (JSD) against $\lambda$ (left) and against mean validation RPS (right) for two regularisation methods: concatenated distance covariance (CCdCov, light blue circles) and joint distance covariance (JdCov, dark red squares). Each point in the right panel corresponds to the same model run shown in the left panel at a given value of $\lambda$; since both panels are based on the same runs, they share the same JSD values. Thus, the left panel shows how JSD varies with $\lambda$, while the right panel re-plots those identical JSD values against the corresponding validation RPS to visualise the fairness–accuracy trade-off. In both panels, the vertical axis reports mean validation JSD. The model was trained on the \texttt{COMPAS} dataset with 10 independent seeds. As described in Section \ref{sec:modelcal}, age was treated as a continuous attribute during training but discretised into three equally sized bins (based on the 33rd and 67th percentiles) to enable the computation of JS-divergence. The results illustrate that stronger regularisation improves fairness (lower JSD) but highlights the trade-off with accuracy, and that for the same level of accuracy, JdCov and CCdCov can yield different fairness outcomes, as seen in the right panel.\end{minipage}}
\end{figure}

\begin{table}[ht]
    \centering
    \caption{Performance metrics of a regularised binary classifier evaluated on validation set (\texttt{COMPAS}) \label{tab:fairness}}
    {\resizebox{\textwidth}{!}{
\begin{tabular}{ccccccccccc}
\toprule
 & \multicolumn{5}{c}{(a) CCdCov} & \multicolumn{5}{c}{(b) JdCov} \\
\cmidrule(lr){2-6} \cmidrule(lr){7-11}
$\lambda$ & RPS & CCdCov & JdCov & JS-divergence & $UF(\hat{\mathbf{Y}})$ & RPS & CCdCov & JdCov & JS-divergence & $UF(\hat{\mathbf{Y}})$ \\
\midrule
0   & 0.2152 & 0.0133 & 0.0236 & 0.1942 & 0.3439 & 0.2152 & 0.0133 & 0.0236 & 0.1942 & 0.3439 \\
10  & 0.2169 & 0.0008 & 0.0071 & 0.0499 & 0.2226 & 0.2199 & 0.0006 & 0.0063 & 0.0286 & 0.1323 \\
20  & 0.2193 & 0.0005 & 0.0063 & 0.0324 & 0.1664 & 0.2223 & 0.0004 & 0.0060 & 0.0213 & 0.0957 \\
40  & 0.2215 & 0.0004 & 0.0061 & 0.0237 & 0.1195 & 0.2242 & 0.0005 & 0.0060 & 0.0191 & 0.0772 \\
60  & 0.2228 & 0.0004 & 0.0060 & 0.0217 & 0.0986 & 0.2254 & 0.0005 & 0.0061 & 0.0194 & 0.0719 \\
80  & 0.2239 & 0.0004 & 0.0060 & 0.0212 & 0.0880 & 0.2265 & 0.0005 & 0.0061 & 0.0196 & 0.0708 \\

\bottomrule
\end{tabular}}}
    \vspace{0.2cm}     \caption*{     \begin{minipage}{0.95\textwidth}     \textit{Note}: This table reports the mean validation metrics, averaged over 10 independent model initialisations. The fairness-regularised models are trained on the \texttt{COMPAS} dataset using the objective function defined in Equation~\eqref{eq:COMPAS_obj}. Results are presented for two regularisers: (a) distance covariance with concatenated protected attributes (CCdCov) and (b) joint distance covariance (JdCov). For each $\lambda \in \{0, 10, 20, 40, 60, 80\}$, we report the predictive accuracy using the Ranked Probability Score (RPS), along with the fairness metrics: CCdCov, JdCov, JS-divergence, and the unfairness measure $UF(\hat{\mathbf{Y}})$.\end{minipage}}
\end{table}

We calibrate our regularisation strength $\lambda$ using the JS-divergence, as motivated in Section~\ref{sec:modelcal}. We set the predetermined range of $\lambda$ from 0 to 80 in increments of 5. Figure~\ref{fig:calibration_diagnostics} shows, for each value of $\lambda$ and averaged over 10 runs with different random seeds, the mean validation JS-divergence and its trade-off with validation RPS under JdCov and CCdCov regularisation. The left panel illustrates the effect of increasing $\lambda$ on JS-divergence, while the right panel shows the corresponding trade-off with validation RPS. Although the binary classifier was trained with age treated as a continuous attribute, for the purpose of computing JS-divergence, we binned age using the 33rd and 67th percentiles.

Table~\ref{tab:fairness} complements Figure~\ref{fig:calibration_diagnostics} by reporting the exact validation metrics across regularisation strengths, including RPS, JS-divergence, the UF measure, and the CCdCov and JdCov values. Consistent with the figure, we observe improved fairness and reduced model accuracy as $\lambda$ increases, reflected by decreasing fairness metrics and increasing RPS. At higher regularisation levels, however, CCdCov exhibits more stable performance across fairness metrics, whereas the fairness metrics for JdCov lose monotonicity.

As seen in the left panel of Figure \ref{fig:calibration_diagnostics}, both regularisers yield a sharp reduction in JS-divergence at $\lambda = 5$ already, with diminishing returns beyond this point. While JdCov generally yields lower JS-divergence, a direct comparison between the two regularisers at the same value of $\lambda$ may not be meaningful, as the effect of regularisation strength on accuracy will generally differ across regularisers. Furthermore, as $\lambda$ increases beyond $\lambda = 40$, the JS-divergence loses its monotonicity, as confirmed by the detailed values in Table~\ref{tab:fairness}. This instability arises due to the association between ethnicity and age, as discussed in Section \ref{sec:regframework}.

The right panel of Figure~\ref{fig:calibration_diagnostics} plots the mean validation JS-divergence against the mean validation RPS across a range of $\lambda$ values. Each point illustrates the trade-off between fairness and accuracy at a given level of regularisation strength. The models using JdCov at $\lambda = 10$ and CCdCov at $\lambda = 25$ yield nearly identical JS-divergence, yet JdCov achieves a lower RPS, indicating better accuracy at the same level of fairness. Similar patterns are observed at other $\lambda$ values where the two regularisers attain comparable JS-divergence. These results suggest that, for the \texttt{COMPAS} dataset, JdCov is a more efficient regulariser at lower levels of $\lambda$. However, at higher levels of $\lambda$, CCdCov may be preferred due to numerical instability in JdCov.

Given the absence of formal fairness requirements in the \texttt{COMPAS} context, we select $\lambda = 10$ for JdCov and $\lambda = 25$ for CCdCov to impose mild regularisation that is comparable across both regularisers, and that strikes a balance between fairness and performance. Again, this choice is purely arbitrary here, and could be adjusted according to exogenous managerial needs.

\begin{remark}
    When using CCdCov as the regulariser, we observe that its value on the validation set does not decrease monotonically as $\lambda$ increases from 20 to 80. However, this behaviour is primarily numerical. In theory, CCdCov can attain zero, and increasing $\lambda$ should reduce the regularisation objective on the training set. In addition, the lack of monotonicity on the validation set may be attributed to generalisation effects, as the regulariser is not directly optimised on the validation data. This interpretation is supported by the consistently decreasing JS-divergence and UF on the validation set (see Figure \ref{fig:calibration_diagnostics} and Table \ref{tab:fairness}), which reflect improved fairness as $\lambda$ increases.
\end{remark}

\begin{remark}
    Note that while the choice of regularisation strength is up to the user, the right hand side of Figure \ref{fig:calibration_diagnostics} teaches us that there are sub-optimal such choices, in the Pareto sense. Our choice of $\lambda$s shows a pair of points with identical JS-divergence whereby JdCov is Pareto optimal (lower RPS). On the other hand, for given RPS around 0.2155, CCdCov displays a much lower JS-divergence, as illustrated by the two leftmost points in the right-hand panel of Figure \ref{fig:calibration_diagnostics}.
\end{remark}

\subsubsection{Test set results and analysis}
Setting $\lambda$ as discussed in the previous section, we re-tune the model hyperparameters and re-train each model on the full training dataset. The details of the hyperparameters are provided in Appendix \ref{apx:compas}. We now evaluate the performance of this final model on the held-out test set, as outlined in Section \ref{sec:modelcal}. 

\begin{figure}[h!]
    \centering
    \caption{Average predicted recidivism rates on test set under different regularisers (\texttt{COMPAS}) \label{fig:compasresult}}
    \includegraphics[width=1\linewidth]{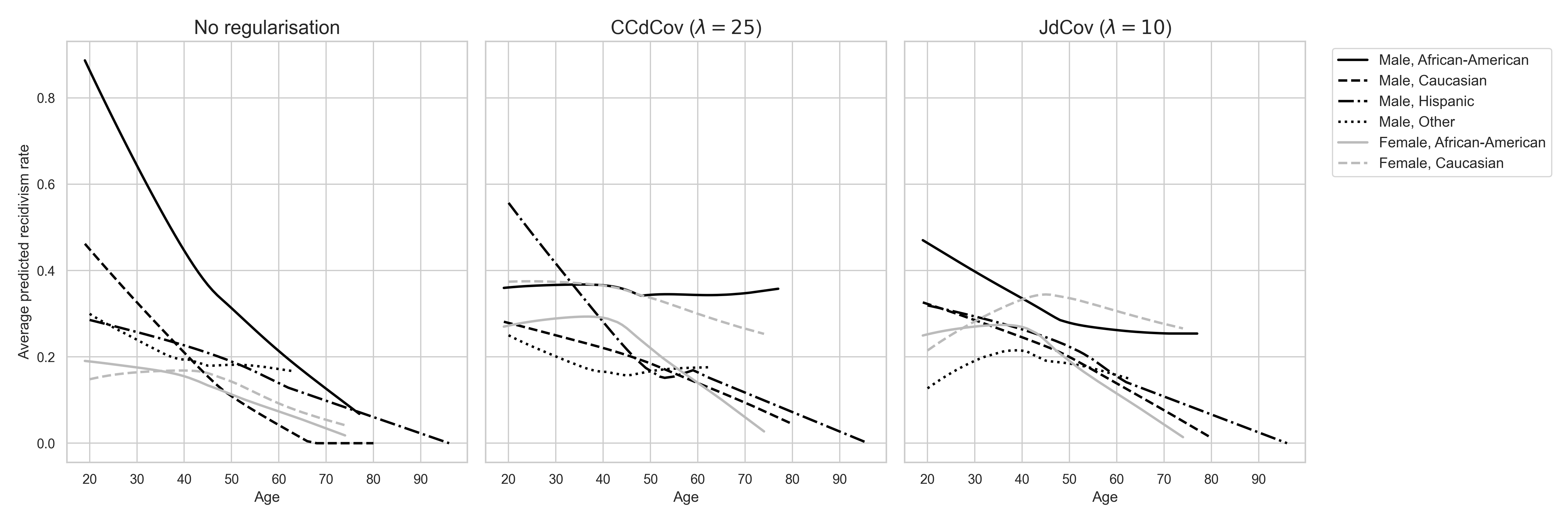}
    \caption*{     \begin{minipage}{0.95\textwidth}\textit{Note}: Average predicted recidivism rates of the test set by age using lowess smoothing (span = 1), split by gender–ethnicity subgroups with at least 100 training samples. Results are shown for three models: (left) no regularisation, (middle) CCdCov regularisation with $\lambda=25$, and (right) JdCov regularisation with $\lambda=10$. Regularisation yields flatter and closer prediction curves across age.\end{minipage}}
\end{figure}

Recall that our objective is to reduce disparities in predicted recidivism probabilities across protected subgroups while maintaining an adequate level of accuracy. Accordingly, we expect the average predicted recidivism rates to become more uniform across age groups and better aligned across subgroups, albeit with a slight trade-off in accuracy. Figure~\ref{fig:compasresult} displays the average predicted recidivism rates across subgroups defined by the intersections of \texttt{Gender}, \texttt{Ethnicity}, and \texttt{Age}, based on predictions from the test set. Without regularisation (left panel), predictions vary substantially across subgroups. Most notably, for African-American males, the average predicted rates decline sharply with age. Similar decreasing patterns are observed in other subgroups.

In contrast, both regularised models (middle and right panels) produce flatter and more closely aligned subgroup predictions across age compared to the unregularised baseline (left panel), suggesting that regularisation clearly improves fairness by reducing disparities associated with age–subgroup interactions. As shown in Figure~\ref{fig:compasresult}, this improvement is visible on the test set, though the curves remain somewhat variable due to limited subgroup sample sizes. The same analysis on the training set (Figure~\ref{fig:compasresult_train}) shows even clearer flattening and alignment, reinforcing that both regularisers reduce subgroup disparities relative to the baseline. This aligns with Table~\ref{tab:test_acc}, where both models achieve similar JS-divergence values, reflecting the intentional selection of $\lambda$ to match validation JS-divergence. Notably, the model using CCdCov achieves lower JS-divergence and UF in Table~\ref{tab:test_acc}, indicating improved fairness with this particular dataset and model.

Nonetheless, some disparities persist among very young and older individuals in certain subgroups. This is likely due to the limited size of the \texttt{COMPAS} training set, which contains approximately 5,000 observations. When the data are partitioned into fine-grained subgroups by age, gender, and ethnicity, many groups contain relatively few samples, limiting the effectiveness of regularisation. We discuss this issue further in Appendix~\ref{apx:compas_discussion}, where we explore resampling-based strategies for improving fairness in underrepresented subgroups.

We also applied the statistical tests described in Section~\ref{sec:fair_met} to formally evaluate whether the model predictions remain statistically dependent on the protected attributes. All resulting $p$-values were close to zero, rejecting the null hypothesis of independence. This outcome is expected, as our selected values of $\lambda$ were not calibrated to enforce complete independence, but rather to reduce disparities while maintaining predictive accuracy.

\begin{table}[ht]
    \centering
    \caption{Performance metrics of a regularised binary classifier evaluated on test set (\texttt{COMPAS}) \label{tab:test_acc}}
    {{
\begin{tabular}{lcccccc}
\toprule
 & \multicolumn{2}{c}{(a) Accuracy metrics} & \multicolumn{4}{c}{(b) Fairness metrics} \\
\cmidrule(lr){2-3} \cmidrule(lr){4-7}
Model & RPS & ACC & CCdCov & JdCov & JS-divergence & $UF(\hat{\mathbf{Y}})$ \\
\midrule
No regularisation & 0.2128 & 0.6628 & 0.0113 & 0.0183 & 0.1879 & 0.2503 \\
CCdCov            & 0.2241 & 0.6317 & 0.0010 & 0.0043 & 0.0302 & 0.0907 \\
JdCov             & 0.2220 & 0.6464 & 0.0008 & 0.0040 & 0.0301 & 0.1049 \\
\bottomrule
\end{tabular}
    }}
    \vspace{0.2cm}     \caption*{     \begin{minipage}{0.95\textwidth}     \textit{Note}: Test performance metrics for three binary classifiers trained on the \texttt{COMPAS} dataset with different regularisers, evaluated at $\lambda=25$ for CCdCov and $\lambda = 10$ for JdCov. Panel (a) reports accuracy metrics (RPS and ACC), while Panel (b) presents fairness metrics: CCdCov, JdCov, JS-divergence, and the unfairness measure $UF(\hat{\mathbf{Y}})$.\end{minipage}}
\end{table}

We now examine how different fairness regularisers affect model accuracy. As shown in Table~\ref{tab:test_acc}, both regularised models exhibit lower ACC and higher RPS values compared to the unregularised baseline, reflecting the expected trade-off between fairness and predictive performance. The $p$-values from the Wilcoxon signed-rank tests (discussed in Section \ref{sec:acc_met}) are close to zero, indicating statistically significant increases in RPS. Specifically, the model using CCdCov demonstrates lower accuracy than the model with JdCov, as reflected in its higher RPS and lower ACC in Table~\ref{tab:test_acc}. These results are consistent with the effects of regularisation, where the fairness penalty leads to moderate reductions in accuracy, and stronger fairness will comes at the cost of predictive performance.

To evaluate model generalisation and check for potential overfitting, we reproduce Figure~\ref{fig:compasresult} and Table~\ref{tab:test_acc} using the training set, rather than the test set used in the main text. This comparison allows us to examine whether the fairness–accuracy patterns observed on the test split also hold on the training data. The corresponding results are shown in Figure~\ref{fig:compasresult_train} and Table~\ref{tab:train_acc} in Appendix~\ref{apx:COMPAS_train}. Both figures exhibit similar patterns across different regularisers, with predicted recidivism rates that are more closely aligned and generally flatter than those of the unregularised model. The slight downward trend in predictions may be explained by the limited sample sizes in older age groups within each gender–ethnicity subgroup. In total, only {71} individuals in the training set are older than {65} years. The most notable discrepancy appears in the ``Male, Other" subgroup, which contains just {5} individuals over age {65}. In terms of performance metrics, we observe a modest reduction in accuracy on the test set, reflected in a higher RPS and lower ACC compared to the training set. At the same time, the model achieves better fairness on the test set, as indicated by lower fairness metric values. Taken together, these results suggest that the model generalises well and does not overfit to the training data.

\subsection{Motor insurance claims dataset}\label{sec:pg15_app}
\subsubsection{Background}
We now apply our framework to a Poisson regression model using the \texttt{pg15training} dataset. This dataset is part of the \texttt{CASdatasets} package in R \citep{dutang2020package}, which provides a broad collection of insurance data.

\subsubsection{Data description}
The dataset consists of 100,000 policies covering up to one year of exposure between 2009 and 2010. Each record corresponds to an individual policyholder and includes attributes describing both the insured and the vehicle. We model the number of third-party property damage claims using Poisson regression, adjusting for exposure via the variable \texttt{Expo}.

We consider two protected attributes in our analysis: \texttt{Female}, a binary indicator of the policyholder's gender, and \texttt{Region}, a categorical variable with ten regions representing the policyholder's residence. \texttt{Region} is treated as a protected attribute because a policyholder's residential area may serve as a proxy for race or socioeconomic status \citep{avraham2013understanding}. Notably, the two protected attributes appear to be independent. This is illustrated from Figure~\ref{fig:pg15eda2} in Appendix~\ref{apx:pg15}, where the proportion of male policyholders remains relatively consistent across regions. To formally assess this, we conducted a $\chi^2$ test of independence between gender and region. The resulting $p$-value of 0.59652 indicates no statistically significant association, supporting the conclusion that gender and region are independent in this dataset. A detailed description of the pre-processing steps and the full set of input variables is provided in Appendix~\ref{apx:pg15}.

Figure \ref{fig:pg15eda} in Appendix~\ref{apx:pg15} displays average claim frequencies by gender and region. Males have higher average number of claims across all regions, and Region R exhibits the highest claim frequency overall. These disparities in the data could persist in the model predictions if no fairness interventions were applied.

\subsubsection{Model specification and regularisation calibration}
We fitted a Poisson regression model on the claim frequency of the portfolio with a neural network subject to the following loss function:
\begin{equation}\label{eq:pg15_obj}
\mbox{\small$
\begin{aligned}
\min_\Theta\Biggl\{&\frac{1}{n}\sum\left[\texttt{Expo}_i\hat{\mathbf{y}}_{\Theta i}-\mathbf{y}_i\log(\hat{\mathbf{y}}_i)\right]+\lambda\cdot\psi\left(\hat{\mathbf{y}},\texttt{\textbf{Female}},\texttt{\textbf{Region}}\right)
\Biggr\},\end{aligned}$
}
\end{equation}
with $\mathbf{y}_i$ being the number of claims recorded for the $i$-th policyholder, and $\hat{\mathbf{y}}_{\Theta i}$ representing the predicted claims frequency for the $i$-th policyholder for a unit of exposure. Note that the protected attributes \texttt{Female} and \texttt{Region} are excluded from the model inputs to prevent \textit{direct} discrimination, which may be prohibited in certain jurisdictions, such as those in the EU \citep{lindholm2022discrimination, eu2004}. A key advantage of our proposed methodology is its ability to construct a predictor that excludes protected attributes as inputs, while still addressing potential dependencies arising from proxy variables during training. We set the predetermined set of $\lambda$ between $0$ to $150$ with an increment of $10$.

\begin{figure}
    \centering
    \caption{JS-divergence vs regularisation strength and accuracy on validation set (\texttt{pg15training}) \label{fig:calibration_diagnostics_pg15}}
    \includegraphics[width = 1\linewidth]{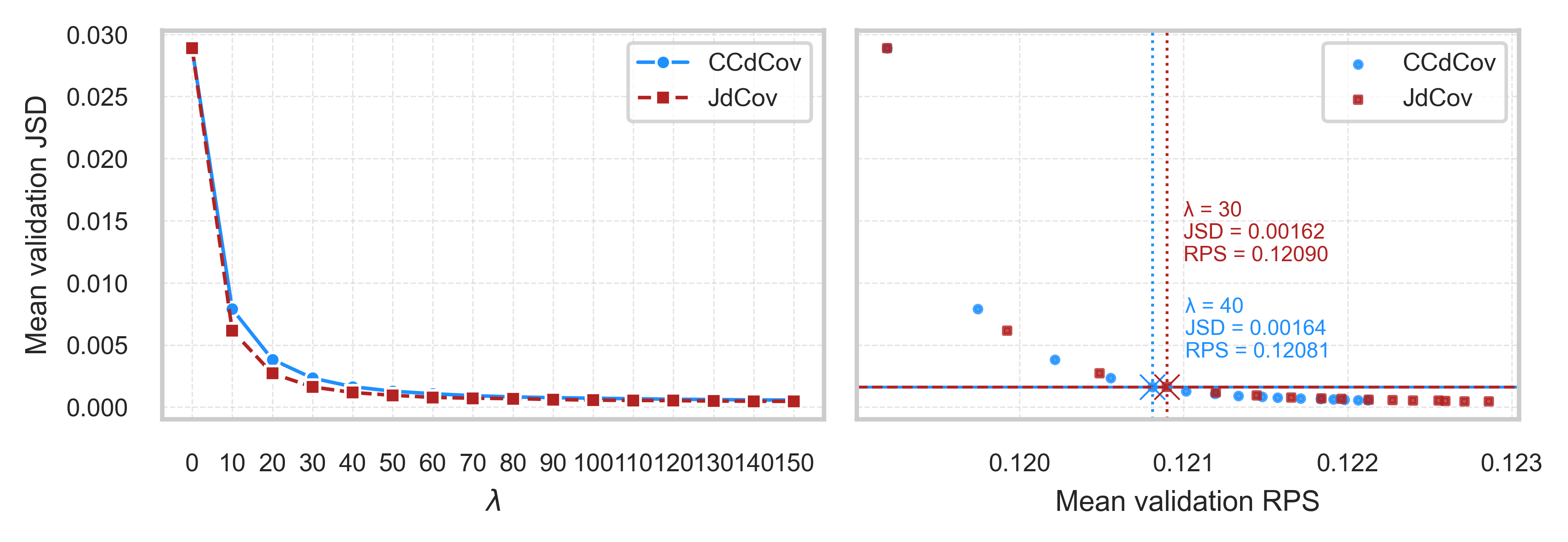}
    \caption*{     \begin{minipage}{0.95\textwidth}     \textit{Note}: This figure illustrates the mean validation JS-divergence (JSD) against $\lambda$ (left) and against mean validation RPS (right) for the Poisson regression model trained on the \texttt{pg15training} dataset under two regularisers: CCdCov (light blue circles) and JdCov (dark red squares), averaged across 10 random seeds. Each point in the right panel corresponds to the same model run shown in the left panel at a given value of $\lambda$; since both panels are based on the same runs, they share the same JSD values. Thus, the left panel shows how JSD varies with $\lambda$, while the right panel re-plots those identical JSD values against the corresponding RPS to visualise the fairness–accuracy trade-off. The results show that increasing regularisation strength reduces JSD, indicating improved fairness, while also demonstrating the trade-off with accuracy. At comparable levels of fairness, CCdCov consistently achieves better accuracy than JdCov on this dataset.\end{minipage}}
\end{figure}

\begin{table}[ht]
    \centering
    \caption{Performance metrics of a regularised Poisson regressor evaluated on validation set (\texttt{pg15training}) \label{tab:fairness_pg15}}
    {\resizebox{\textwidth}{!}{
\begin{tabular}{ccccccccccc}
\toprule
 & \multicolumn{5}{c}{(a) CCdCov} & \multicolumn{5}{c}{(b) JdCov} \\
\cmidrule(lr){2-6} \cmidrule(lr){7-11}
{$\lambda$} & RPS & CCdCov & JdCov & JS-divergence & $UF(\hat{\mathbf{Y}})$ & RPS & CCdCov & JdCov & JS-divergence & $UF(\hat{\mathbf{Y}})$ \\
\midrule
0   & 0.1192 & 9.1380e-04 & 1.4778e-03 & 0.0289 & 0.0774 & 0.1192 & 9.1380e-04 & 1.4778e-03 & 0.0289 & 0.0774 \\
10  & 0.1197 & 1.9653e-04 & 5.4627e-04 & 0.0079 & 0.0203 & 0.1199 & 1.5219e-04 & 4.8335e-04 & 0.0061 & 0.0159 \\
20  & 0.1202 & 9.4145e-05 & 4.1285e-04 & 0.0038 & 0.0100 & 0.1205 & 6.9255e-05 & 3.8005e-04 & 0.0027 & 0.0074 \\
40  & 0.1208 & 4.4518e-05 & 3.4792e-04 & 0.0016 & 0.0047 & 0.1212 & 3.3815e-05 & 3.3314e-04 & 0.0012 & 0.0036 \\
80  & 0.1214 & 2.5133e-05 & 3.2185e-04 & 0.0008 & 0.0026 & 0.1220 & 2.1168e-05 & 3.1529e-04 & 0.0007 & 0.0024 \\
120 & 0.1219 & 2.0275e-05 & 3.1479e-04 & 0.0006 & 0.0022 & 0.1226 & 1.7017e-05 & 3.0845e-04 & 0.0005 & 0.0022 \\
150 & 0.1221 & 1.8772e-05 & 3.1228e-04 & 0.0006 & 0.0021 & 0.1229 & 1.4751e-05 & 3.0393e-04 & 0.0005 & 0.0021 \\
\bottomrule
\end{tabular}
    }}
    \vspace{0.2cm}     \caption*{     \begin{minipage}{0.95\textwidth}     \textit{Note}: This table reports mean validation metrics based on 10 independent runs of a fairness-regularised Poisson regression model trained on the \texttt{pg15training} dataset. The models are trained using the objective function defined in Equation~\eqref{eq:pg15_obj}. Results are presented for two regularisers: CCdCov (Panel (a)) and JdCov (Panel (b)). For each $\lambda \in \{0, 10, 20, 40, 80, 120, 150\}$, we report the validation Ranked Probability Score (RPS), CCdCov, JdCov, JS-divergence, and the unfairness measure $UF(\hat{\mathbf{Y}})$.
\end{minipage}}
\end{table}

Figure \ref{fig:calibration_diagnostics_pg15} displays the JS-divergence and RPS for each level of $\lambda$. Table \ref{tab:fairness_pg15} reports the corresponding fairness metrics for each setting of $\lambda$. In this scenario, increasing $\lambda$ is unlikely to result in numerical instability under JdCov, as the protected attributes (region and gender) do not appear to be strongly associated as discussed earlier in this section. Under such conditions, the higher-order associations can, in theory, be reduced or mitigated by adjusting the model predictions during optimisation, as discussed in Section \ref{sec:jdreg}.

We again compare the RPS of the two models at similar levels of JS-divergence. The right panel of Figure~\ref{fig:calibration_diagnostics_pg15} displays the mean validation JS-divergence and RPS for each regulariser across different $\lambda$ values. In contrast to the \texttt{COMPAS} application, CCdCov consistently achieves lower RPS values than JdCov at comparable levels of JS-divergence. For example, comparing JdCov at $\lambda = 30$ and CCdCov at $\lambda = 40$, the latter yields a lower RPS, indicating more accurate predictions. A similar pattern is observed for other $\lambda$ pairs where the two models attain comparable fairness levels. These results indicate that, for this dataset, CCdCov outperforms JdCov in terms of accuracy at a given level of fairness, suggesting that CCdCov is the more efficient regulariser in this context.

\subsubsection{Results and analysis}
To ensure regulatory compliance on non-discriminatory pricing, we select $\lambda = 30$ for JdCov and $\lambda = 40$ for CCdCov as our final model, adopting stronger regularisation to provide additional assurance of fairness in deployment.

\begin{figure}[h]
    \centering
    \caption{Kernel density estimates and empirical CDF of predicted claim frequency on test set (\texttt{pg15training}) \label{fig:pg15_results}}
    \includegraphics[width=1\linewidth]{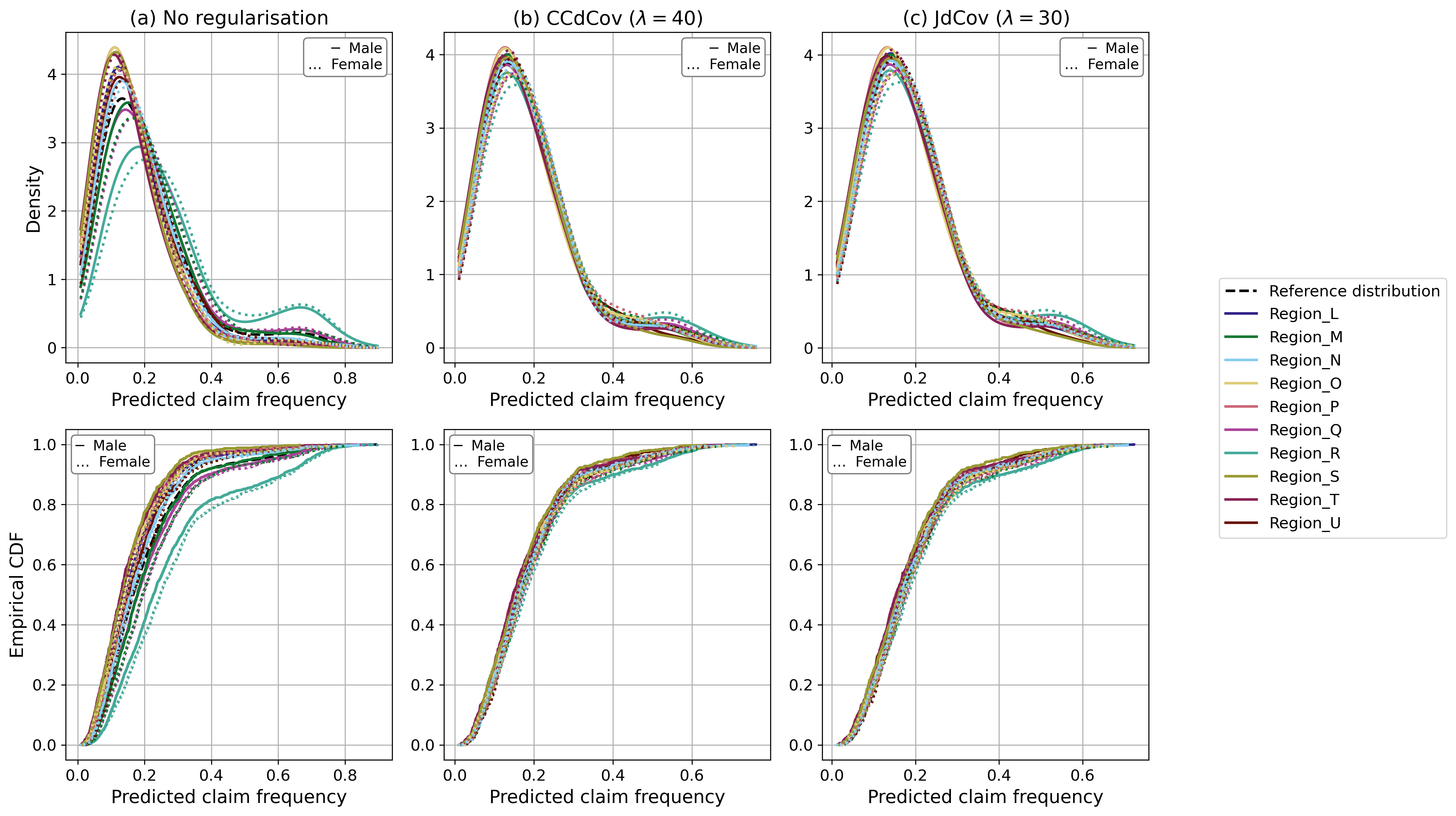}
    
    \vspace{0.2cm}     \caption*{     \begin{minipage}{0.95\textwidth}     \textit{Note}: This figure presents the predicted claim frequency distributions of the test set under three regularisation settings: (a) no regularisation, (b) regularisation using CCdCov, and (c) regularisation using JdCov. The top row shows kernel density estimates (KDEs), while the bottom row displays the corresponding empirical CDFs on the test set. Each color corresponds to a specific region, and line styles distinguish gender (solid for male, dashed for female). The black dashed line represents the reference KDE of the unconditional prediction distribution $\mathbb{P}(\hat{\mathbf{Y}})$. The plots highlight how regularisation aligns the conditional prediction distributions across protected subgroups with the overall distribution.\end{minipage}}
\end{figure}

\begin{table}[ht]
    \centering
    \caption{Performance metrics of a regularised Poisson regressor evaluated on test set (\texttt{pg15training}) \label{tab:test_acc_pg15}}
    {{
\begin{tabular}{lcccccc}
\toprule
 & \multicolumn{2}{c}{(a) Accuracy metrics} & \multicolumn{4}{c}{(b) Fairness metrics} \\
\cmidrule(lr){2-3} \cmidrule(lr){4-7}
Model & RPS & Poisson deviance & CCdCov & JdCov & JS-divergence & $UF(\hat{\mathbf{Y}})$ \\
\midrule
No regularisation & 0.1225 & 0.5450 & 1.0083e-03 & 1.3230e-03 & 0.0282 & 0.0808 \\
CCdCov            & 0.1231 & 0.5494 & 5.8460e-05 & 8.4519e-05 & 0.0018 & 0.0062 \\
JdCov             & 0.1237 & 0.5538 & 5.2875e-05 & 7.5221e-05 & 0.0015 & 0.0057 \\
\bottomrule
\end{tabular}
    }}
    \vspace{0.2cm}
    \caption*{\begin{minipage}{0.95\textwidth}   \textit{Note}: Test performance metrics for Poisson regression models trained with different fairness regularisers: no regularisation, concatenated distance covariance (CCdCov), and joint distance covariance (JdCov). Metrics are reported for the model selected with $\lambda = 40$ for CCdCov and $\lambda=30$ for JdCov, which yielded identical JS-divergence in the validation set. Panel (a) reports accuracy metrics, including the Ranked Probability Score (RPS) and Poisson deviance. Panel (b) reports fairness metrics. Both regularised models reduce disparities between predictions and protected attributes, with CCdCov offering slightly better accuracy with a slight decrease in fairness.\end{minipage}}
\end{table}

Figure~\ref{fig:pg15_results} presents the predicted claim frequency distributions obtained for the test set under different regularisation settings, using kernel density estimates (KDEs) in the top row and empirical cumulative distribution functions (CDFs) in the bottom row. KDEs provide a smooth approximation of the prediction distributions for a given subgroup, but they require bandwidth selection and can be sensitive to the choice of smoothing parameters. In contrast, CDFs offer an unmodified view of distributional differences without need for smoothing. Notably, the area above each subgroup’s CDF corresponds to its mean prediction. Thus, closer alignment across subgroup CDFs implies smaller variation in subgroup means and, consequently, a lower value of the unfairness metric (UF).

Both rows of Figure~\ref{fig:pg15_results} reveal substantial differences in subgroup prediction distributions when no regularisation is applied. As seen in the KDEs, policyholders in region R exhibit a heavier right tail, indicating a higher likelihood of large predicted claim frequencies. This reflects the underlying data, where policyholders in region R indeed have higher average claim frequencies, as shown in Figure~\ref{fig:pg15eda} in Appendix~\ref{apx:pg15}. Applying regularisation with either CCdCov or JdCov visibly reduces these disparities. The KDEs become more similar across subgroups, consistent with reductions in JS-divergence. Similarly, the CDFs become more closely aligned, and the UF values reported in Table~\ref{tab:test_acc_pg15} confirm improved distributional fairness. Between the two methods, CCdCov achieves slightly better accuracy at comparable levels of fairness, as reflected by its lower RPS and UF in Table~\ref{tab:test_acc_pg15}, consistent with the right panel of Figure \ref{fig:calibration_diagnostics_pg15}.

Finally, we assess whether the model overfits to the training data by replicating the main results, which were based on the test set and presented in Figure~\ref{fig:pg15_results} and Table~\ref{tab:test_acc_pg15}. We reproduce these results using the training set, shown in Figure~\ref{fig:pg15_results_train} and Table~\ref{tab:test_acc_pg15_train} in Appendix~\ref{apx:COMPAS_train}. Both the KDEs and CDFs display similar shapes across models, and the metrics show comparable RPS and Poisson deviance, with slightly better accuracy on the training set. Fairness metrics are also similar, with slightly lower values on the training set indicating improved fairness. Again, CCdCov consistently performs slightly better than JdCov in terms of accuracy at similar fairness levels. These findings suggest that the model generalises well and does not exhibit signs of overfitting on this dataset.

\begin{remark}
    In insurance pricing, it is essential for the claim frequency to be unbiased at the portfolio level, so that the predicted claim frequency does not overestimate or underestimate the total claim frequency on average \citep{lindholm2024balance}. However, it is likely that our model is biased, given that we used neural networks with early stopping and included a regularisation term \citep{wuthrich2020bias}. In that case, we suggest applying a bias correction method, such as the simple post-processing method introduced by \citet{lindholm2024balance}, to ensure the balance property of our predicted claim frequencies, so that the sum of predicted claim frequencies is the same as the sum of the observed claim frequencies. In theory, the balance property guarantees an unbiased claim frequency for any future portfolio whose risk profile, that is, the distribution of rating factors, matches that of the training data.
\end{remark}

\section{Conclusion}\label{sec:conclusion}
Fairness in machine learning has attracted growing attention in recent years. However, many existing methods rely on materially simplified assumptions, such as treating protected attributes as binary or considering only a single protected attribute. In this paper, we propose a flexible regularisation framework based on distance covariance and its extensions. Specifically, we initially adopt the joint distance covariance (JdCov) introduced by \citet{chakraborty2019distance}, which captures the mutual association between model predictions and protected attributes. However, JdCov inherently penalises dependencies among the protected attributes themselves, which may be undesirable when these attributes are dependent, and can lead to numerical instability in such settings. To address these issues, we introduce a new regulariser, CCdCov, which computes the distance covariance between model predictions and the joint distribution of protected attributes. This regulariser is designed to parsimoniously handle multiple protected attributes of arbitrary types while effectively addressing fairness gerrymandering.

Our approach allows users to control the trade-off between fairness and accuracy through a tunable regularisation strength. To support practical implementation, we introduced an intuitive and visual calibration strategy using the Jensen-Shannon divergence (JSD), which quantifies differences in prediction distributions across protected subgroups. Importantly, we do not rely on JSD as a standalone criterion. Instead, we combine it with an accuracy metric to jointly assess model accuracy and fairness, ensuring that improvements in fairness do not come at the expense of unacceptable predictive performance.

We demonstrated the effectiveness of our method on two real-world datasets: a binary classification task using the \texttt{COMPAS} dataset, and a Poisson regression task for a large motor insurance claims dataset. The results show that our framework is versatile across different learning tasks, supports multiple and mixed-type protected attributes, and effectively mitigates disparities in predictions across intersectional subgroups.

Future work may extend our regularisation framework to other fairness criteria beyond demographic parity, and develop alternative tools for selecting regularisation strength, particularly for continuous protected attributes without binning. Additionally, methods that are specifically designed to handle limited subgroup sizes would enhance the applicability of fairness approaches in real-world settings with complex group structures.

\section*{Acknowledgments}
We acknowledge the valuable feedback received during presentations at the European Actuarial Journal Conference in Tartu (August 2022), the UNSW Workshop on Risk and Actuarial Frontiers in Sydney (December 2022), the Ph.D. Colloquium at the University of Amsterdam (December 2022), the PARTY Workshop in Valencia (February 2023), the International Congress on Insurance: Mathematics and Economics in Edinburgh (July 2023), the Dependence Modelling (DEMO) Workshop in Agistri, Greece (September 2023), the Bayes Business School Seminar Series in London (September 2023), and the Australasian Actuarial Education and Research Symposium (AAERS) 2024. We thank all participants for their helpful comments.

This research was supported by a joint Ph.D. scholarship between the University of Melbourne and KU Leuven. Benjamin Avanzi acknowledges support from the Australian Research Council’s Discovery Project funding scheme (DP200101859). Katrien Antonio gratefully acknowledges funding from the FWO and Fonds De La Recherche Scientifique - FNRS (F.R.S.-FNRS) under the Excellence of Science (EOS) program, project ASTeRISK Research Foundation Flanders [grant number 40007517]. The views expressed herein are those of the authors and do not necessarily reflect those of the supporting organisations.

\section*{Data and code}
All results in this paper are reproducible using Python, with code and data available on Zenodo (\url{https://zenodo.org/records/16193169}) and GitHub (\url{https://github.com/agi-lab/dCov-fairness}). The code was run in an Azure ML notebook on a \texttt{Standard\_E16s\_v3} virtual machine with 16 cores, which supports the full pipeline but was used specifically for the hyperparameter tuning phase. The Azure notebook is available on Zenodo, while a Quarto-adapted version was used for the final runs and is provided on GitHub. For the \texttt{COMPAS} dataset, hyperparameter tuning took approximately 4 hours. For the \texttt{pg15training} dataset, tuning was stopped after 24 hours, as continuing would have required considerably more time and computational resources. This early stopping is not expected to materially affect the results, as the large dataset size supports good generalisation performance. The remaining parts of the pipeline (training and evaluation using the selected hyperparameters) were run in Quarto locally, taking approximately 5 hours for \texttt{COMPAS} and 7 hours for \texttt{pg15training}.

\section*{References}
\bibliographystyle{plainnat}  
\bibliography{references}

\clearpage
\begin{appendices}
\section{Details on distance covariance}\label{apx:dcov}
To define distance covariance, we first follow the approach by \citet{szekely2007measuring} in defining the characteristic function of a random vector. Let $\mathbf{X}$ be an $m$-dimensional random vector. The characteristic function of $\mathbf{X}$ is given by
\begin{equation}
    \phi_\mathbf{X}(\mathbf{t}) = \mathbb{E}\left[\exp\{\sqrt{-1}\langle \mathbf{t},\mathbf{X}\rangle\}\right],\quad \mathbf{t}\in\mathbb{R}^m
\end{equation}
where $\langle \mathbf{a},\mathbf{b}\rangle = \sum_{i=1}^m a_ib_i$ is the dot product of two vectors $\mathbf{a},\mathbf{b}\in\mathbb{R}^m$. Let $\mathbf{Y}$ be another random vector of dimension $n$. We say $\mathbf{X}$ and $\mathbf{Y}$ are independent if and only if their characteristic function satisfies
\begin{equation}
\begin{aligned}
    &\phi_\mathbf{X}(\mathbf{t})\phi_\mathbf{Y}(\mathbf{s}) = \phi_{\mathbf{X},\mathbf{Y}}(\mathbf{t},\mathbf{s}) \\
    &\qquad= \mathbb{E}[\exp\{\sqrt{-1}\langle \mathbf{t},\mathbf{X}\rangle + \sqrt{-1}\langle \mathbf{s},\mathbf{Y}\rangle\}],\quad \mathbf{t}\in\mathbb{R}^m, \mathbf{s}\in\mathbb{R}^n,
\end{aligned}
\end{equation}
which follows directly from the definition of independence \citep{billingsley2017probability}.  This motivates the development of a measure of distance between $\phi_\mathbf{X}(t)\phi_\mathbf{Y}(s)$ and $\phi_{\mathbf{X},\mathbf{Y}}(t,s)$. Let $f(\cdot)$ be a function with domain $\Omega$ and $w(\cdot)$ be a nonnegative weight function. The weighted $L_2$ norm of $f$ is defined as
\begin{equation}\label{eq:norm}
    ||f||^2_w = \int_\Omega\overline{f(x)}f(x)w(x)dx,
\end{equation}
where $\overline{f(x)}$ is the complex conjugate of $f(x)$. For clarity, if $z=a+b\sqrt{-1}\in\mathbb{C}$, the complex conjugate is $\overline{z} = a-b\sqrt{-1}$.

Using \eqref{eq:norm}, \citet{szekely2007measuring} developed the distance covariance for two random vectors $\mathbf{X}$ and $\mathbf{Y}$, which is the weighted $L_2$ norm of the difference between $\phi_{\mathbf{X},\mathbf{Y}}(t,s)$ and $\phi_\mathbf{X}(t)\phi_\mathbf{Y}(s)$:
\begin{equation}\label{eq:dcovdef}
\mbox{\small$
\begin{aligned}
    &dCov^2(\mathbf{X},\mathbf{Y}) = ||\phi_{\mathbf{X},\mathbf{Y}}(\mathbf{t},\mathbf{s}) - \phi_\mathbf{X}(\mathbf{t})\phi_\mathbf{Y}(\mathbf{s})||^2_w\ \\
    &\qquad = \int_{\mathbb{R}^{m+n}}|\phi_{\mathbf{X},\mathbf{Y}}(\mathbf{t},\mathbf{s}) - \phi_\mathbf{X}(\mathbf{t})\phi_\mathbf{Y}(\mathbf{s})|^2w(\mathbf{t},\mathbf{s})d\mathbf{t}d\mathbf{s},
\end{aligned}$
}
\end{equation}
where 
\begin{subequations}
\begin{align}
    w(\mathbf{t},\mathbf{s}) &= (c_mc_n|\mathbf{t}|^{1+m}_m|\mathbf{s}|^{1+n}_n)^{-1}, \\
    c_d &= \frac{\pi^{(1+d)/2}}{\Gamma((1+d)/2)}, \\
    \Gamma(z) &= \int^\infty_0 t^{z-1}e^{-t}dt.
\end{align}
\end{subequations}
The unbiased estimator of the distance covariance, computed from a sample of $(\mathbf{X},\mathbf{Y})$ of size $n$, denoted by $(\mathbf{x},\mathbf{y})$, is defined as \citep{szekely2014partial}:
\begin{equation}\label{eq:dcov_unbiased}
    \widetilde{dCov}^2(\mathbf{x},\mathbf{y}) = \frac{1}{n(n-3)}\sum^n_{i = 1}\sum^n_{j = 1}\widetilde{U}_\mathbf{x}(i,j)\widetilde{U}_\mathbf{y}(i,j),
\end{equation}
where the $\mathcal{U}$-centred matrix $\widetilde{U}$, also computed on the sample, is defined as
\begin{equation}\label{eq:u_centred}
\widetilde{U}_\mathbf{x}(i,j) =
\begin{cases}
\begin{aligned}
    &|\mathbf{x}_i-\mathbf{x}_j| - \frac{1}{n-2} \sum_{l=1}^{n} |\mathbf{x}_i-\mathbf{x}_l| \\
    &\qquad - \frac{1}{n-2} \sum_{k=1}^{n} |\mathbf{x}_k-\mathbf{x}_j| \\
    &\qquad + \frac{1}{(n-1)(n-2)} \sum_{k=1}^{n}\sum_{l=1}^{n} |\mathbf{x}_k-\mathbf{x}_l|,
\end{aligned}
& i \neq j; \\
0, & i = j.
\end{cases}
\end{equation}
with $|\cdot|$ denoting the Euclidean norm. The $\mathcal{U}$-centred matrix is specifically designed so that the resulting estimator of distance covariance is unbiased under independence, meaning its expectation is zero when the variables are independent \citep{szekely2023energy}. Furthermore, following the proof of Proposition 1 in \citep{szekely2014partial}, we can expand Equation \eqref{eq:dcov_unbiased} into an equivalent form:
\begin{equation}\label{eq:dcov_unbiased_2}
\begin{aligned}
&\widetilde{dCov}^2(\mathbf{x},\mathbf{y}) = \frac{1}{n(n-3)}\Biggl(\sum_{k=1}^n\sum_{l=1}^n|\mathbf{x}_k-\mathbf{x}_l||\mathbf{y}_k-\mathbf{y}_l|\\
&\qquad +\frac{1}{(n-1)(n-2)}\biggl(\sum_{k=1}^n\sum_{l=1}^n|\mathbf{x}_k-\mathbf{x}_l|\biggr)\biggl(\sum_{k=1}^n\sum_{l=1}^n|\mathbf{y}_k-\mathbf{y}_l|\biggr) \biggr.\\
&\qquad - \biggl. \frac{2}{n-2}\sum_k^n\biggl(\sum^n_{l=1}|\mathbf{x}_k-\mathbf{x}_l|\biggr)\biggl(\sum^n_{l=1}|\mathbf{y}_k-\mathbf{y}_l|\biggr)\Biggr).
\end{aligned}
\end{equation}
which will be used in the proof of Theorem \ref{thm:1}. In this paper, we implemented the distance covariance numerically using \eqref{eq:dcov_unbiased} through computing the $\mathcal{U}$-centred distance matrices defined in \eqref{eq:u_centred}.

\section{Measuring mutual association for multiple random vectors}\label{apx:jdcov}
\subsection{Mutual independence}
Let $\mathbf{X}_1, \ldots, \mathbf{X}_d $ be random vectors with dimensions $ p_1, \ldots, p_d $, taking values in $ \mathbb{R}^{p_1}, \ldots, \mathbb{R}^{p_d}$. Note that this also covers one-hot–encoded categorical variables, which can be treated as a \emph{multinoulli} distribution \citep{bishop2006pattern, murphy2012machine}. 

For example, suppose $\mathbf{X}_i = (x_{i,1},x_{i,2},...,x_{i,K})$ encodes a categorical variable with $K$ classes, so $p_i = K$. Each component satisfies $x_{i,k}\in\{0,1\}$ and exactly one entry is 1, i.e. $\sum^K_{k=1}x_{i,k} = 1$. With class probabilities $\boldsymbol{\mu} = (\mu_1,...,\mu_K)$ (where $\sum_{k=1}^K\mu_k=1$), its probability mass function is
\begin{equation}
    p(\mathbf{X}_i = (x_{i,1},x_{i,2},...,x_{i,K})|\boldsymbol{\mu} = (\mu_1,...,\mu_K)) = \prod^K_{k=1}\mu_k^{x_{i,k}}.
\end{equation}

The random vectors $ \mathbf{X}_1, \ldots, \mathbf{X}_d $ are said to be \emph{mutually independent} if and only if for every collection of Borel measurable sets $ \mathbf{A}_1 \subseteq \mathbb{R}^{p_1}, \ldots, \mathbf{A}_d \subseteq \mathbb{R}^{p_d} $,
\begin{equation}
\mathbb{P}(\mathbf{X}_1 \in \mathbf{A}_1, \ldots, \mathbf{X}_d \in \mathbf{A}_d) = \prod_{i=1}^d \mathbb{P}(\mathbf{X}_i \in \mathbf{A}_i).
\end{equation}

\subsection{Joint distance covariance}
To measure the mutual association among an arbitrary number of random vectors, \citet{chakraborty2019distance} developed the joint distance covariance (JdCov), defined as follows. Let $\mathbf{X}_1,\ldots,\mathbf{X}_d$ be $d$ random vectors with dimensions $p_1,\ldots,p_d$, respectively, and let $p_0 = \sum_{i=1}^dp_i$. Then:
\begin{equation}
\begin{aligned}
    &JdCov^2(\mathbf{X}_1,...,\mathbf{X}_d) = \sum_{(\tau_1,\tau_2)\in\mathcal{I}_2}dCov^2(\mathbf{X}_{\tau_1}, \mathbf{X}_{\tau_2}) \\
    &\qquad+ \sum_{(\tau_1,\tau_2,\tau_3)\in\mathcal{I}_3}dCov^2(\mathbf{X}_{\tau_1}, \mathbf{X}_{\tau_2},\mathbf{X}_{\tau_3})\\
    &\qquad+\cdots+dCov^2(\mathbf{X}_1,...,\mathbf{X}_d),
\end{aligned}
\end{equation}
where 
\begin{equation}
\begin{aligned}
    &dCov^2(\mathbf{X}_{\tau_1},...,\mathbf{X}_{\tau_k}) \\
    &\qquad= \int_{\mathbb{R}^{p_{0}}}\left| \mathbb{E}\left[\prod^{k}_{i=1} \left(\phi_{\mathbf{X}_{\tau_i}}(\mathbf{x}_{\tau_i})-\exp\left\{\sqrt{-1}\langle \mathbf{x}_{\tau_i}, \mathbf{X}_{\tau_i}\rangle\right\}\right)\right]\right|^2dw,
\end{aligned}
\end{equation}
with $dw = \prod^{k}_{i=1}\left(c_{p_i}dx_{\tau_i}\right)$ and $\mathcal{I}_k = \left\{ (\tau_1, \ldots, \tau_k) \in \{1, \ldots, d\}^k : \tau_1 < \tau_2 < \cdots < \tau_d \right\}$ being the set of index tuples of size $d$. Intuitively, it is the summation of the pairwise associations ($dCov^2(\cdot,\cdot)$) and all the higher-order associations ($dCov^2(\cdot,\cdot,\cdots,\cdot)$) between the random vectors. The joint distance covariance characterises mutual independence if and only if this quantity is 0. To estimate this quantity from an available sample $\mathbf{x}_1,\ldots,\mathbf{x}_d$ of size $n$, where each $\mathbf{x}_i\in\mathbb{R}^{p_i}$ corresponds to $\mathbf{X}_i$, \citet{chakraborty2019distance} defined the bias-corrected estimator of the joint distance covariance as:
\begin{equation}\label{eq:jdcov}
\begin{aligned}
    &\widetilde{JdCov}^2(\mathbf{x}_1,...,\mathbf{x}_d) \\
    &\qquad= \frac{1}{n(n-3)}\sum_{k=1}^n\sum_{l=1}^n\prod_{i=1}^d\left(\widetilde{U}_{\mathbf{x}_{i}}(k,l)+1\right)-\frac{1}{n(n-3)}\\
    &\qquad=\sum_{1\leq i,j\leq d}\widetilde{dCov}^2(\mathbf{x}_i,\mathbf{x}_j)+ \zeta(\mathbf{x}_1,\ldots,\mathbf{x}_d),
\end{aligned}
\end{equation}
using the $\mathcal{U}$-centred matrices as defined in Equation \eqref{eq:u_centred}, and $\zeta(\mathbf{x}_1,\ldots,\mathbf{x}_d)$ is defined as:
\begin{equation}
    \zeta(\mathbf{x}_1,\ldots,\mathbf{x}_d) = \frac{1}{n(n-3)} \sum_{k=1}^n \sum_{l=1}^n \prod_{i \in \mathcal{I}} \widetilde{U}_{\mathbf{x}_i}(k,l),
\end{equation}
with $\mathcal{I}\subseteq\{1,\ldots,d\}$ and $|\mathcal{I}|\geq 3$. In the setting of model predictions and $d\geq 2$ protected attributes, as in the regularisation term $\psi$ used in Equations \eqref{eq:COMPAS_obj} and \eqref{eq:pg15_obj}, we have:
\begin{equation}\label{eq:jdcov_ys}
\begin{aligned}
    &\widetilde{JdCov}^2(\hat{\mathbf{y}},\mathbf{s}_1,...,\mathbf{s}_d)=\sum_{i=1}^d\widetilde{dCov}^2(\hat{\mathbf{y}},\mathbf{s}_i) \\
    &\qquad+\sum_{1\leq i,j\leq d}\widetilde{dCov}^2(\mathbf{s}_i,\mathbf{s}_j)+ \zeta(\hat{\mathbf{y}},\mathbf{s}_1,...,\mathbf{s}_d),
\end{aligned}
\end{equation}
where the term $\sum_{1\leq i,j\leq d}\widetilde{dCov}^2(\mathbf{s}_i,\mathbf{s}_j)+ \zeta(\hat{\mathbf{y}},\mathbf{s}_1,...,\mathbf{s}_d)$ captures the association within the protected attributes and the extra associations as illustrated in the right panel of Figure \ref{fig:asso1}.

\section{Proof of Theorem \ref{thm:1}}\label{apx:thm1}
For notation simplicity, denote ${\mathbf{s}} = (\mathbf{s}_1,\ldots,\mathbf{s}_d)$ the sample of the protected attributes of size $n$ and ${\mathbf{s}}_k = (\mathbf{s}_{k,1},\ldots,\mathbf{s}_{k,d})$ be the $k$-th sample of ${\mathbf{s}}$. By Equation \eqref{eq:dcov_unbiased_2}, we can estimate the distance covariance between random vector of predictions $\hat{\mathbf{Y}}$ and the vector of protected attributes $\mathbf{S}$ as follows:
\begin{equation}\label{eq:ccdcov_expand}
\begin{aligned}
&\widetilde{dCov}^2(\hat{\mathbf{y}},{\mathbf{s}}) = \frac{1}{n(n-3)}\Biggl[\sum_{k=1}^n\sum_{l=1}^n|\hat{\mathbf{y}}_k-\hat{\mathbf{y}}_l||{\mathbf{s}}_k-{\mathbf{s}}_l|\\
&\qquad +\frac{1}{(n-1)(n-2)}\biggl(\sum_{k=1}^n\sum_{l=1}^n|\hat{\mathbf{y}}_k-\hat{\mathbf{y}}_l|\biggr)\biggl(\sum_{k=1}^n\sum_{l=1}^n|{\mathbf{s}}_k-{\mathbf{s}}_l|\biggr) \biggr.\\
&\qquad - \biggl. \frac{2}{n-2}\sum_{k=1}^n\biggl(\sum^n_{l=1}|\hat{\mathbf{y}}_k-\hat{\mathbf{y}}_l|\biggr)\biggl(\sum^n_{l=1}|{\mathbf{s}}_k-{\mathbf{s}}_l|\biggr)\Biggr].
\end{aligned}
\end{equation}
For $1\leq k,l\leq n$, we can decompose the Euclidean distance $|{\mathbf{s}}_k-{\mathbf{s}}_l|$:
\begin{equation}\label{eq:s_expand}
\begin{aligned}
|{\mathbf{s}}_k-{\mathbf{s}}_l| &= \sqrt{\sum_{i=1}^d|{\mathbf{s}}_{k,i}-{\mathbf{s}}_{l,i}|^2} \\
&= \sum_{i=1}^d|{\mathbf{s}}_{k,i}-{\mathbf{s}}_{l,i}| +\xi(k,l),
\end{aligned}
\end{equation}
where
\begin{equation}
    \xi(k,l) = \begin{cases}
        - \frac{\sum_{1\leq i,j\leq d}2|{\mathbf{s}}_{k,i}-{\mathbf{s}}_{l,i}||{\mathbf{s}}_{k,j}-{\mathbf{s}}_{l,j}|}{|{\mathbf{s}}_{k}-{\mathbf{s}}_{l}|+\sum_{i=1}^d|{\mathbf{s}}_{k,i}-{\mathbf{s}}_{l,i}|}&,{\mathbf{s}}_k\neq{\mathbf{s}}_l\\
        0&, {\mathbf{s}}_k={\mathbf{s}}_l.
    \end{cases}
\end{equation}
Note that the second line of Equation \eqref{eq:s_expand} is coming from the equality:
\begin{equation}
    \sqrt{\sum_{i=1}^d a_i} = \sum_{i=1}^d\sqrt{a_i} - \frac{\sum_{1\leq i,j\leq d}2\sqrt{a_ia_j}}{\sqrt{\sum_{i=1}^d a_i} + \sum_{i=1}^d\sqrt{a_i}},
\end{equation}
which is directly derived by expanding the expression
\begin{equation}
    \left(\sqrt{\sum_{i=1}^d a_i}-\sum_{i=1}^d\sqrt{a_i}\right)\cdot\left(\sqrt{\sum_{i=1}^d a_i}+\sum_{i=1}^d\sqrt{a_i}\right).
\end{equation}
Combining Equations \eqref{eq:ccdcov_expand} and \eqref{eq:s_expand}, we have:
\begin{equation}\label{eq:dcov_decomp}
\widetilde{dCov}^2(\hat{\mathbf{y}},{\mathbf{s}}) = \sum_{i=1}^d\widetilde{dCov}^2(\hat{\mathbf{y}},\mathbf{s}_i) + \eta(\hat{\mathbf{y}},\mathbf{s}_1,\ldots,\mathbf{s}_d)
\end{equation}
with
\begin{equation}\label{eq:eta}
\begin{aligned}
&\eta(\hat{\mathbf{y}},\mathbf{s}_1,\ldots,\mathbf{s}_d) = -\frac{1}{n(n-3)}\Biggl(\sum^n_{k=1}\sum^n_{l=1}|\hat{\mathbf{y}}_k-\hat{\mathbf{y}}_l|\xi(k,l)\\
&\qquad+\frac{1}{(n-1)(n-2)}\biggl(\sum^n_{k=1}\sum^n_{l=1}|\hat{\mathbf{y}}_k-\hat{\mathbf{y}}_l|\biggr)\biggl(\sum^n_{k=1}\sum^n_{l=1}\xi(k,l)\biggr)\\
&\qquad -\frac{2}{n-2}\sum_{k=1}^n\biggl(\sum^n_{l=1}|\hat{\mathbf{y}}_k-\hat{\mathbf{y}}_l|\biggr)\biggl(\sum^n_{l=1}\xi(k,l)\biggr)\Biggr).
\end{aligned}
\end{equation}
Here, each $\widetilde{dCov}^2(\hat{\mathbf{y}},\mathbf{s}_i)$ is given by the unbiased estimator in Equation \eqref{eq:dcov_unbiased_2}. Thus, Equation \eqref{eq:dcov_decomp} can be understood as decomposing the joint distance covariance into the sum of the pairwise distance covariances and the residual intersection term $\eta(\hat{\mathbf{y}},\mathbf{s}_1,\ldots,\mathbf{s}_d)$.

The term $\eta(\hat{\mathbf{y}},\mathbf{s}_1,\ldots,\mathbf{s}_d)$ captures the dependence that arises only when the protected features are considered together. It is the residual link between the prediction $\hat{\mathbf{Y}}$ and the joint vector $\mathbf{S}=(\mathbf{S}_1,\ldots,\mathbf{S}_d)$ after removing the sum of the marginal dependencies with each $\mathbf{S}_i$. If $\eta$ is large, the prediction is reacting to the intersections of the protected attributes. In that case, even if the prediction and each protected attribute have a small distance covariance, indicating empirical independence, the predictions and the protected attributes are empirically mutually dependent. If $\hat{\mathbf{Y}}$ is independent of the whole vector $\mathbf{S}$, then the terms on the RHS of Equation \eqref{eq:dcov_decomp} should be close to zero; i.e., $\hat{\mathbf{Y}}$ is independent of each protected attribute, and $\hat{\mathbf{Y}}$ is also independent of every subset of the protected attributes, leading to mutual independence.

\section{Details on model training and tuning}\label{apx:model_arch}
\subsection{Justification of applying gradient descent}
Our aim is to minimise the objective function:
\begin{equation}
\frac{1}{n}\sum^n_{i=1}
\mathcal{L}(\hat{\mathbf{y}}_{\Theta i},\mathbf{y}_i) +
\lambda \cdot \psi\left(\hat{\mathbf{y}}, \mathbf{s}_1,\ldots,\mathbf{s}_d\right),
\end{equation}
by adjusting the model parameters $\Theta$, where $\mathcal{L}(\cdot,\cdot)$ is a task-specific loss function that is assumed to be differentiable. To enable gradient-based optimisation using AdaHessian, our implementation requires the availability of gradient and Hessian information with respect to $\Theta$ for the full objective. Given that $\mathcal{L}$ is differentiable, we show that our regulariser $\psi$ also satisfies these properties, inspired by \citet{schmidt2007fast} and \citet{shalev2014understanding}. This allows efficient second-order optimisation of the regularised loss by leveraging curvature information to accelerate convergence.

Let $\Theta = (\theta_1, \dots, \theta_m) \in \mathbb{R}^m$ denote the model parameters, and let $\hat{\mathbf{y}}_\Theta = \{f(\mathbf{x}_i; \Theta)\}_{i=1}^n$ denote the model predictions produced by a neural network. From Equation \eqref{eq:dcov_unbiased} and \eqref{eq:jdcov}, our regularisers are computed with $\mathcal{U}$-centred matrices in \eqref{eq:u_centred}, which involve terms of the form:
\begin{equation}\label{outputdiff} 
|\hat{\mathbf{y}}_{\Theta k} - \hat{\mathbf{y}}_{\Theta l}| = |f(\mathbf{x}_k; \Theta) - f(\mathbf{x}_l; \Theta)|, 
\end{equation} 
for all $k, l \in {1, \dots, n}$. Define $g_{kl}(\Theta) = f(\mathbf{x}_k; \Theta) - f(\mathbf{x}_l; \Theta)$ for notational simplicity.

We first consider the trivial case where $\mathbf{x}_k = \mathbf{x}_l$. Then for all $\Theta$, we have $f(\mathbf{x}_k; \Theta) = f(\mathbf{x}_l; \Theta)$, so $g_{kl}(\Theta) = 0$ and consequently we have zero gradient everywhere.

Now consider the case where $\mathbf{x}_k \ne \mathbf{x}_l$. If $g_{kl}(\Theta) \ne 0$, then $|\cdot|$ is differentiable and the gradient is given by:
\begin{equation}
\begin{aligned}
    \nabla_\Theta |g_{kl}(\Theta)| &= \begin{bmatrix} 
    \dfrac{\partial |g_{kl}(\Theta)|}{\partial\theta_{1}} & \dots  & \dfrac{\partial  |g_{kl}(\Theta)|}{\partial\theta_{m}}
    \end{bmatrix}\\
    &=\text{sign}(g_{kl}(\Theta))\begin{bmatrix} 
    \dfrac{\partial g_{kl}(\Theta)}{\partial\theta_{1}} & \dots  & \dfrac{\partial g_{kl}(\Theta)}{\partial\theta_{m}}
    \end{bmatrix}\\
    &=\text{sign}(g_{kl}(\Theta))\nabla_\Theta(g_{kl}(\Theta)),
\end{aligned}
\end{equation}
where $\text{sign}(g_{kl}(\Theta)) = 1$ if $g_{kl}(\Theta) > 0$ and $-1$ if $g_{kl}(\Theta) < 0$.

At the non-differentiable point where $g_{kl}(\Theta) = 0$, we rely on the concept of subgradients \citep{clarke1990optimization}:
\begin{definition}
Let $h: \mathbb{R}^m \to \mathbb{R}$. A vector $\mathbf{v} \in \mathbb{R}^m$ is a subgradient of $h$ at $\mathbf{x}$  if for all $\mathbf{x}' \in \mathbb{R}^m$,
\begin{equation}
    h(\mathbf{x}')\geq h(\mathbf{x}) + \mathbf{v}^T(\mathbf{x}'-\mathbf{x}).
\end{equation}
\end{definition}
According to Theorem 2.3.10 in \citet{clarke1990optimization}, the subgradient of $|g_{kl}(\Theta)|$ at $g_{kl}(\Theta) = 0$ can be expressed as:
\begin{equation}
\{\gamma\nabla_\Theta(g_{kl}(\Theta))\mid\gamma\in[-1,1]\}.
\end{equation}
In our implementation, we use PyTorch (v2.5.1), which supports automatic differentiation through subgradients at non-differentiable points via its autograd engine \citep{pytorchautograd}. This allows us to apply gradient-based optimisation even when the loss function is not fully differentiable.

For second-order information, we use AdaHessian to approximate the diagonal of the Hessian via Hutchinson’s method \citep{yao2021adahessian}. Our objective is piecewise differentiable. Away from points where $g_{kl}(\Theta)\neq 0$, autograd provides exact derivatives and AdaHessian computes Hessian vector products as usual. At $g_{kl}(\Theta)=0$ the true second-order derivative is not defined; PyTorch returns a particular subgradient and an effective zero curvature at that point. Therefore, the curvature used by AdaHessian at the non-smooth locations should be viewed as a practical approximation rather than a formal second-order quantity.

\subsection{Hyperparameter tuning process}
As mentioned in Section~\ref{sec:modelcal}, the hyperparameter tuning process involves Gaussian process-based Bayesian optimisation \citep{snoek2012practical}. We used the \texttt{gp\_minimize} function from the \texttt{skopt} library in Python to perform this tuning. Algorithm \ref{alg:bayes_cv} provides the pseudocode of our hyperparameter search procedure.
\begin{algorithm}[h!]
\begin{small}
\captionsetup{font=small}
\caption{Bayesian optimisation with 5-fold cross validation for hyperparameter tuning}
\label{alg:bayes_cv}
\begin{algorithmic}[1]
\Require Learning algorithm $\mathcal{A}$, hyperparameter space $\mathcal{H}$, training dataset $\mathcal{D}_{\text{train}}$, number of iterations $\mathcal{I}$, number of initial random trials $n_{\text{rand}}$
\Ensure Optimal hyperparameter set $h^* \in \mathcal{H}$
\State Split $\mathcal{D}_{\text{train}}$ into 5 disjoint stratified subsets $\mathcal{D}_1, \dots, \mathcal{D}_5$
\State Initialise best loss $L^* \gets \infty$
\For{$i = 1$ to $\mathcal{I}$}
    \If{$i \le n_{\text{rand}}$}
        \State Randomly select candidate hyperparameter set $h_i \in \mathcal{H}$
    \Else
        \State Select $h_i \in \mathcal{H}$ using Bayesian optimisation based on previous evaluations
    \EndIf
    \For{$k = 1$ to $5$}
        \State $\mathcal{D}_{\text{valid}} \gets \mathcal{D}_k$
        \State $\mathcal{D}_{\text{subtrain}} \gets \bigcup_{\substack{j = 1 \\ j \neq k}}^{5} \mathcal{D}_j$
        \State Train model $\mathcal{M}_{h_i}^{(k)} \gets \mathcal{A}(h_i, \mathcal{D}_{\text{subtrain}})$
        \State Evaluate $\mathcal{M}_{h_i}^{(k)}$ on $\mathcal{D}_{\text{valid}}$ to obtain loss $L_{h_i,k}$
    \EndFor
    \State Compute $L_{h_i} = \frac{1}{5} \sum_{k=1}^{5} L_{h_i,k}$
    \If{$L_{h_i} < L^*$}
        \State $L^* \gets L_{h_i}$
        \State $h^* \gets h_i$
    \EndIf
    \State Update Bayesian optimiser with $(h_i, L_{h_i})$
\EndFor
\State \Return $h^*$
\end{algorithmic}
\end{small}
\end{algorithm}

\section{Details on model performance metrics}\label{apx:performance}
Here we provide the details on the model performance metrics discussed in Section \ref{sec:modelcal}.
\subsection{Details on hypothesis tests for independence}\label{apx:performance_fair}
When we use hypothesis tests to calibrate $\lambda$, we proceed as follows. After training the model on the subtraining set, we apply the test on the validation set and record the corresponding $p$-value for each candidate $\lambda$. We then choose the \emph{smallest} $\lambda$ for which the null of independence cannot be rejected, i.e., $p \ge \alpha$ for a chosen threshold $\alpha$. This keeps regularisation as weak as possible, thereby minimising the accuracy trade-off, while still achieving statistical (mutual) independence between the model predictions and the protected attributes. Below, we provide the details of the hypothesis tests.
\vspace{0.5cm}
\begin{enumerate}
    \item \underline{Distance correlation $\chi^2$-test}: This test was developed by \citet{shen2022chi}. It is based on the distance correlation, which is a standardised version of the distance covariance as discussed in Appendix \ref{apx:dcov}. Let $(\mathbf{x},\mathbf{y})$ be the corresponding sample of size $n$ of $(\mathbf{X}, \mathbf{Y})$. The distance correlation is  defined as:
    \begin{equation}
        \widetilde{dCorr}^2(\mathbf{x}, \mathbf{y}) = \frac{\widetilde{dCov}^2(\mathbf{x}, \mathbf{y})}{\sqrt{\widetilde{dCov}^2(\mathbf{x},\mathbf{x})\cdot\widetilde{dCov}^2(\mathbf{y},\mathbf{y})}}\in[-1,1]
    \end{equation}
    and is set to $0$ if the denominator is not a positive real number. This test is used to test the hypothesis stated in Equation \eqref{eq:hyp_joint} and the $p$-value is computed as:
    \begin{equation}
        p = \mathbb{P}\left(\chi^2_1>n\cdot \widetilde{dCorr}^2\left(\hat{\mathbf{y}}, (\mathbf{s}_1,\ldots,\mathbf{s}_d)\right)\right).
    \end{equation}

    \item \underline{Permutation test \citep{szekely2009brownian, chakraborty2019distance}}: There are two types of permutation tests that can be used to assess statistical dependence in our setting. The first type tests the independence between the model predictions and the joint distribution of the protected attributes, as stated in Equation~\eqref{eq:hyp_joint}. This is achieved by permuting the protected attribute samples randomly, generating datasets consistent with the null hypothesis of independence between the model predictions and the protected attributes.

    Let $T_0 = \widetilde{dCov}^2(\hat{\mathbf{y}}, (\mathbf{s}_1, \ldots, \mathbf{s}_d))$ denote the test statistic computed from the original data. To approximate the null distribution, we generate $R$ permuted versions of the protected attributes. This is done by randomly shuffling the protected attribute samples as a whole, and therefore we shuffle the row numbers of the protected attributes. This preserves the relationships between the protected attributes, but breaks any link they might have with the model predictions. This simulates a situation where the predictions and protected attributes are independent, which we use to compare against the original data.
    
    For each permutation $r = 1, \ldots, R$, we compute the test statistic $T^{(r)} = \widetilde{dCov}^2(\hat{\mathbf{y}},(\mathbf{s}_1,\ldots,\mathbf{s}_d)^{(r)})$ where $(\mathbf{s}_1, \ldots, \mathbf{s}_d)^{(r)}$ is the $r$-th permuted sample. The $p$-value is then computed as: 
    \begin{equation} \label{eq:perm_test}
    p = \frac{1 + \sum_{r=1}^{R} \mathbb{I}(T^{(r)} > T_0)}{1 + R}.
    \end{equation}

    Intuitively, permuting the sample breaks any dependency structure between $\hat{\mathbf{y}}$ and $(\mathbf{s}_1,\ldots,\mathbf{s}_d)$. In that case, $T^{(r)}$ should be close to zero. If $p$ is large, which means that many $T^{(r)}$, the distance covariances for supposedly independent predictions and protected attributes, are larger than the distance covariance of the original sample $T_0$, then it is very likely that our original sample is already independent.
    
    The second type of permutation test looks at the mutual independence between the model prediction and each protected attribute, as defined in Equation~\eqref{eq:hyp_mutual}. Similar to the first test, we compute the $p$-value using the same formula in Equation~\eqref{eq:perm_test}, but here the test statistic is based on joint distance covariance: $T_0 = \widetilde{JdCov}^2(\hat{\mathbf{y}},\mathbf{s}_1,\ldots,\mathbf{s}_d)$ and $T^{(r)} = \widetilde{JdCov}^2(\hat{\mathbf{y}},\mathbf{s}_1^{(r)},\ldots,\mathbf{s}_d^{(r)})$, as defined in Equation \eqref{eq:jdcov_ys}.
    
    In this case, we permute each protected attribute separately. This breaks not only the association between the model prediction and each protected attribute, but also the dependencies among the protected attributes themselves. As a result, when the protected attributes are correlated, this test is more likely to reject the null hypothesis of mutual independence, potentially leading to uninformative $p$-values as discussed in Section \ref{sec:sig_metric}.
\end{enumerate}

\subsection{Details on ranked probability score}\label{apx:performance_acc}
Let $K$ denote the number of ordered outcome categories in the prediction task (e.g. $0$ and $1$ for binary classification, $1$ to $K$ for $K$-class classification). The Ranked Probability Score (RPS), defined in \citet{weigel2007discrete}, for the $i$-th prediction is defined as:
\begin{equation}
RPS_i = \sum_{k=1}^K \left(Y_{ik} - O_{ik}\right)^2,
\end{equation}
where $Y_{ik} = \sum_{j=1}^k y_{ij}$ is the cumulative predicted probability for the $i$-th prediction up to category $k$, with $y_{ij}$ denoting the predicted probability that prediction $i$ falls in category $j$. Similarly, $O_{ik} = \sum_{j=1}^k o_{ij}$ is the cumulative observed indicator, where $o_{ij} = 1$ if the $i$-th prediction belongs to category $j$, and $0$ otherwise.

Intuitively, RPS evaluates how well the predicted probabilities match the true outcome by summing the squared differences between the predicted and observed cumulative probabilities across all categories. It penalises predictions that place too much probability mass far from the true class, encouraging forecasts that assign high probability to the correct category and its neighbouring lower categories.

The average RPS across all $n$ predictions is then given by:
\begin{equation}
\frac{1}{n} \sum_{i=1}^n RPS_i.
\end{equation}

For Poisson regression tasks, since the number of possible ordered outcome categories is infinite, we choose a sufficiently large maximum value $K$ to truncate the range and compute the RPS accordingly.

\section{Pseudocode for regularisation strength calibration}\label{apx:lambdacal}
\begin{algorithm}
\begin{small}
\captionsetup{font=small}
\caption{Calibrating Regularisation Strength: The procedure evaluates model performance for each predetermined value of $\lambda$ and allows users to select the optimal regularisation strength $\lambda^*$.}
\label{alg:lambdatune}
\begin{algorithmic}[1]
\Require Model $\mathcal{M}$, optimal hyperparameters $h^*$, training dataset $\mathcal{D}_{\text{train}}$, candidate regularisation strengths $\Lambda$
\Ensure Selected regularisation strength $\lambda^* \in \Lambda$
\State Split $\mathcal{D}_{\text{train}}$ into subtraining set $\mathcal{D}_{\text{subtrain}}$ (70\%) and validation set $\mathcal{D}_{\text{valid}}$ (30\%) with stratification on the model output and the protected attributes
\State Train baseline model $\mathcal{M}_0$ with $\lambda = 0$ and hyperparameters $h^*$ on $\mathcal{D}_{\text{subtrain}}$
\State Evaluate baseline fairness $F_0$ and accuracy loss $L_0$ on $\mathcal{D}_{\text{valid}}$ \Comment{In our applications, we used JS-divergence for $F$ and $RPS$ for $L$.}
\State Initialise fairness set $\mathcal{F} \gets \{F_0\}$ and accuracy set $\mathcal{L} \gets \{L_0\}$
\For{$\lambda \in \Lambda$}
    \State Train model $\mathcal{M}_\lambda$ with regularisation strength $\lambda$ and hyperparameters $h^*$ on $\mathcal{D}_{\text{subtrain}}$ 
    \State Evaluate fairness $F_\lambda$ and loss $L_\lambda$ on $\mathcal{D}_{\text{valid}}$
    \State Append $F_\lambda$ to $\mathcal{F}$ and $L_\lambda$ to $\mathcal{L}$
\EndFor
\State Select $\lambda^*$ based on the trade-off between fairness $F_{\lambda^*}\in\mathcal{F}$ and accuracy $L_{\lambda^*}\in\mathcal{L}$ \Comment{e.g. take the “elbow’’ of the JSD–loss curve to pick the smallest $\lambda$ such that increase $\lambda$ further will bring negligible reductions in JSD.}
\State \Return $\lambda^*$
\end{algorithmic}
\end{small}
\end{algorithm}

\section{Details of the \texttt{COMPAS} data}\label{apx:compas}
\subsection{Data pre-processing}\label{apx:compas_prep}
In our application, we followed most of the data pre-processing as done by ProPublica in their analysis \citep{angwin2016machine}. The following data cleaning steps were applied to prepare the data for analysis:
\begin{itemize}
    \item Removed observations without a \texttt{COMPAS} case.
    \item Removed records where the crime charge date that triggered the \texttt{COMPAS} score was not within 30 days of the arrest date, to ensure temporal consistency.
    \item Excluded cases involving ordinary traffic offences.
    \item Removed observations with unknown \texttt{COMPAS} scores.
\end{itemize}
Table \ref{tab:details_COMPAS} provides the details of variables used in our model training.
\begin{table}[ht]
    \centering
    \caption{Details of variables in \texttt{COMPAS} used for model training \label{tab:details_COMPAS}}
    {\resizebox{\textwidth}{!}{
\begin{tabular}{p{3cm} p{4cm} p{4cm} p{5cm}}
\toprule
\textbf{Variable} & \textbf{Type and Range} & \textbf{Pre-processing} & \textbf{Notes} \\
\midrule
\multicolumn{4}{l}{\textit{Response Variable}} \\
\texttt{two\_year\_recid} & Binary (0,1) & None & Indicates whether the defendant recidivate within two years. \\
\midrule
\multicolumn{4}{l}{\textit{Defendant’s attributes}} \\
\texttt{Female} & Categorical (Male, Female) & One-hot encoding (Male=0, Female=1)  & Renamed from \texttt{sex}. Protected attribute representing the defendant's gender. Included in model input. \\
\texttt{Ethnicity} & Categorical (4 classes) & One-hot encoding & Protected attribute included in the input. Includes African-American, Caucasian, Hispanic, Others. Originally 6 categories; Asian, Native American, and Others grouped into \texttt{Others} as they contributed less than 1\% of the data. Renamed from \texttt{race}. \\
\texttt{Felnoy} & Categorical (Felony, Misdemeanor) & One-hot encoding (F=1, M=0) & Indicates whether the offence is a felony. Renamed from \texttt{c\_charge\_degree}. \\
\texttt{age} & Integer (18-96) & Min-max scaling & Protected attribute included in the input. Also binned by percentiles (33rd, 67th) for EDA and computing JS-divergence, but binning not used in training \\
\texttt{priors\_count} & Integer (0–38) & Min-max scaling & Number of prior offenses \\
\bottomrule
\end{tabular}
    }}
    \vspace{0.2cm}
    \caption*{     \begin{minipage}{0.95\textwidth}\textit{Note}: This table provides details of the variables used in model training for the recidivism prediction application in Section \ref{sec:COMPAS_app}, including each variable’s type and range, pre-processing steps, and additional notes.\end{minipage}}
\end{table}

\subsection{Neural network structure}
In this application, we used a sigmoid activation function in the output layer. We also employed mini-batch learning for model training and applied dropout to prevent overfitting. Table~\ref{tab:hp_COMPAS} summarises the hyperparameter settings used for the final model under each regularisation configuration.
\begin{table}[ht]
    \centering
    \caption{Hyperparameters of a neural network binary classifier applied on training set (\texttt{COMPAS}) \label{tab:hp_COMPAS}}
    {\resizebox{\textwidth}{!}{
\begin{tabular}{lccccccc}
\toprule
Model & Learning rate & Batch size & No. of layers & No. of nodes & Dropout & Hessian power & AdaHessian $(\beta_1,\beta_2)$ \\
\midrule
No regularisation & 0.0014 & 512 & 2 & 256 & 0.0749 & 1.0 & (0.95, 0.999)\\
CCdCov            & 0.01  & 256 & 3 & 128 & 0.0755 & 0.5 & (0.95, 0.999)\\
JdCov             & 0.01 & 1024 & 2 & 512 & 0.05 & 0.75 & (0.95, 0.999)\\
\bottomrule
\end{tabular}
    }}
    \vspace{0.2cm}
    \caption*{     \begin{minipage}{0.95\textwidth}\textit{Note}: This table reports the detailed hyperparameter settings for the binary classifier trained on the \texttt{COMPAS} dataset. Each row corresponds to a model with a specific regulariser. The hyperparameters shown are those re-tuned after selecting the regularisation strength $\lambda$, using the full training set with 5-fold cross-validation.\end{minipage}}
\end{table}

\clearpage
\subsection{Training set results}\label{apx:COMPAS_train}
\begin{figure}[h!]
    \centering
    \caption{Average predicted recidivism rates on training set under different regularisers (\texttt{COMPAS}) \label{fig:compasresult_train}}
    \includegraphics[width=1\linewidth]{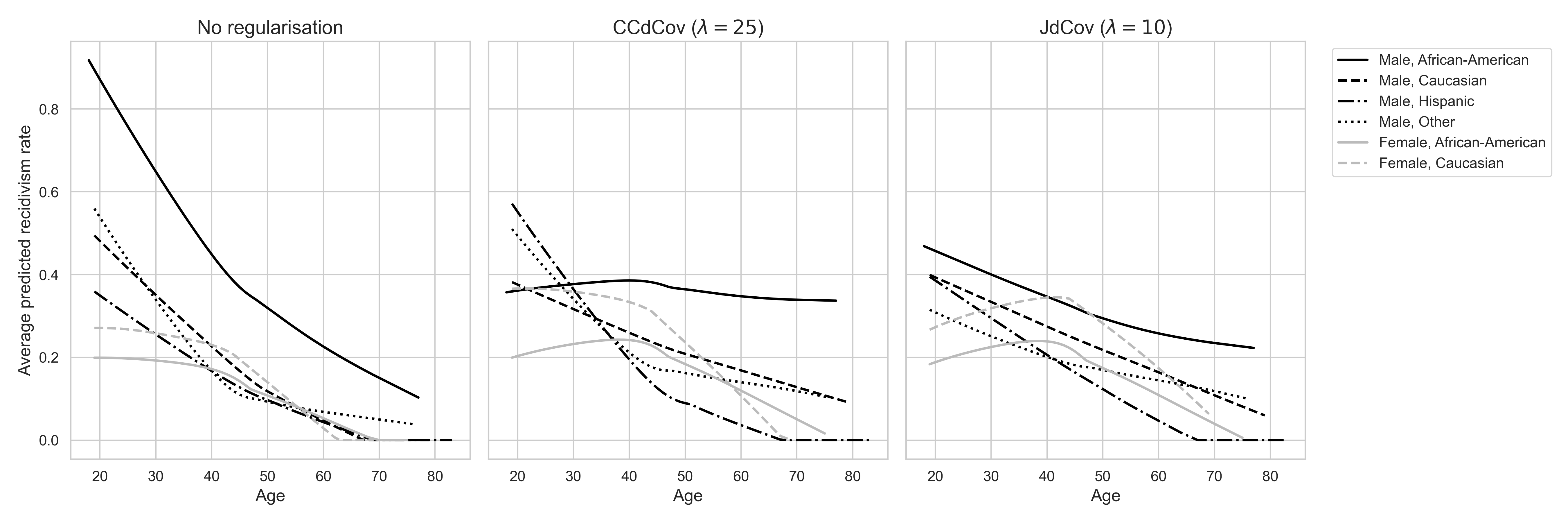}
    \caption*{     \begin{minipage}{0.95\textwidth}\textit{Note}: Average predicted recidivism rates of the training set by age using lowess smoothing (span = 1), split by gender–ethnicity subgroups with at least 100 training samples. Results are shown for three models: (left) no regularisation, (middle) CCdCov regularisation with $\lambda=25$, and (right) JdCov regularisation with $\lambda=10$. Regularisation yields flatter and closer prediction curves across age.\end{minipage}}
\end{figure}

\begin{table}[h!]
    \centering
    \caption{Performance metrics of a regularised binary classifier evaluated on training set (\texttt{COMPAS}) \label{tab:train_acc}}
    {{
\begin{tabular}{lcccccc}
\toprule
 & \multicolumn{2}{c}{(a) Accuracy metrics} & \multicolumn{4}{c}{(b) Fairness metrics} \\
\cmidrule(lr){2-3} \cmidrule(lr){4-7}
Model & RPS & ACC & CCdCov & JdCov & JS-divergence & $UF(\hat{\mathbf{Y}})$ \\
\midrule
No regularisation & 0.2018 & 0.6887 & 0.0110 & 0.0192 & 0.2020 & 0.3310 \\
CCdCov            & 0.2117 & 0.6731 & 0.0009 & 0.0048 & 0.0318 & 0.1354 \\
JdCov             & 0.2101 & 0.6737 & 0.0009 & 0.0048 & 0.0316 & 0.1087 \\
\bottomrule
\end{tabular}
    }}
    \vspace{0.2cm}
    \caption*{     \begin{minipage}{0.95\textwidth}\textit{Note}: Training set performance metrics for three binary classifiers trained on the \texttt{COMPAS} dataset with different regularisers, evaluated at $\lambda = 10$ for JdCov and $\lambda=25$ for CCdCov. Panel (a) reports accuracy metrics (RPS and ACC), while Panel (b) presents fairness metrics: CCdCov, JdCov, JS-divergence, and the unfairness measure $UF(\hat{\mathbf{Y}})$.\end{minipage}}
\end{table}

\section{Details of the \texttt{pg15training} data}\label{apx:pg15}
\subsection{Data pre-processing}
There are no missing or \texttt{NA} values in the dataset. However, as noted by \citet{xin2024antidiscrimination}, the first 21 observations are duplicate records with zero third-party property damage claims. After review, we confirmed the duplication and have removed these records accordingly. Table \ref{tab:details_pg15} displays the details of the variables used in our application.
\begin{table}[ht]
    \centering
    \caption{Details of variables in \texttt{pg15training} used for model training \label{tab:details_pg15}}
    {\resizebox{\textwidth}{!}{
\begin{tabular}{p{3cm} p{4cm} p{4cm} p{5cm}}
\toprule
\textbf{Variable} & \textbf{Type and Range} & \textbf{Pre-processing} & \textbf{Notes} \\
\midrule
\multicolumn{4}{l}{\textit{Response Variable}} \\
\texttt{nclaims} & Count (0--7) & None & Renamed from \texttt{Numtppd}, counts third-party material claims. \\
\midrule
\multicolumn{4}{l}{\textit{Driver's attributes}} \\
\texttt{Female} & Categorical (Male, Female) & One-hot encoding (Male=0, Female=1) & Renamed from \texttt{Gender}. Protected attribute representing the policyholder's gender. Excluded from model input. \\
\texttt{Occupation} & Categorical (5 classes) & One-hot encoding & Includes Employed, Unemployed, Housewife, Self-employed, Retired. \\
\texttt{Region} & Categorical (10 classes) & One-hot encoding & Protected attribute representing the policyholder's living region. Renamed from \texttt{Group2} with 10 classes (region L to region U), excluded from input. \\
\texttt{Age} & Integer (18--75) & Min-max scaling & -- \\
\texttt{Bonus} & Integer (-50--150) & Min-max scaling & Represents no-claim discount (bonus-malus). \\
\texttt{Exposure} & Continuous (91--365 days) & Converted to years & Renamed from \texttt{Exppdays}, used as model offset. \\
\texttt{PolDur} & Continuous (0--15) & Min-max scaling & Policy duration in years. \\
\texttt{Density} & Continuous (14.38--297.39) & Min-max scaling & Population density per km$^2$. \\
\midrule
\multicolumn{4}{l}{\textit{Vehicle's attributes}} \\
\texttt{CarCat} & Categorical (Small, Medium, Large) & Ordinal encoding (Small=0, Medium=1, Large=2)& Renamed from \texttt{Category}. \\
\texttt{CarType} & Categorical (A--F) & One-hot encoding & Renamed from \texttt{Type}. \\
\texttt{CarGroup} & Categorical (1--20) & One-hot encoding & Renamed from \texttt{Group1}. \\
\texttt{Value} & Continuous (1,000--49,995) & Min-max scaling & Vehicle's market value in Euros. \\
\bottomrule
\end{tabular}
    }}
    \vspace{0.2cm}
    \caption*{     \begin{minipage}{0.95\textwidth}\textit{Note}: This table provides details of the variables used in model training for the motor insurance claims application in Section \ref{sec:pg15_app}, including each variable’s type and range, pre-processing steps, and additional notes.\end{minipage}}
\end{table}

\clearpage
\subsection{Exploratory data analysis}
\begin{figure}[h]
    \centering
    \caption{Average Claim Frequency by Region and Gender \label{fig:pg15eda}}
    {\includegraphics[width=0.6\linewidth]{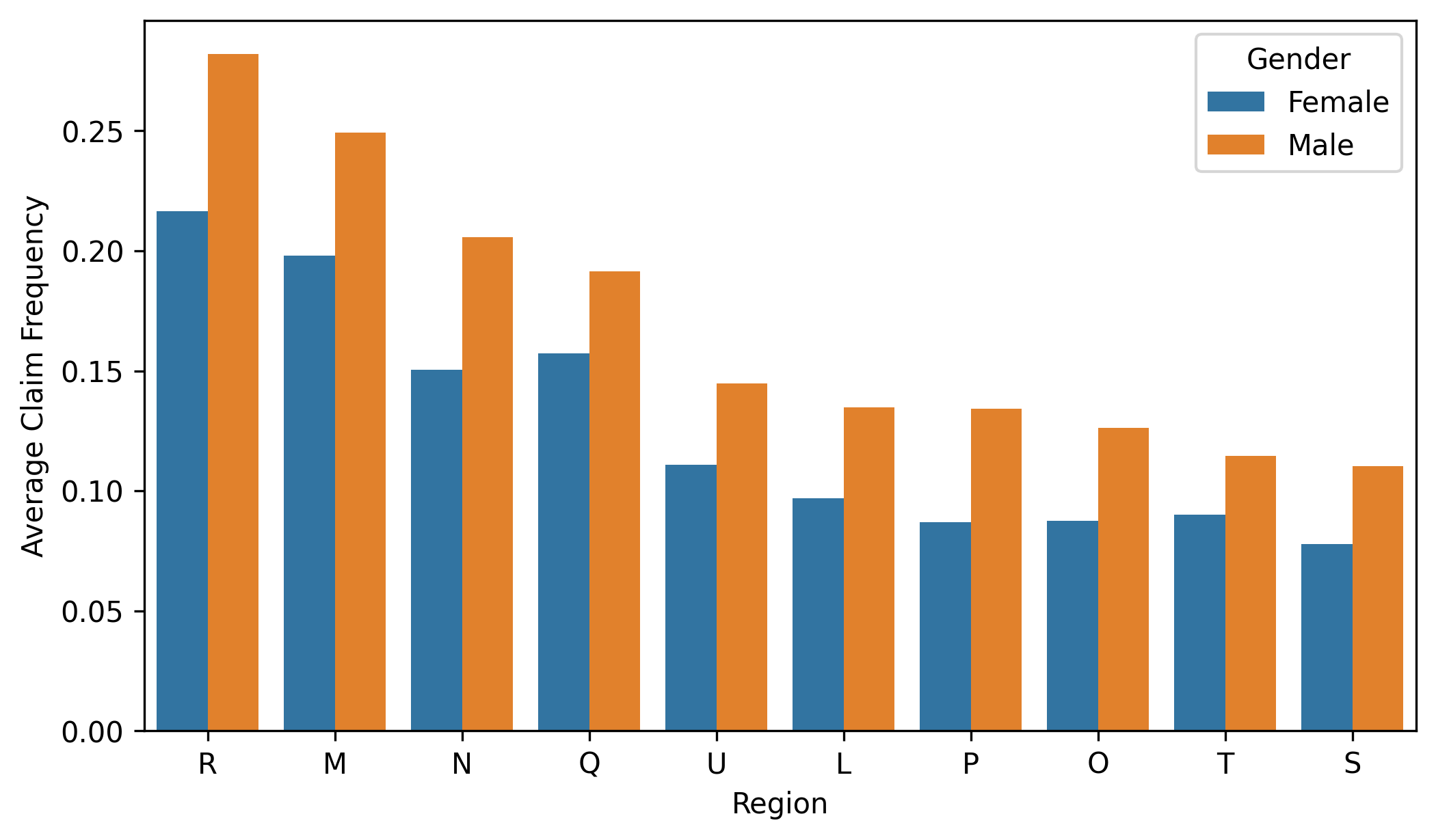}}
    \caption*{     \begin{minipage}{0.95\textwidth}\textit{Note}: This bar chart shows the average claim frequency of third-party property damage claims by region and gender. Claim frequency is calculated as the number of claims divided by the exposure, measured as the policyholder’s time at risk during the policy year.\end{minipage}}
\end{figure}
\begin{figure}[h]
    \centering
    \caption{Proportion of Gender by Region \label{fig:pg15eda2}}
    \includegraphics[width=0.6\linewidth]{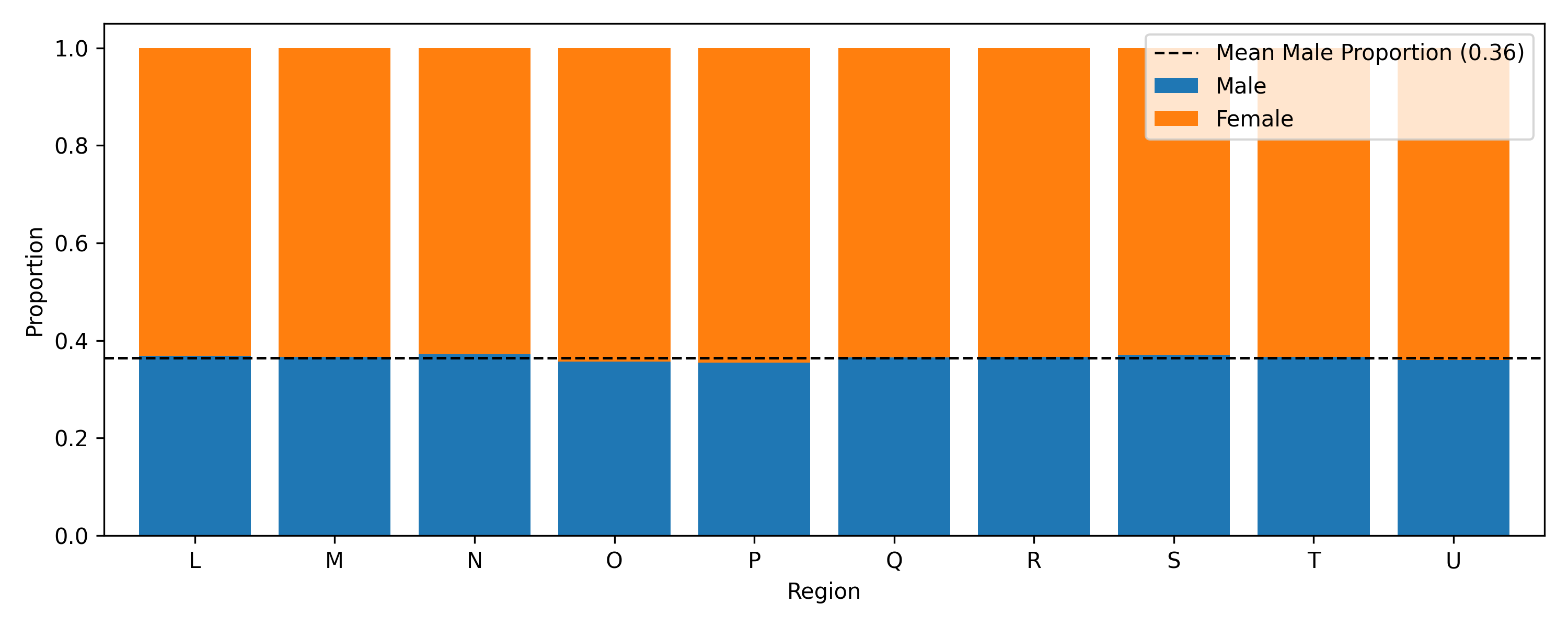}
    \caption*{     \begin{minipage}{0.95\textwidth}\textit{Note}: This bar chart illustrates the proportion of males and females across different regions. The dashed line indicates the average male proportion across all regions. As shown in the plot, the male proportions remain relatively consistent across regions, suggesting that there is little to no association between region and gender.\end{minipage}}
\end{figure}

\subsection{Neural network structure}
In this application, we used a Poisson activation function with offset in the output layer. We also employed mini-batch learning for model training and applied dropout to prevent overfitting. Table~\ref{tab:hp_pg15} summarises the hyperparameter settings used for the final model under each regularisation configuration.
\begin{table}[ht]
    \centering
    \caption{Hyperparameters of a neural network Poisson regressor applied on training set (\texttt{pg15training}) \label{tab:hp_pg15}}
    {\resizebox{\textwidth}{!}{
\begin{tabular}{lccccccc}
\toprule
Model & Learning rate & Batch size & No. of layers & No. of nodes & Dropout & Hessian power & AdaHessian $(\beta_1,\beta_2)$ \\
\midrule
No regularisation & 9.3106e-03 & 128 & 4 & 64 & 0.3398 & 0.5 & (0.85, 0.95)\\
CCdCov            & 1.6007e-03  & 128 & 2 & 256 & 0.2598 & 1.0 & (0.85, 0.95)\\
JdCov             & 1.1594e-03 & 128 & 2 & 64 & 0.1129 & 0.75 & (0.85, 0.95)\\
\bottomrule
\end{tabular}
    }}
    \vspace{0.2cm}
    \caption*{     \begin{minipage}{0.95\textwidth}\textit{Note}: This table reports the detailed hyperparameter settings for the Poisson regressor trained on the \texttt{pg15training} dataset. Each row corresponds to a model with a specific regulariser. The hyperparameters shown are those re-tuned after selecting the regularisation strength $\lambda$, using the full training set with 5-fold cross-validation.\end{minipage}}
\end{table}

\newpage
\subsection{Training set results}\label{apx:pg15_train}
\begin{figure}[h]
    \centering
    \caption{Kernel density estimates and empirical CDF of predicted claim frequency on training set (\texttt{pg15training}) \label{fig:pg15_results_train}}
    \includegraphics[width=1\linewidth]{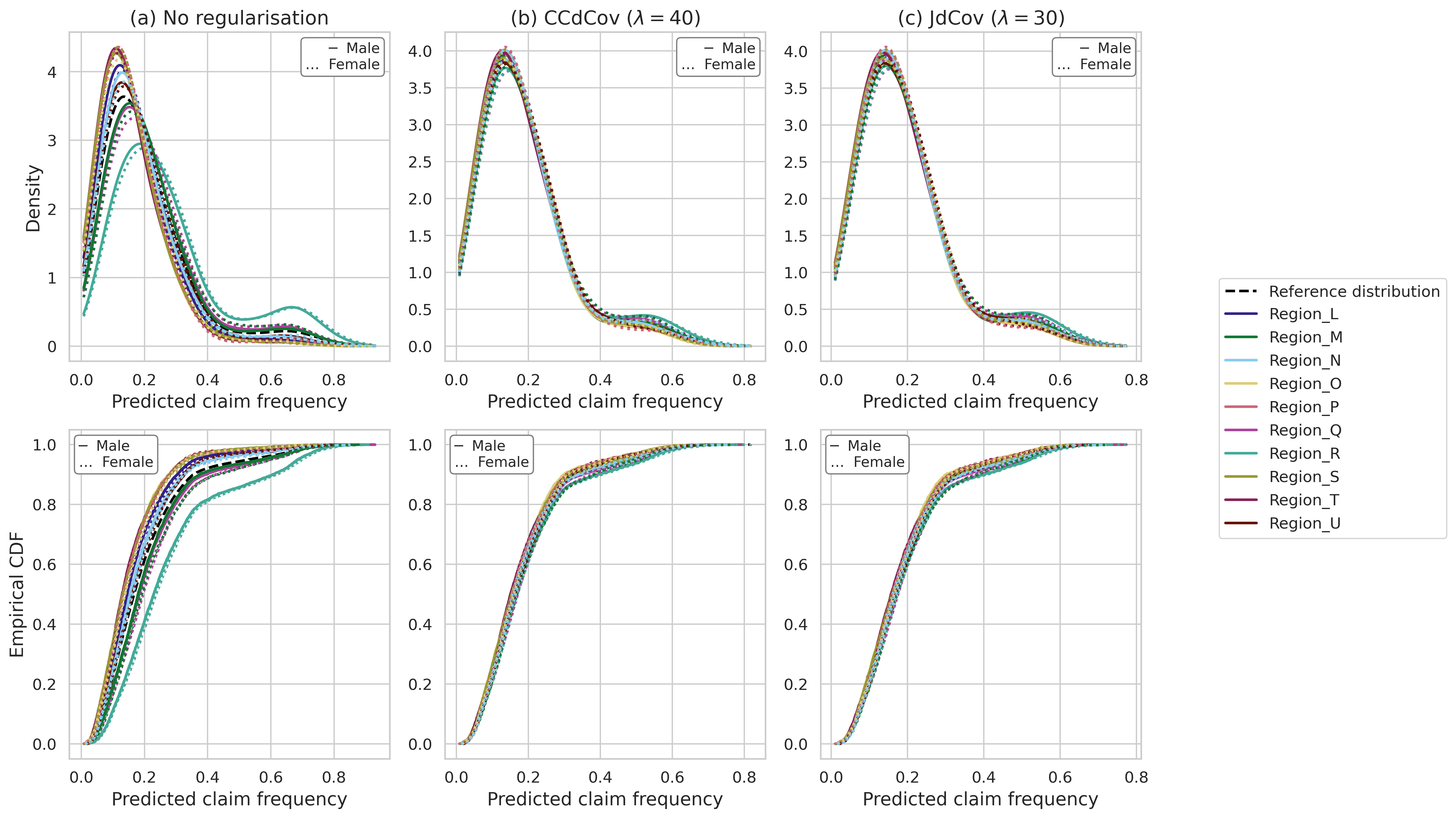}
    \caption*{     \begin{minipage}{0.95\textwidth}\textit{Note}: This figure presents the predicted claim frequency distributions of the training set under three regularisation settings: (a) no regularisation, (b) regularisation using CCdCov, and (c) regularisation using JdCov. The top row shows kernel density estimates (KDEs), while the bottom row displays the corresponding empirical CDFs on the test set. Each color corresponds to a specific region, and line styles distinguish gender (solid for male, dashed for female). The black dashed line represents the reference KDE of the unconditional prediction distribution $\mathbb{P}(\hat{\mathbf{Y}})$. The plots highlight how regularisation aligns the conditional prediction distributions across protected subgroups with the overall distribution.\end{minipage}}
\end{figure}

\begin{table}[ht]
    \centering
    \caption{Performance metrics of a regularised Poisson regressor evaluated on training set (\texttt{pg15training}) \label{tab:test_acc_pg15_train}}
    {{
\begin{tabular}{lcccccc}
\toprule
 & \multicolumn{2}{c}{(a) Accuracy metrics} & \multicolumn{4}{c}{(b) Fairness metrics} \\
\cmidrule(lr){2-3} \cmidrule(lr){4-7}
Model & RPS & Poisson deviance & CCdCov & JdCov & JS-divergence & $UF(\hat{\mathbf{Y}})$ \\
\midrule
No regularisation & 0.1185 & 0.5106 & 9.6276e-04 & 1.2568e-03 & 0.0478 & 0.0752 \\
CCdCov            & 0.1198 & 0.5207 & 4.2713e-05 & 6.7472e-05 & 0.0029 & 0.0041 \\
JdCov             & 0.1205 & 0.5266 & 3.7555e-05 & 5.9485e-05 & 0.0032 & 0.0050 \\

\bottomrule
\end{tabular}
    }}
    \vspace{0.2cm}
    \caption*{     \begin{minipage}{0.95\textwidth}\textit{Note}: Training set performance metrics for Poisson regression models trained with different fairness regularisers: no regularisation, concatenated distance covariance (CCdCov), and joint distance covariance (JdCov). Metrics are reported for the model selected with $\lambda = 40$ for CCdCov and $\lambda=30$ for JdCov, which yielded identical JS-divergence in the validation set. Panel (a) reports accuracy metrics, including the Ranked Probability Score (RPS) and Poisson deviance. Panel (b) reports fairness metrics. Both regularised models reduce disparities between predictions and protected attributes, with CCdCov offering slightly better accuracy under similar level of fairness.\end{minipage}}
\end{table}

\section{Further discussions in limited sample size}\label{apx:compas_discussion}
Assume that the results we have achieved so far are not achieving sufficient fairness for our purposes. In this instance, this is probably triggered by limited data. As discussed in the \texttt{COMPAS} application, when data is divided by multiple protected attributes, some subgroups inevitably contain very few samples, especially when the overall dataset is small. For example, in our data, there are fewer than 20 Hispanic male observations for any given integer age, since age is treated as a continuous variable. Typically, large datasets are required to achieve reliable generalisation.

We now explore two strategies to push our results towards more fairness, we explore two strategies: increasing the regularisation strength and applying oversampling. Note these could be applied in conjunction or separately of one another.

\begin{figure}[h!]
    \centering
    \caption{Average predicted recidivism rates regularised with CCdCov using limited sample size mitigation strategies \label{fig:compas_lim_sample}}
    \includegraphics[width=1\linewidth]{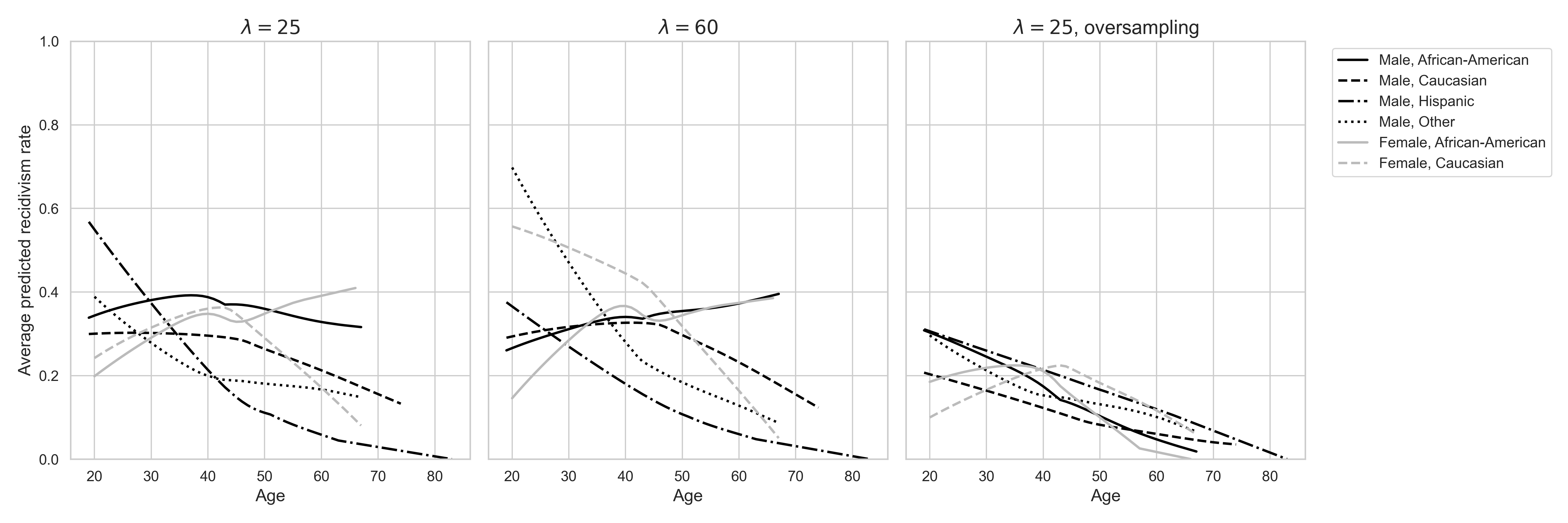}
    \caption*{     \begin{minipage}{0.95\textwidth}\textit{Note}: The figure displays average predicted recidivism rates by age, smoothed using lowess (span = 1), across gender--ethnicity subgroups. Models are regularised with CCdCov and evaluated under three strategies: (left) $\lambda = 25$, (middle) $\lambda = 60$, and (right) $\lambda = 25$ with resampling. Only gender-ethnicity subgroups with at least 100 training samples are shown. Stronger regularisation flattens predicted rates but fails to generalise at extreme ages. Resampling improves consistency across age by reinforcing rare subgroups.\end{minipage}}
\end{figure}

According to Figure~\ref{fig:compas_lim_sample}, increasing the regularisation strength from $\lambda = 25$ (left panel) to $\lambda = 60$ (middle panel) improves fairness, as shown by the flatter predicted recidivism rates across most age groups. However, disparities remain at the youngest and oldest ages, likely due to limited training data for those subgroups.

To mitigate this issue, we apply oversampling by duplicating training samples within each combination of age, gender, and ethnicity to ensure a minimum of 30 observations per group. We also apply regularisation with $\lambda=25$. The model is then re-trained (right panel). This enhances the effect of regularisation, particularly for small or underrepresented subgroups. For example, predicted rates for males of other ethnicity become more stable across age.

Nonetheless, oversampling has limitations. For instance, the predicted recidivism rate for older Hispanic males remains at zero, as individuals aged 83 appear only in the validation set and are not present in the training data. Since oversampling only replicates existing observations, it cannot create new age values. As a result, regularisation cannot reduce disparities in regions that are completely unrepresented in the training data.

In addition, we report the accuracy metrics for each model in Table \ref{tab:COMPAS_lim_sample}. As regularisation strength increases, model accuracy declines, reflecting a trade-off between fairness and predictive performance. Specifically, for the model with resampling, even with a lower regularisation strength of $\lambda = 25$, accuracy drops below that of the $\lambda = 60$ model without resampling. This is expected: by enforcing a stronger regularisation effect through oversampling, the model focuses more on fairness, further sacrificing predictive accuracy.

\begin{table}[ht]
    \centering
    \caption{Validation RPS of a regularised binary classifier with remedies for small subgroup sizes \label{tab:COMPAS_lim_sample}}
    {{
\begin{tabular}{lccc}
\toprule
Model & Original & Increased $\lambda$ & Resampling + original $\lambda$ \\
\midrule
CCdCov & 0.2180 & 0.2225 & 0.2333 \\
JdCov  & 0.2203 & 0.2214 & 0.2347  \\
\bottomrule
\end{tabular}
    }}
    \vspace{0.2cm}
    \caption*{     \begin{minipage}{0.95\textwidth}\textit{Note}: Validation RPS for binary classifiers trained with regularisation using CCdCov and JdCov under different settings: with the original model, with increased $\lambda$, and with original $\lambda$ and oversampling. Higher $\lambda$ and resampling both increase the RPS, indicating a lower predictive accuracy.\end{minipage}}
\end{table}
\begin{figure}[h!]
    \centering
    \caption{Average predicted recidivism rates regularised with JdCov using limited sample size mitigation strategies \label{fig:compas_lim_sample2}}
    \includegraphics[width=1\linewidth]{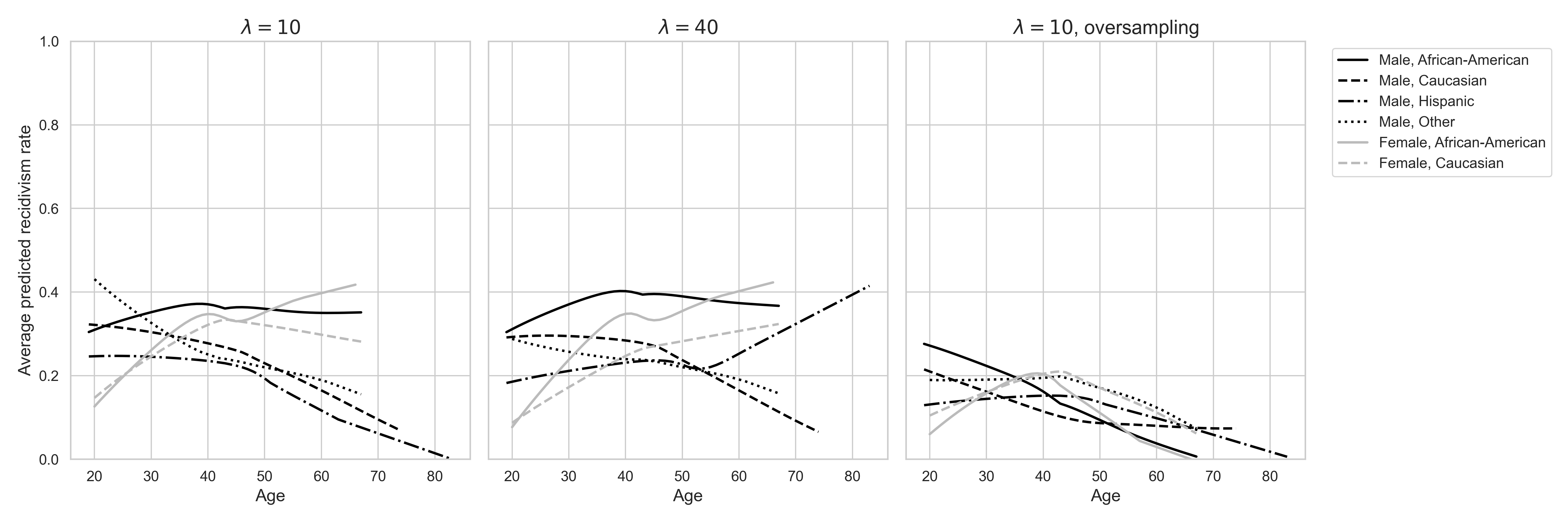}
    \caption*{     \begin{minipage}{0.95\textwidth}\textit{Note}: The figure shows average predicted recidivism rates by age under the same setting as Figure~\ref{fig:compas_lim_sample}, but using JdCov as the regulariser. In this case, increasing $\lambda$ does not reduce disparities by much. Moreover, predictions for underrepresented age groups remain poorly generalised as $\lambda$ increases. In contrast, applying oversampling with $\lambda = 10$ results in more aligned predicted rates across subgroups.\end{minipage}}
\end{figure}

We now discuss the effect of limited sample size remedies for models regularised with JdCov. Figure~\ref{fig:compas_lim_sample2} shows the average predicted recidivism rates under different settings.

When a stronger regularisation is applied (middle panel), the regularisation effect appears to deteriorate. The average predicted rates begin to increase with age for some subgroups. This are two potential reasons: first, the JdCov regulariser does not exclusively penalise the dependence between model predictions and the joint distribution of the protected attributes. It also captures higher order associations, which may divert the regularisation focus away from improving fairness across subgroups. Second, as discussed in Section~\ref{sec:jdreg}, JdCov can cause numerical instability when it penalises dependencies that cannot be removed through training.

With oversampling and $\lambda=10$ (right panel), the average predicted rates flatten across age for most subgroups. However, the curve for African-American females still shows a noticeable hump. This is likely due to the limited sample size in that subgroup. As shown in Table~\ref{tab:COMPAS_lim_sample}, oversampling with regularisation again yields the least accurate predictions.
\end{appendices}

\clearpage
\section*{Online Appendix}
\addcontentsline{toc}{section}{Online Appendix}
\noindent\rule{\textwidth}{1pt}
\small

\setcounter{section}{0}
\setcounter{subsection}{0}
\renewcommand{\thesection}{\Alph{section}}
\renewcommand{\thesubsection}{\thesection.\arabic{subsection}}

\section{Numerical example of decomposition in Theorem 3.2.1}\label{oapx:thm1}
We provide two examples illustrating how the correction term $\eta$ in Equation (3.9) captures the association between a random variable and the joint distribution of other random variables.

First, consider three random variables $X$, $Y$, and $Z$, where
\begin{subequations}
\begin{align}
    X&\sim\mathcal{N}(0,1),\\
    Y&\sim\mathcal{N}(0,1),\\
    Z&=sgn(XY).
\end{align}
\end{subequations}
Under this setting, $ Z $ is a discrete random variable taking values in $\{-1, 1\}$, with $ \mathbb{P}(Z = 1) = \mathbb{P}(Z = -1) = 0.5 $. Moreover, we have the marginal independencies $ X \perp Y $, $ Z \perp X $, and $ Z \perp Y $. However,  $ Z $ can be determined by the joint values of $ X $ and $ Y $, meaning that $ Z $ is associated with the joint distribution $ (X, Y) $.

In this case, both $ \widetilde{dCov}^2(Z, X) $ and $ \widetilde{dCov}^2(Z, Y) $ should be close to zero, since $ Z $ has no marginal association with $ X $ or $ Y $ individually. In contrast, the correction term $ \eta $ should capture the dependence between $ Z $ and the joint distribution $ (X, Y) $, ensuring that $ \widetilde{dCov}^2(Z, (X, Y)) $ properly reflects this higher-order association.

We simulated 1000 samples and obtained the following:
    \begin{equation}
    \mbox{\small $
        \begin{aligned}
            &\widetilde{dCov}^2(Z, (X, Y)) \\
            &\qquad= \widetilde{dCov}^2(Z,X) + \widetilde{dCov}^2(Z,Y)+\eta(Z,(X,Y))\\
            &\qquad= 0.0003 + (-0.0005) + 0.0472= 0.0470.
        \end{aligned}$
        }
    \end{equation}
As shown, the distance covariances between $Z$ and $X$, and between $Z$ and $Y$, are both close to zero, indicating the marginal associations are negligible. However, the correction term $\eta$ captures the higher-order association between $Z$ and the joint distribution $(X, Y)$, resulting in a non-negligible overall distance covariance.

On the other hand, it is important to show that $\eta$ does not capture extra associations that may exist within the protected attributes themselves (e.g., association between the categorical variables \texttt{Gender} and \texttt{Ethnicity}). Otherwise, we may face numerical instability in our model training as discussed in Section 3.2.2. 

Consider another example. Let $(X,Y,Z)\sim \mathcal{N}(\mu,\Sigma)$ be multivariate Gaussian with
\begin{equation}
    \mu = \begin{bmatrix}
        0 \\
        0\\
        0
    \end{bmatrix},\quad
    \Sigma = \begin{bmatrix}
        1 & 0.8 & 0\\
        0.8 & 1 & 0\\
        0 & 0 & 1
    \end{bmatrix}.
\end{equation}
In this case, $X$ and $Y$ are correlated, but $Z$ is independent of both $X$ and $Y$, as well as their joint distribution $(X, Y)$. As a result, the measures $\widetilde{dCov}^2(Z, X)$, $\widetilde{dCov}^2(Z, Y)$, and $\widetilde{dCov}^2(Z, (X, Y))$ should all be close to $0$. This also implies that $\eta$, which captures the additional association between $Z$ and $(X, Y)$ beyond the individual associations, should also be close to $0$. 

Through 1000 simulated samples, we have
\begin{equation}
\mbox{\small$
    \begin{aligned}
        &\widetilde{dCov}^2(Z, (X, Y)) \\
        &\qquad= \widetilde{dCov}^2(Z,X) + \widetilde{dCov}^2(Z,Y) +\eta(Z,(X,Y))\\
        &\qquad= (-0.0006) + (-0.0003) + 0.0002 = -0.0007.
    \end{aligned}
    $
}
\end{equation}
These two examples provide an intuitive justification for including $\eta$ in the regulariser. Including this term ensures that the regulariser captures associations between the model output and protected attributes both marginally and jointly, while ignoring other irrelevant associations.

\section{Details on feed-forward neural network}\label{oapx:ffnn}
We implement a feed-forward neural network (FFNN), also known as a multilayer perceptron (MLP), for both classification and regression tasks. The architecture consists of an input layer, multiple hidden layers, and an output layer. Each layer applies a linear transformation followed by a nonlinear activation function.

Let $\mathbf{x} \in \mathbb{R}^p$ denote the input vector. The FFNN consists of $K$ hidden layers. For each hidden layer $k = 1, \dots, K$, let $\mathbf{W}^{[k]}$ and $\mathbf{b}^{[k]}$ denote the weight matrix and bias vector respectively. The transformation at layer $k$ is given by:
\begin{equation}
    \mathbf{z}^{[k]} = \sigma\left(\mathbf{W}^{[k]} \mathbf{z}^{[k-1]} + \mathbf{b}^{[k]}\right),
\end{equation}
with $\mathbf{z}^{[0]} = \mathbf{x}$, and $\sigma(\cdot)$ denoting a nonlinear activation function.

The output layer performs a final transformation:
\begin{equation}
    \hat{\mathbf{y}} = g\left(\mathbf{W}^{[K+1]} \mathbf{z}^{[K]} + \mathbf{b}^{[K+1]}\right),
\end{equation}
where $g(\cdot)$ is an activation function specific to the prediction task.

\section{Details on AdaHessian}\label{oapx:adahessian}
Let $\Theta = \begin{bmatrix}
    \theta_1 & \ldots &\theta_m
\end{bmatrix} \in \mathbb{R}^m$ denote the vector of all learnable model parameters, which includes all scalar entries of $\mathbf{W}^{[1]}, \mathbf{b}^{[1]}, \ldots, \mathbf{W}^{[K+1]}, \mathbf{b}^{[K+1]}$. Let $\mathcal{L}(\Theta)$ be the loss function evaluated on the training data. Our objective is to minimise $\mathcal{L}(\Theta)$ with respect to $\Theta$. To achieve this, we adopt the AdaHessian algorithm \citep{yao2021adahessian}, a second-order optimisation method that leverages both gradient and curvature (Hessian) information of the loss function.

We first initialise the parameters with some starting value $\Theta_0$. Then, let $\mathbf{g}_t = \nabla_\Theta \mathcal{L}(\Theta_t)$ denote the gradient vector and $\mathbf{H}_t = \nabla^2_\Theta \mathcal{L}(\Theta_t)$ the Hessian matrix at iteration $t$. The parameter vector is updated using: 
\begin{equation} 
\Theta_{t+1} = \Theta_t - \eta_t \cdot \frac{\mathbf{m}_t}{\mathbf{v}_t}, 
\end{equation} 
where $\eta_t$ is the learning rate, and the first and second moment estimates are defined as: 
\begin{subequations} 
\begin{align} 
\mathbf{m}_t &= \frac{(1 - \beta_1) \sum_{i=1}^t \beta_1^{t-i} \mathbf{g}_i}{1 - \beta_1^t}, \\
\mathbf{v}_t &= \sqrt{ \frac{(1 - \beta_2) \sum_{i=1}^t \beta_2^{t-i} \mathbf{D}_i^{(s)} \mathbf{D}_i^{(s)}}{1 - \beta_2^t} }, 
\end{align} 
\end{subequations}
with $\mathbf{D}_i^{(s)}$ denoting the spatially averaged approximation to the diagonal of the Hessian $\text{diag}(\mathbf{H}_i)$ computed using the method described in Equation (10) of \citet{yao2021adahessian}, and where $\beta_1$ and $\beta_2$ are first-order and second-order moment hyperparameters that can be tuned respectively.

\section{Details on complementary accuracy metrics}\label{oapx:performance_acc}
\vspace{0.2cm}
\begin{enumerate}
    \item \underline{Wilcoxon signed-ranked test \citep{wilcoxon1992individual}}: To compare model accuracy before and after regularisation, we apply the one-sided Wilcoxon signed-rank test, a nonparametric test for comparing paired samples. Let ${\hat{\mathbf{y}}_{0,i}}$ and ${\hat{\mathbf{y}}_{\lambda,i}}$ denote the predictions for each observation $i = 1, \ldots, n$ under the unregularised model and the model with regularisation strength $\lambda$ respectively. We compute the RPS for each prediction to obtain two paired sets of values: ${RPS_{0,i}}$ and ${RPS_{\lambda,i}}$.
    We aim to test whether regularisation results in a statistically significant increase in RPS. The hypotheses are:
    \begin{equation} 
    \begin{aligned} 
    H_0 &: \text{Median}(RPS_{\lambda,i} - RPS_{0,i}) = 0 \\
    H_A &: \text{Median}(RPS_{\lambda,i} - RPS_{0,i}) > 0, 
    \end{aligned} 
    \end{equation}
    where rejecting $H_0$ indicates that the regularised model has significantly higher RPS (lower accuracy) than the unregularised one.
    The test statistic is computed as follows:
    \begin{enumerate} 
    \item Compute the differences $d_i = RPS_{\lambda,i} - RPS_{0,i}$. 
    \item Remove any $d_i = 0$. 
    \item Rank the absolute differences $|d_i|$ in increasing order and  assign average ranks for ties. 
    \item Apply the sign of each $d_i$ to its corresponding rank. 
    \item Compute $W^+ = \sum \text{Ranks of all positive } d_i$.
    \end{enumerate}
    Under $H_0$, the distribution of $W^+$ can be approximated by a normal distribution. The $p$ value is computed as:
    \begin{equation} \mathbb{P}\left(Z > \frac{W^+ - \frac{n(n+1)}{4} - 0.5}{\sqrt{\frac{n(n+1)(2n+1)}{24}}} \right), \quad Z \sim \mathcal{N}(0, 1), \end{equation}
    where $n$ is the number of nonzero differences and the continuity correction term $-0.5$ improves the approximation.
    
    \item \underline{Binary classification accuracy (ACC)}: In binary classification tasks, ACC measures the overall correctness of a classifier. In our \texttt{COMPAS} application (Section 5.1), we classify whether an individual will recidivate within two years. We define individuals who do recidivate as having a true label, and those who do not as having a false label. Let:
    \begin{itemize} 
    \item $TP$: True positives — the number of instances correctly predicted as true, where the actual label is also true. 
    \item $TN$: True negatives — the number of instances correctly predicted as false, where the actual label is false. 
    \item $FP$: False positives — the number of instances incorrectly predicted as true, where the actual label is false. 
    \item $FN$: False negatives — the number of instances incorrectly predicted as false, where the actual label is true. 
    \end{itemize}
    Then, the classification accuracy is given by: 
    \begin{equation} 
    \text{ACC} = \frac{TP + TN}{TP + TN + FP + FN}, 
    \end{equation}
    and a higher value of ACC represents a better classification performance.

    \item \underline{Poisson deviance}: In count-based prediction tasks, such as our claim counts application in Section 5.2, we can assess the model performance using Poisson deviance, which measures the discrepancy between predicted and observed counts under the Poisson assumption. Let $y_i$ denote the observed count and $\hat{y}_i$ the predicted mean for the $i$-th observation. The Poisson deviance for a prediction is defined as: 
    \begin{equation} 
    D_i = 2\left[y_i \log\left(\frac{y_i}{\hat{y}_i}\right) - (y_i - \hat{y}_i)\right], 
    \end{equation}
    with $y_i \log(y_i/\hat{y}_i) = 0$ when $y_i = 0$.
    The average Poisson deviance over $n$ observations is given by: \begin{equation} 
    \text{Poisson Deviance} = \frac{1}{n}\sum_{i=1}^{n} D_i. 
    \end{equation}
    Here, a lower Poisson Deviance indicates better alignment between predicted and observed counts.
\end{enumerate}

\end{document}